\documentclass{article}

\PassOptionsToPackage{numbers, compress}{natbib}

\usepackage[eandd, final]{neurips_2026}

\usepackage[utf8]{inputenc} 
\usepackage[T1]{fontenc}    
\usepackage{hyperref}       
\usepackage{etoc}           
\usepackage{url}            
\usepackage{booktabs}       
\usepackage{array}          
\usepackage{amsfonts}       
\usepackage{amssymb}        
\usepackage{amsmath}        
\usepackage{graphicx}       
\usepackage{nicefrac}       
\usepackage[protrusion=false,expansion=false]{microtype} 
\usepackage{xcolor}         
\usepackage[most]{tcolorbox} 
\usepackage{caption}        
\usepackage{subcaption}
\usepackage{textcomp}       

\title{EHRNote-ChatQA: A Benchmark for Evidence-Grounded Multi-Turn Clinical Question Answering over Longitudinal Discharge Summaries}

\author{%
  \textbf{Jiyoun Kim$^{1}$},
  \textbf{Muhan Yeo$^{2,3}$},
  \textbf{Eunhye Jang$^{4}$}, 
  \textbf{Jeewon Yang$^{1}$}, \\
  \textbf{Hangyul Yoon$^{1}$},
  \textbf{Su Ji Lee$^{5}$}, 
  \textbf{Hee Jo Han$^{5,6}$},
  \textbf{Hee-Jae Jung$^{7}$},
  \textbf{Doyun Kwon$^{8}$},\\
  \textbf{Jun young Lee$^{3,9}$},
  \textbf{Jaehun Lee$^{10}$},
  \textbf{Jung-Oh Lee$^{11}$},
  \textbf{Sunjun Kweon$^{1}$}, \\
  \textbf{Jong Hak Moon$^{1}$},
  \textbf{Daseul Kim$^{12}$},
  \textbf{Minjae Cho$^{3}$},
  \textbf{Edward Choi$^{1}$}\thanks{Corresponding author} \\
  $^{1}$KAIST \;
  $^{2}$Seoul National University \;
  $^{3}$Seoul National University Bundang Hospital \; \\
  $^{4}$SAIHST, Sungkyunkwan University \;
  $^{5}$Yonsei University College of Medicine \; \\
  $^{6}$Severance Hospital \;
  $^{7}$Seoul Medical Center \;
  $^{8}$Gangnam Severance Hospital \;\\
  $^{9}$Seoul National University Hospital \; 
  $^{10}$National Cancer Center \;\\
  $^{11}$Icahn School of Medicine at Mount Sinai \;
  $^{12}$Samsung Medical Center \\
  \texttt{\{jiyoun.kim, edwardchoi\}@kaist.ac.kr}
}

\begin{document}

\maketitle

\begin{abstract}
Discharge summaries are crucial clinical documents containing the context of a patient's overall hospital stay, and are routinely reviewed by medical experts for patient readmission, ongoing care, and diagnostic decision-making. 
When reviewing them, medical experts often must iteratively synthesize information across multiple discharge summaries while verifying the evidence supporting each answer.
Although large language models (LLMs) are increasingly explored for clinical question answering, existing benchmarks do not sufficiently reflect this setting: they often evaluate exam-style medical knowledge or focus on single-turn question answering with limited evidence-grounding evaluation. 
We introduce \textbf{EHRNote-ChatQA}, the first benchmark for evidence-grounded multi-turn clinical question answering over patients' multiple discharge summaries. 
Built from de-identified MIMIC-IV discharge summaries, EHRNote-ChatQA contains 967 patient-level multi-turn samples spanning one to five notes and 16,072 medical-expert-verified QA pairs (8,036 content questions, each paired with an evidence-grounding question) across eight clinical categories. 
The benchmark is constructed through an expert-informed pipeline combining discharge-summary structuring schema, expert-curated multi-turn QA templates, and LLM-based generation, followed by review and revision of every single QA sample by 12 medical experts (11 physicians and 1 nurse). 
Benchmarking 24 proprietary and open-weight LLMs reveals several challenges, including that LLMs struggle more with evidence grounding than content answering, multi-turn errors compound across turns, and single-turn clinical QA performance does not reliably transfer to this setting. 
These findings establish EHRNote-ChatQA as a rigorous and practical benchmark for evaluating clinical QA systems.
The dataset is available through PhysioNet credentialed access at \url{https://doi.org/10.13026/x533-f455}, and the code is available at \url{https://github.com/ji-youn-kim/EHRNote-ChatQA}.
\end{abstract}

\etocdepthtag.toc{mainmatter}

\section{Introduction}

Patient discharge summaries are comprehensive clinical documents that describe a patient's entire clinical course from admission to discharge. 
They record clinical events such as diagnoses, medications, procedures, and laboratory tests, along with contextual information including causal relationships, diagnostic reasoning, and patient outcomes.
These documents are routinely reviewed by medical experts (e.g., physicians, nurses) to understand a patient's prior clinical status in settings such as readmission, longitudinal care, and diagnostic decision-making.
However, discharge summaries are often written by different physicians, follow inconsistent formats, and contain extensive medical abbreviations.
Moreover, a new discharge summary is generated for each admission, accumulating multiple notes for a single patient over time.
As a result, reviewing a patient's discharge summaries requires navigating long, fragmented documents, making it difficult to extract relevant information.

Recent advances in large language models (LLMs) have enabled question answering (QA) over clinical documents~\cite{singhal2023large, nori2023capabilities}, allowing medical experts to retrieve targeted information without manually reviewing each note.
Yet, for LLMs to be safely integrated into clinical workflows, two fundamental requirements must be met: (1) accurately answering medical experts’ iterative, multi-turn questions, and (2) grounding each answer in traceable evidence from the source documents.

\begin{figure}[t]
\centering
\includegraphics[width=\linewidth]{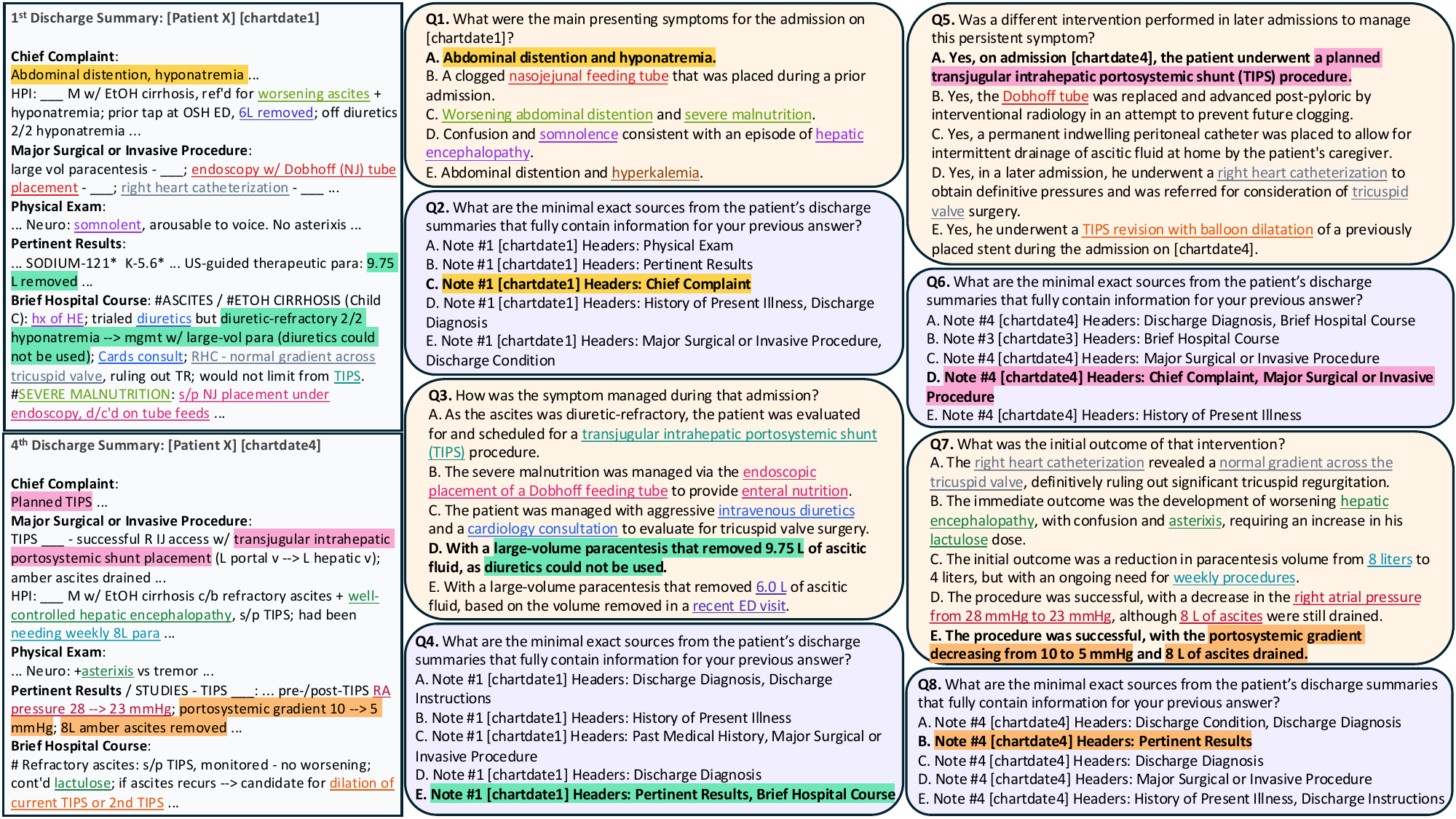}
\caption{An illustrative multi-turn QA sample of EHRNote-ChatQA (only parts of Notes~1 and~4 are shown for readability; the actual chart date of Note \#N is replaced with [chartdateN]). Given a single patient's entire set of discharge summaries, EHRNote-ChatQA consists of a sequence of \emph{content} questions about the notes (Q1, Q3, Q5, Q7), each paired with an \emph{evidence-grounding} question that requires identifying the exact source supporting the preceding answer (Q2, Q4, Q6, Q8).}
\label{fig:intro-example}
\end{figure}

In practice, clinical inquiries are rarely isolated; they are inherently multi-turn~\cite{ely2005answering}, typically starting with an initial query and proceeding through related follow-ups that build on prior context.
For example, a medical expert may first ask, ``What symptoms led to the patient's admission?'', followed by ``What did the evaluation of those symptoms reveal?'', and then ``Did any of those symptoms precipitate the patient's next hospitalization?''.
Moreover, beyond answer correctness, traceability back to source is equally critical in clinical settings, since LLMs may generate hallucinated or unverifiable responses~\cite{ji2023survey,pal2023med}.
It is therefore essential to evaluate whether model output can be grounded in verifiable evidence~\cite{bohnet2022attributed}, letting users trace each answer to specific segments of the source documents.

Despite these requirements, existing clinical QA benchmarks fail to capture key aspects of real-world clinical usage.
Benchmarks such as MedQA~\cite{jin2021disease}, MedMCQA~\cite{pal2022medmcqa}, and PubMedQA~\cite{jin2019pubmedqa} evaluate medical knowledge through exam-style questions and do not involve patient records.
Electronic health record (EHR)-based benchmarks such as EHRNoteQA~\cite{kweon2024ehrnoteqa}, emrQA~\cite{pampari2018emrqa}, and DiSCQ~\cite{lehman2022learning} include expert-curated questions over discharge summaries, but are limited to single-turn QA and lack evidence-grounding questions that ask for the source supporting each answer.

To address this gap, we introduce \textbf{EHRNote-ChatQA}, the first evidence-grounded, multi-turn, multi-note clinical QA benchmark over patient discharge summaries.
Built from de-identified MIMIC-IV~\cite{johnson2023mimic} discharge summaries, our benchmark covers 967 patients (one to five notes each), yielding 967 multi-turn samples with 16,072 QA pairs: 8,036 content questions, each paired with an evidence-grounding question that identifies the supporting source location.
Each sample provides the patient's full sequence of discharge summaries as context. 
Examples are shown in Appendix~\ref{app:example}.

To generate multi-turn questions that reflect medical expert inquiries in clinical practice, with clinically accurate answers grounded in the patient's notes, we develop an expert-informed, structured QA generation pipeline in collaboration with 4 medical experts (3 physicians and 1 nurse).
The pipeline uses a comprehensive discharge-summary structuring schema and multi-turn QA templates (both defined with the experts) as seeds for LLM-based generation with Gemini 2.5 Pro.
For each content question, we also generate a paired evidence-grounding question asking for the source location supporting the answer.
Every QA sample is then reviewed and revised by 12 medical experts (11 physicians and 1 nurse) over three weeks to ensure clinical naturalness, accuracy, and evidence-grounding reliability.

Using this benchmark, we comprehensively evaluate 24 proprietary and open-weight LLMs.
Our evaluations reveal several findings, including the following: (1) models exhibit a substantial gap between content and evidence-grounding accuracy; (2) errors propagate across turns; and (3) strong performance on existing single-turn benchmarks does not reliably transfer to multi-turn evaluation.

Our main contributions are as follows:
\begin{itemize}
\item \textbf{Benchmark.} We introduce EHRNote-ChatQA, the first evidence-grounded, multi-turn, multi-note clinical QA benchmark over patient discharge summaries, comprising 967 multi-turn samples and 16,072 medical-expert-verified QA pairs across eight categories.
\item \textbf{Dataset Construction Pipeline.} We develop a structured, medical-expert-informed pipeline for generating clinically grounded multi-turn QA data, combining a discharge-summary structuring schema, expert-curated multi-turn QA templates, and LLM-based generation, followed by medical expert validation of every single QA sample.
\item \textbf{Evaluation.} We benchmark 24 LLMs and conduct a comprehensive analysis, including content and evidence-grounding accuracy, multi-turn dependency and error propagation, comparison with existing single-turn clinical QA benchmark, and comparison between multiple-choice and open-ended LLM-as-a-judge evaluation.
\end{itemize}

\section{Related Work}

\paragraph{Multi-turn QA Benchmarks.}
Several benchmarks evaluate LLMs on multi-turn dialogue and sequential QA.
MT-Bench~\cite{zheng2023judging} and MT-Eval~\cite{kwan2024mt} cover general-domain conversational topics (e.g., writing, reasoning, STEM) and use GPT-4 as a judge to score open-ended responses.
CoQA~\cite{reddy2019coqa} and QuAC~\cite{choi2018quac} are reading-comprehension dialogue benchmarks in which a model answers a sequence of questions grounded in a single text passage. 
While these benchmarks have been central to evaluating multi-turn question answering, none target patient electronic health records (EHRs), and none include evidence-grounding questions asking for the source document of each answer.

\paragraph{Clinical QA Benchmarks.}

Existing clinical QA benchmarks cover diverse settings, including medical knowledge evaluation and question answering over patient records. 
Benchmarks such as MedQA~\cite{jin2021disease} (USMLE), MedMCQA~\cite{pal2022medmcqa} (Indian medical entrance exams), and PubMedQA~\cite{jin2019pubmedqa} (PubMed-abstract comprehension) evaluate the parametric medical knowledge of LLMs through exam-style or literature-comprehension questions, without grounding in a specific patient's record. 
More closely related to our setting, benchmarks such as emrQA~\cite{pampari2018emrqa}, EHRNoteQA~\cite{kweon2024ehrnoteqa}, and DiSCQ~\cite{lehman2022learning} are built on patient EHR notes (e.g., discharge summaries), but are limited to single-turn questions and do not pair each content question with an evidence-grounding question asking for the source evidence supporting the answer.

\paragraph{Evidence-Grounding Benchmarks.}
Evidence grounding evaluation has also been studied in general-domain QA and long-form generation. 
Attributed QA~\cite{bohnet2022attributed} evaluates whether a model can provide both an answer and a supporting attribution to a source corpus, while ALCE~\cite{gao2023enabling} evaluates LLM generations with citations over open-domain QA tasks such as ASQA~\cite{stelmakh2022asqa}, QAMPARI~\cite{amouyal2023qampari}, and ELI5~\cite{fan2019eli5}. 
These benchmarks emphasize the importance of verifiable question answering, but they are not specific to clinical records and do not evaluate multi-turn QAs. 
Among patient-record benchmarks, emrQA~\cite{pampari2018emrqa} includes answer evidence pairs derived from clinical annotations, and ArchEHR-QA~\cite{soni2025overview} provides sentence-level evidence annotations for patient-style questions over discharge-summary excerpts. 
However, neither spans the patient's multiple discharge summaries (they are only single note) or evaluates evidence grounding under multi-turn dialogue.

To our knowledge, no prior benchmark combines the three properties EHRNote-ChatQA targets: multi-turn questions reflecting realistic clinical record review of medical experts, multi-note grounding across a patient's longitudinal discharge summaries, and per-turn evidence-grounding questions that identify the source supporting each answer.
Table~\ref{tab:dataset-comparison} (in Appendix~\ref{app:dataset-comparison}) compares EHRNote-ChatQA with prior patient-record QA benchmarks along these dimensions.

\section{Methods}

\subsection{Data Source and Cohort Construction}\label{sec:cohort}

We use de-identified discharge summaries from MIMIC-IV~\cite{johnson2023mimic}, a database of patient electronic health records (EHRs) from Beth Israel Deaconess Medical Center in Boston, USA.
As approximately 90\% of patients in MIMIC-IV have 1--5 discharge summaries, we sampled 1{,}000 patients across five note-count groups, with 200 patients per group.
We further assigned patients uniformly across eight expert-defined question categories (see Section~\ref{sec:schema-templates})---Diagnosis, Symptom, Procedure, Medication, Microbiology, Clinical Assessment, Clinical Outcome, and Discharge Plan---yielding 25 patients for each category--note-count combination.

\subsection{Discharge Summary Structuring Schema and Multi-turn QA Template Design}\label{sec:schema-templates}

Our benchmark is built around a central principle: the multi-turn questions should closely reflect how medical experts actually inquire and follow up when reviewing discharge summaries in clinical practice, and the answers must be clinically accurate and grounded in the patient's notes.

To this end, we develop a data generation pipeline in collaboration with 4 medical experts (3 physicians and 1 nurse).
We first define a schema for structuring discharge summary content, consisting of 19 clinical entity types, 3--10 attributes per entity type, and 23 relationship types (full schema in Appendix~\ref{app:schema}).
Although prior work has proposed partial schemas for structuring discharge summaries~\cite{uzuner2010extracting,uzuner20112010,sun2013evaluating,henry20202018}, these omit several important entity types (e.g., microbiology, allergy, activity) and merge clinically distinct concepts into broad categories---for example, diagnosis, symptom, and finding under a ``Problem'' entity type, and procedure and medication under a ``Treatment'' entity type.
Such coarse-grained representations limit the detailed clinical attributes and relationships needed for realistic clinical questioning.
We therefore build upon prior discharge-summary schemas and refine them with medical experts to obtain more comprehensive and granular representations.

Together with the medical experts, we also define eight question categories reflecting common inquiries during discharge-summary review: Diagnosis, Symptom, Procedure, Medication, Microbiology, Clinical Assessment, Clinical Outcome, and Discharge Plan.
For each category, we then design multi-turn QA templates with these experts so that the questions reflect how clinicians inquire about patient information when reviewing discharge summaries.
Each template is built on the structuring schema: entities and attributes support entity-specific questions (e.g., a medication's dosage or route), and relationships link entities across turns (e.g., the symptoms and outcomes that followed a medication), so that each answer is grounded in the patient’s discharge summaries.
This reflects how clinicians review discharge summaries: identifying what occurred during the patient's care, asking about the details, and tracing causal, indication, and temporal relations across these events.
Each category has five single-note and five multi-note templates (80 in total: 8 categories $\times$ 2 settings $\times$ 5), where the multi-note templates capture cross-admission inquiry patterns such as disease course and discharge-plan evolution.
The complete set of templates is provided in Appendix~\ref{app:template}.

Note that these QA templates are not used to generate samples by rule-based slot filling.
Instead, they serve as examples of clinically realistic inquiry patterns, which the LLM (Gemini-2.5-Pro~\cite{comanici2025gemini}) uses as seeds to generate patient-specific, contextually appropriate multi-turn QA samples from each patient's discharge summaries.

\begin{figure}[t]
\centering
\includegraphics[width=\linewidth]{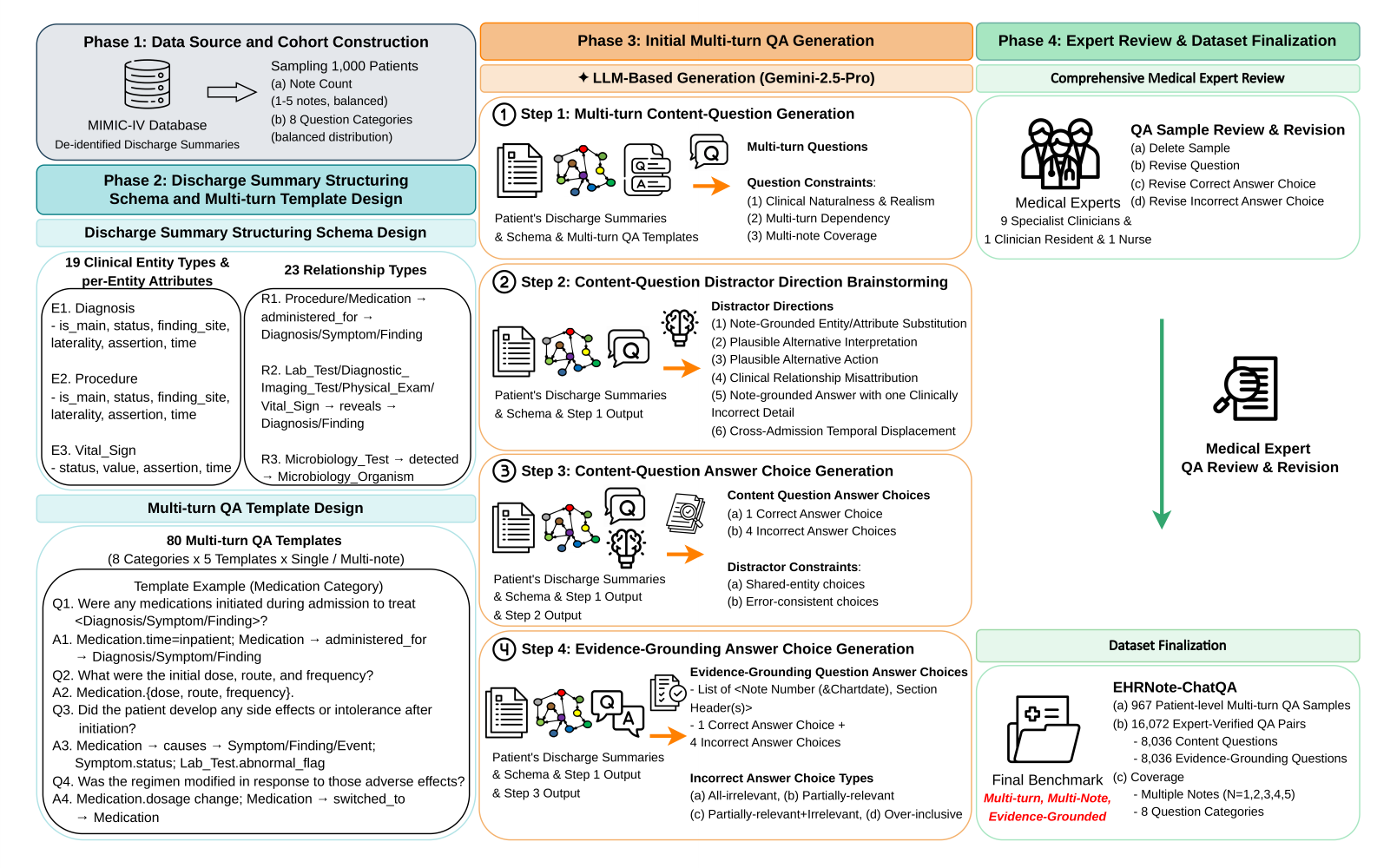}
\caption{Overview of the EHRNote-ChatQA construction pipeline.}
\label{fig:method-overview}
\end{figure}

\subsection{Initial Multi-turn QA Generation}\label{sec:generation}

Using the expert-defined schema and multi-turn QA templates, we generate initial QA samples with Gemini-2.5-Pro.
The QA samples contain two question types: (1) \emph{content questions} that ask about the clinical content of the discharge summaries, and (2) \emph{evidence-grounding questions}, each paired with a content question, that ask which part of the source notes supports the answer to the content question.
All questions are generated in a 5-way multiple-choice format (one correct choice, four incorrect answer choices).
We use multiple-choice as the primary evaluation format because (1) it enables objective, consistent evaluation via rule-based grading; and (2) it is scalable: rule-based grading avoids the costs of both medical expert manual review and LLM-as-a-judge evaluation.
In Section~\ref{sec:openended}, we additionally evaluate models in an open-ended setting with an LLM judge.

Each sample is generated through a four-step pipeline: (1)~generate multi-turn content questions that reflect medical expert inquiries; (2)~brainstorm plausible directions for incorrect answers to each content question; (3)~generate one correct and four incorrect content-answer choices; and (4)~generate the evidence-grounding answer choices.
We detail the design principles below (see Appendix~\ref{app:generation-details}--\ref{app:step-outputs}, \ref{app:prompts} for additional details on the generation process and prompts):

\textbf{(1) Multi-turn content questions (Step~1)}. We prompt Gemini-2.5-Pro to generate the multi-turn content questions given a single patient's full sequence of discharge summaries, under the following constraints in priority order. \textbf{(a)~Clinical naturalness and realism} (highest priority): questions must reflect inquiries that medical experts realistically pose when reviewing discharge summaries. The model is seeded with the expert-curated multi-turn QA templates as exemplars and is encouraged to include \emph{why}/\emph{how} reasoning questions when the reasoning can be inferred from the patient's notes. Every question must require both clinical knowledge \emph{and} the patient's discharge-summary context to answer: we forbid questions answerable from external medical knowledge alone, and encourage questions that require clinical reasoning to interpret the notes rather than purely surface-level text retrieval. \textbf{(b)~Multi-turn dependency}: answering a later question should require referencing prior questions and/or their answers, so that the multi-turn questions capture the cumulative context of chart review rather than collapsing into independent single-turn questions. \textbf{(c)~Multi-note coverage}: for patients with multiple discharge summaries, questions should span across multiple notes when clinically meaningful questions can be asked (e.g., medication changes or diagnosis evolution across admissions). However, clinical naturalness takes priority; if a note cannot be covered without forcing a contrived question, it should not be generated.

\textbf{(2) Content-question answer choices (Steps~2--3)}. We prompt Gemini-2.5-Pro to first brainstorm plausible distractor directions per question (Step~2), then generate one correct and four incorrect content-answer choices (Step~3). \textbf{(a)~Correct answer choice}: should provide a complete answer to the question, grounded in the patient's notes (Generated in Step~3 with the four incorrect answer choices per question). \textbf{(b)~Incorrect answer choices}: For brainstorming in Step~2, six types of plausible distractor directions are defined with medical experts (see Appendix~\ref{app:generation-details}). These directions are derived from one or both of two sources: (1) \textbf{note-grounded} distractors, which are incorrect interpretations of the patient's notes and (2) \textbf{clinical-knowledge-grounded} distractors, which are content not present in the patient's notes but clinically plausible given the patient's clinical context. Together, these sources ensure correct answering requires both interpreting discharge-summary content \emph{and} applying relevant clinical knowledge. We incorporate two additional design principles in Steps~2-3. First, we encourage \textbf{cross-turn error chaining} where clinically appropriate distractors should form coherent chains across turns, such that if a model selects an incorrect answer in an earlier question, subsequent incorrect choices include plausible continuations of that earlier error. This allows mistakes to compound across turns, which is a key property of realistic multi-turn QA.
Second, we encourage \textbf{shared-entity choices}: some distractors should reference the same key entities as the correct answer but differ in relationships, attributes, or clinical interpretations applied to them. This prevents the trivial strategy of selecting whichever answer choice mentions entities from the notes, and instead requires evaluating the clinical reasoning tied to those entities.

\textbf{(3) Evidence-grounding-question answer choices (Step~4)}. Each content question is paired with an evidence-grounding question: ``\emph{What are the minimal exact sources from the patient's discharge summaries that fully contain information for your previous answer?}''. Each answer choice for the evidence-grounding question consists of $\langle$note number (\& chart date), section header(s)$\rangle$ tuples drawn from the patient's discharge summaries (example section headers include Chief Complaint, Major Surgical or Invasive Procedure, Brief Hospital Course, Pertinent Results, and Discharge Medications). We prompt Gemini-2.5-Pro to generate the evidence-grounding answer choices as follows.
\textbf{(a)~Correct answer choice (Minimal-yet-sufficient constraint)}: the smallest set of $\langle$note number (\& chart date), section header(s)$\rangle$ tuples that fully supports the answer to the corresponding content question. \emph{Sufficiency} requires that every clinical detail of the answer be grounded in the selected sections, while \emph{minimality} requires removing any section whose supporting information is already contained in another selected section. \textbf{(b)~Incorrect answer choices}: distractors that fail to fully support the answer via one of four error patterns: (i) all headers irrelevant, (ii) headers covering only part of the answer, (iii) a mix of partially relevant and irrelevant headers, or (iv) the full correct set plus irrelevant additions. Irrelevant headers are encouraged to be \emph{semantic traps}: names that plausibly align with the answer's wording but whose contents do not support it, requiring models to read the section content rather than rely on surface header-name matching.

\subsection{Medical Expert Review and Dataset Finalization}\label{sec:expert-review}

To ensure clinical naturalness, correctness, and grounding in the discharge summaries, each of the 1,000 multi-turn QA samples (one per patient) produced by the four-step pipeline was comprehensively reviewed by medical experts.

\textbf{Reviewers and protocol.}
Each sample was reviewed by one of 12 medical experts---11 physicians (10 board-certified specialists and 1 fellow) and 1 nurse---over three weeks, covering 16{,}072 QA pairs. 
Reviewer clinical backgrounds are detailed in Appendix~\ref{app:annotation-backgrounds}.
Reviewers were provided with a Streamlit interface that displayed each sample's multi-turn QA sequence, with one correct and four incorrect choices per question, alongside the patient's discharge summaries.
They were provided a separate spreadsheet to record revisions. 
Full annotation instructions and interface screenshots are in Appendix~\ref{app:annotation}.

\textbf{Review criteria.} Reviewers were instructed to apply the following decisions per sample: 
\textbf{(1) Delete sample} if the overall multi-turn question sequence was clinically unnatural and could not be revised into a realistic clinical multi-turn inquiry, or if the source headers supporting the correct answer had been incorrectly de-identified in MIMIC-IV, making the evidence-grounding question unanswerable; 
\textbf{(2) Revise question} if an individual question was not clinically natural but could be edited into a question that medical experts would ask in clinical practice; 
\textbf{(3) Revise correct answer choice} if its content was partially or fully incorrect against the discharge summaries; 
\textbf{(4) Revise incorrect answer choice} if it was actually correct and therefore invalid as an incorrect answer choice.

\textbf{Outcome.}
After review, 33 of the 1{,}000 samples were removed, yielding 967 final samples. 
Each content question has one correct and four incorrect choices, yielding 8{,}036 correct and 32{,}144 incorrect content answer choices; the same counts apply to evidence-grounding questions.
Among retained content questions, \textbf{125} questions (1.6\% of 8{,}036), \textbf{101} correct choices (1.3\%), and \textbf{156} incorrect choices (0.5\% of 32{,}144) were revised.
For evidence-grounding questions, \textbf{425} correct choices (5.3\% of 8{,}036) and \textbf{1{,}676} incorrect choices (5.2\% of 32{,}144) were revised.
The correct-answer position (A--E) was then shuffled to be approximately uniform across the dataset.
Table~\ref{tab:dataset-stats} reports the final dataset statistics.

\textbf{Multi-note coverage.}
To assess whether the final dataset's multi-turn sequences genuinely draw evidence from multiple discharge summaries, we examine the notes included in the correct answer choices of the evidence-grounding questions.
Of the 767 multi-note samples (patients with at least two discharge summaries), 724 (94.4\%) reference at least two distinct notes.
On average, the notes referenced across each multi-note sample's question sequence account for 85.2\% of the patient's available discharge summaries.
These results show that most multi-note sequences draw on multiple discharge summaries and reference most of each patient's available discharge summaries.

\begin{table}[t]
\centering
\caption{Statistics of the final EHRNote-ChatQA dataset. Each patient-level sample consists of a multi-turn QA sequence over a single patient's entire set of discharge summaries. Each content question is followed by an evidence-grounding question, so the two question counts are equal.}
\label{tab:dataset-stats}
\footnotesize
\setlength{\tabcolsep}{4pt}
\resizebox{\linewidth}{!}{%
\begin{tabular}{l r @{\hspace{1.4em}} l r r c @{\hspace{1.4em}} c r r c}
\toprule
\multicolumn{2}{c}{\textbf{Overall}} &
\multicolumn{4}{c}{\textbf{By question category}} &
\multicolumn{4}{c}{\textbf{By \# discharge summaries}} \\
\cmidrule(r){1-2}
\cmidrule(lr){3-6}
\cmidrule(l){7-10}
Metric & Count &
Category & Samples & \shortstack{Total\\questions} & \shortstack{Total questions\\per sample} &
Notes & Samples & \shortstack{Total\\questions} & \shortstack{Total questions\\per sample} \\
\midrule
Patient-level samples & 967
& Diagnosis & 117 & 1{,}994 & 17.0
& 1 & 200 & 2{,}360 & 11.8 \\

Discharge summaries / patient (mean) & 2.95
& Symptom & 122 & 2{,}176 & 17.8
& 2 & 200 & 3{,}238 & 16.2 \\

Content questions & 8{,}036
& Procedure & 120 & 1{,}994 & 16.6
& 3 & 198 & 3{,}594 & 18.2 \\

Evidence-grounding (Source) questions & 8{,}036
& Medication & 124 & 2{,}094 & 16.9
& 4 & 187 & 3{,}444 & 18.4 \\

Total questions (Content + Source) & 16{,}072
& Microbiology & 121 & 1{,}714 & 14.2
& 5 & 182 & 3{,}436 & 18.9 \\

Min total questions / sample & 8
& Clinical Assessment & 120 & 2{,}120 & 17.7
& & & & \\

Mean total questions / sample & 16.6
& Clinical Outcome & 122 & 2{,}050 & 16.8
& & & & \\

Max total questions / sample & 26
& Discharge Plan & 121 & 1{,}930 & 16.0
& & & & \\

\cmidrule(lr){3-6}
\cmidrule(l){7-10}
&
& \textbf{Total} & \textbf{967} & \textbf{16{,}072} & \textbf{16.6}
& \textbf{Total} & \textbf{967} & \textbf{16{,}072} & \textbf{16.6} \\
\bottomrule
\end{tabular}%
}
\end{table}

\section{Experiments}

\subsection{Experimental Settings}
\paragraph{Models.}\label{sec:models}
We benchmark 24 LLMs on EHRNote-ChatQA, spanning proprietary frontier models (Google Gemini~\cite{comanici2025gemini}, OpenAI GPT~\cite{singh2025openai}), open-weight general-purpose models (Qwen3~\cite{yang2025qwen3}, Llama-4~\cite{adcock2026llama}, DeepSeek-R1-Distill~\cite{guo2025deepseek}, Ministral~\cite{liu2026ministral}, Phi~\cite{abdin2024phi,abdin2024phi3}, Gemma~3~\cite{gemmateam2025gemma3}), and medical-domain specialized models (MedGemma~\cite{sellergren2025medgemma}, MediPhi~\cite{corbeil2025modular}).

\paragraph{Metrics.}
For each model, we report nine accuracy metrics combining three correctness criteria with three aggregation levels.
The correctness criteria are: \emph{Content} for content-question accuracy, \emph{Source} for evidence-grounding accuracy, and \emph{Paired} for accuracy when both the content answer and its paired evidence grounding answer are correct.
The aggregation levels are: \emph{QA-level}, a micro-average over all turns of all samples, where each turn contributes equally; \emph{Sample-level (mean)}, a macro-average of per-sample accuracies, so each sample contributes equally regardless of turn count; and \emph{Sample-level (0/1)}, which counts a sample correct only if every turn is correct.

\begin{table}[t]
\centering
\caption{Main results on EHRNote-ChatQA (accuracy in \%). \emph{Content}: content-question accuracy. \emph{Source}: evidence-grounding accuracy. \emph{Paired}: both the content answer and its paired evidence-grounding answer correct. \emph{QA-level} micro-averages over all turns; \emph{Sample-level (mean)} macro-averages per-sample accuracy; \emph{Sample-level (0/1)} requires every turn in a sample correct. \emph{EHRNoteQA}~\cite{kweon2024ehrnoteqa} is an external single-turn reference. Best per column and model-category in \textbf{bold}.}
\label{tab:main-results}
\setlength{\tabcolsep}{2.5pt}
\footnotesize
\resizebox{\linewidth}{!}{%
\begin{tabular}{l c ccc ccc ccc c}
\toprule
& & \multicolumn{9}{c}{Multi-Turn \& Evidence-Grounding (Ours)} & Single-Turn \\
\cmidrule(lr){3-11} \cmidrule(lr){12-12}
& & \multicolumn{3}{c}{QA-level} & \multicolumn{3}{c}{Sample-level (mean)} & \multicolumn{3}{c}{Sample-level (0/1)} & \\
\cmidrule(lr){3-5} \cmidrule(lr){6-8} \cmidrule(lr){9-11}
Model & \shortstack{Params\\(Active / Total)} & Content & Source & Paired & Content & Source & Paired & Content & Source & Paired & EHRNoteQA \\
\midrule
\multicolumn{12}{l}{\textit{Proprietary frontier models}} \\
gpt-5.4                & --- & \textbf{98.28} & 88.38 & 87.39 & \textbf{98.28} & 88.67 & 87.68 & \textbf{87.49} & 41.78 & 38.47 & \textbf{95.43} \\
gpt-5.4-mini           & --- & 92.36 & 75.68 & 70.69 & 92.62 & 75.95 & 71.10 & 57.39 & 14.17 & 9.93 & 92.00 \\
gemini-3-flash-preview & --- & 97.91 & \textbf{93.22} & \textbf{91.69} & 97.94 & \textbf{93.38} & \textbf{91.89} & 84.90 & \textbf{58.74} & \textbf{52.12} & 92.00 \\
\midrule
\multicolumn{12}{l}{\textit{Open-weight general-purpose models}} \\
Llama-4-Scout-17B-16E-Instruct & 17B / 109B & 92.78 & 78.45 & 73.71 & 93.09 & 78.78 & 74.23 & 59.46 & 19.65 & 13.75 & 93.35 \\
Qwen3-Next-80B-A3B-Instruct    & 3B / 80B   & \textbf{95.87} & \textbf{78.99} & \textbf{76.54} & \textbf{95.91} & \textbf{79.22} & \textbf{76.76} & \textbf{73.84} & \textbf{20.89} & \textbf{17.37} & 93.76 \\
Qwen3-30B-A3B-Instruct-2507    & 3B / 30B   & 92.82 & 70.97 & 66.90 & 93.00 & 70.83 & 66.86 & 57.60 & 9.51 & 6.00 & 94.39 \\
Qwen3-4B-Instruct-2507         & 4B         & 88.84 & 59.13 & 53.57 & 89.22 & 59.52 & 54.08 & 43.85 & 4.55 & 2.48 & 92.31 \\
DeepSeek-R1-Distill-Llama-70B  & 70B        & 94.84 & 75.12 & 72.05 & 94.98 & 75.52 & 72.51 & 67.53 & 14.06 & 11.27 & \textbf{94.80} \\
DeepSeek-R1-Distill-Qwen-32B   & 32B        & 92.02 & 68.37 & 64.04 & 92.12 & 68.54 & 64.24 & 55.33 & 9.00 & 6.51 & 93.87 \\
DeepSeek-R1-Distill-Qwen-14B   & 14B        & 87.42 & 58.31 & 52.48 & 87.69 & 58.39 & 52.63 & 37.13 & 2.79 & 1.96 & 92.62 \\
DeepSeek-R1-Distill-Llama-8B   & 8B         & 74.33 & 48.95 & 38.02 & 74.89 & 49.13 & 38.32 & 13.65 & 1.24 & 0.52 & 89.29 \\
DeepSeek-R1-Distill-Qwen-7B    & 7B         & 43.29 & 31.55 & 15.51 & 45.30 & 33.00 & 17.12 & 2.38 & 0.52 & 0.00 & 78.90 \\
Ministral-3-14B-Instruct-2512  & 14B        & 87.18 & 68.53 & 60.69 & 87.71 & 68.80 & 61.26 & 28.44 & 7.65 & 4.45 & 64.03 \\
Ministral-3-8B-Instruct-2512   & 8B         & 91.09 & 66.03 & 60.86 & 91.30 & 66.05 & 60.97 & 50.57 & 6.72 & 4.14 & 90.75 \\
Ministral-3-3B-Instruct-2512   & 3B         & 83.01 & 44.80 & 37.54 & 83.50 & 45.17 & 38.09 & 29.37 & 0.62 & 0.21 & 84.10 \\
gemma-3-27b-it                 & 27B        & 89.91 & 75.25 & 68.67 & 90.51 & 75.76 & 69.62 & 50.57 & 15.62 & 9.93 & 92.52 \\
gemma-3-4b-it                  & 4B         & 66.69 & 47.66 & 32.19 & 67.06 & 48.25 & 32.78 & 6.93 & 1.96 & 0.31 & 80.87 \\
Phi-4-mini-instruct            & 3.8B       & 71.42 & 41.36 & 30.49 & 72.12 & 41.57 & 30.94 & 11.17 & 0.72 & 0.31 & 85.24 \\
Phi-3.5-mini-instruct          & 3.8B       & 72.06 & 36.97 & 26.59 & 72.80 & 37.24 & 27.07 & 12.00 & 0.62 & 0.10 & 84.30 \\
\midrule
\multicolumn{12}{l}{\textit{Medical-domain specialized models}} \\
medgemma-27b-it  & 27B  & \textbf{88.10} & \textbf{70.71} & \textbf{63.03} & \textbf{88.79} & \textbf{70.93} & \textbf{63.71} & \textbf{43.43} & \textbf{9.82} & \textbf{5.79} & \textbf{91.16} \\
medgemma-4b-it   & 4B   & 62.08 & 49.50 & 31.20 & 62.47 & 49.85 & 31.60 & 4.96 & 1.34 & 0.31 & 80.98 \\
MediPhi-Instruct & 3.8B & 64.92 & 37.10 & 24.02 & 65.65 & 37.45 & 24.46 & 6.72 & 0.93 & 0.21 & 84.41 \\
MediPhi-Clinical & 3.8B & 70.69 & 40.07 & 28.80 & 71.49 & 40.30 & 29.22 & 12.41 & 0.83 & 0.31 & 83.89 \\
MediPhi-PubMed   & 3.8B & 72.13 & 36.90 & 26.68 & 72.87 & 37.20 & 27.16 & 11.48 & 0.41 & 0.00 & 84.41 \\
\bottomrule
\end{tabular}%
}
\end{table}

\paragraph{Evaluation protocol.}

We evaluate models in a multi-turn chat setting, following common protocols such as MT-Bench~\cite{zheng2023judging} and MT-Eval~\cite{kwan2024mt}.
Prior user messages (questions and answer choices) and assistant responses are retained as conversation history using each model's native chat template, and the model is instructed to select one of A/B/C/D/E at each turn.
The first-turn user message contains the patient's entire sequence of discharge summaries, the first question, and its five answer choices; later turns contain only the next question and its answer choices.
We use greedy decoding (temperature $0$) wherever temperature can be set, except the DeepSeek-R1-Distill family, which uses $0.6$ following the official recommendation~\cite{guo2025deepseek}; gpt-5.4 and gpt-5.4-mini, whose temperature cannot be set, use the default API setting.
Implementation details (e.g., inference framework, HIPAA-compliant API access, evaluation cost) are in Appendix~\ref{app:implementation}.

\subsection{Main Results}

Table~\ref{tab:main-results} reports the benchmark results across all 24 models.

\paragraph{Content-question accuracy.}
For QA-level content-question accuracy, proprietary frontier models perform best, with gpt-5.4 and gemini-3-flash-preview achieving 98.28\% and 97.91\%, respectively.
Among open-weight models, the Qwen3 family performs strongly across sizes, with the 80B variant achieving the highest open-weight accuracy (95.87\%) and the 30B and 4B variants reaching 92.82\% and 88.84\%; Llama-4-Scout-17B also performs strongly at 92.78\%.
The DeepSeek-R1-Distill family shows substantial variation across base model and size, with accuracies of 94.84\%, 92.02\%, 87.42\%, 74.33\%, and 43.29\% for the 70B, 32B, 14B, 8B, and 7B variants, respectively.
For the Ministral-3-Instruct family, accuracy does not increase monotonically with model size: the 14B, 8B, and 3B variants achieve 87.18\%, 91.09\%, and 83.01\%, respectively.

Among medical-domain open-weight models, medgemma-27b-it achieves the highest content accuracy at 88.10\%.
However, despite continued medical-domain pretraining and instruction tuning, it underperforms gemma-3-27b-it (89.91\%), the general-purpose instruction-tuned model derived from the same pretrained model, gemma-3-27b-pt.
Similarly, medgemma-4b-it (62.08\%) underperforms not only gemma-3-4b-it (66.69\%), likewise instruction-tuned from the same pretrained model (i.e., gemma-3-4b-pt), but also other general-purpose models of comparable size, such as Qwen3-4B (88.84\%) and Phi-4-mini (71.42\%).
The MediPhi variants likewise show no consistent gain over their base model, Phi-3.5-mini-instruct (72.06\%): MediPhi-Instruct, MediPhi-Clinical, and MediPhi-PubMed achieve 64.92\%, 70.69\%, and 72.13\%, respectively.
These results indicate that medical-domain specialization does not necessarily improve performance on multi-turn clinical QA over patient discharge summaries.

\paragraph{Evidence-grounding accuracy.}
QA-level Source accuracy is strongly correlated with Content accuracy (Spearman $\rho = 0.90$), indicating that models with higher content-question accuracy also tend to perform better on evidence grounding.
Nevertheless, Source accuracy is consistently and substantially lower: across all 24 models, it is 22.0~percentage points (pp) lower than Content accuracy on average.
This gap varies with model capability: frontier models exhibit smaller gaps, such as 9.9~pp for gpt-5.4 and 4.7~pp for gemini-3-flash-preview, whereas smaller open-weight models show much larger gaps, such as 38.2~pp for Ministral-3-3B-Instruct and 35.1~pp for Phi-3.5-mini-instruct.
These results show that models struggle more with evidence-grounding questions than with content questions, particularly smaller open-weight models, underscoring the need to evaluate evidence grounding separately from content accuracy.

\paragraph{Multi-turn consistency evaluation.}
While QA-level results evaluate individual turns, practical clinical use requires models to remain correct across an entire multi-turn interaction.
We therefore examine Sample-level (0/1) accuracy, which counts a sample as correct only if every turn is correct.
Compared with QA-level accuracy, it drops by 44.2, 49.9, and 44.5~pp on average for Content, Source, and Paired accuracy, respectively.
Under the strictest metric, Sample-level (0/1) Paired accuracy, the best proprietary models reach only 52.12\% (gemini-3-flash-preview) and 38.47\% (gpt-5.4), and the best open-weight models only 17.37\% (Qwen3-Next-80B-A3B-Instruct) and 13.75\% (Llama-4-Scout-17B).
These results show that current LLMs still have substantial room for improvement in consistently answering complete multi-turn clinical QA sequences.

\paragraph{Comparison with single-turn discharge-summary QA.}
We compare EHRNote-ChatQA with EHRNoteQA~\cite{kweon2024ehrnoteqa}, an external single-turn clinical QA benchmark containing physician-curated questions over patient discharge summaries.
QA-level Content accuracy is lower on EHRNote-ChatQA for 17 of 24 models, with an average decrease of 5.81~pp.
The drop is concentrated in smaller models: proprietary frontier models and open-weight models with at least 27B total parameters perform comparably on both benchmarks (0.16~pp higher on average), whereas smaller open-weight models drop by 10.07~pp on average.
One exception is Ministral-3-14B-Instruct-2512, which scores 23.15~pp higher on EHRNote-ChatQA than on EHRNoteQA.
This model often returns an empty response to the first question of a conversation, which leaves 310 of 962 single-turn EHRNoteQA questions unanswered but affects only the first turn of each EHRNote-ChatQA sample.
Overall, these results suggest that single-turn clinical QA performance does not reliably transfer to multi-turn clinical QA, underscoring the need for evaluating on multi-turn clinical QA benchmarks.

\subsection{Analysis}\label{sec:analysis}

\paragraph{Content-question inter-turn dependency.}
To test whether errors arise independently at each turn or propagate to later turns, we measure per-turn content accuracy conditioned on the correctness of the immediately preceding content turn (Figure~\ref{fig:dependency-source}(a) in Appendix~\ref{app:analysis}).
For all 24 models, the dependency gap $\Delta = \mathrm{Acc}(Q_t^{\mathrm{cont}} \mid Q_{t-1}^{\mathrm{cont}}\ \mathrm{correct}) - \mathrm{Acc}(Q_t^{\mathrm{cont}} \mid Q_{t-1}^{\mathrm{cont}}\ \mathrm{wrong})$ is positive.
Averaged across models, content-question accuracy is 83.33\% when the previous content turn was answered correctly versus 74.94\% when incorrectly answered, yielding a gap of $\bar{\Delta}=8.40$~pp.
These results show that EHRNote-ChatQA exhibits a critical property of multi-turn multiple-choice QA---multi-turn dependency and cross-turn error chaining---built by design (Section~\ref{sec:generation}).

\paragraph{Effect of note count.}
To examine how performance changes with the number of notes, we measure QA-level accuracy by the number of discharge summaries per patient (1--5; Figure~\ref{fig:dependency-source}(b) in Appendix~\ref{app:analysis}).
Across all three metrics (Content, Source, and Paired), accuracy declines for most models (23 out of 24) as note count increases; averaged across models, accuracy drops by 8.9, 5.5, and 8.7~pp, respectively, from 1 to 5 notes.
These results suggest that models misidentify information more often as the number of notes increases, likely because they must process longer multi-note contexts and integrate information across admissions.

\paragraph{Category-level performance.}
Variation in accuracy across categories is small relative to differences between models (Figure~\ref{fig:category} in Appendix~\ref{app:analysis}): the per-model standard deviation across the eight categories, averaged over all models, is 2.78, 3.02, and 3.17~pp for Content, Source, and Paired accuracy, compared with a standard deviation of 14.02, 18.05, and 22.07~pp in overall accuracy across models.
Consequently, model rankings are nearly unchanged across categories (pairwise Spearman's $\rho \geq 0.96$).
For Content accuracy, higher-accuracy models are more uniform across categories (Spearman's $\rho=-0.89$ between a model's average accuracy and its standard deviation): gpt-5.4 and gemini-3-flash-preview vary by only 0.70 and 0.42~pp, whereas the 14 smaller open-weight models (fewer than 27B total parameters) vary by 3.68~pp on average.

\subsection{Open-Ended Generation Evaluation}\label{sec:openended}

The multiple-choice format enables objective and scalable grading (Section~\ref{sec:generation}), but whether it reflects models' ability to answer the same clinical questions in free text remains an important question.
We therefore evaluate open-ended generation on a random subset of 200 samples (3{,}286 QA pairs).
At each turn, the model generates a free-text answer, which an LLM judge (gpt-5.4) scores against the reference content answer or source set.
Of the 24 models in Table~\ref{tab:main-results}, we exclude gpt-5.4 because it serves as the judge, and evaluate the remaining 23.
Across these models, multiple-choice and open-ended QA-level accuracy are strongly correlated for both content questions (Pearson $r=0.948$, Spearman $\rho=0.879$) and evidence-grounding questions (Pearson $r=0.934$, Spearman $\rho=0.890$).
These results suggest that multiple-choice evaluation is a practical, scalable option for EHRNote-ChatQA, while the benchmark can also be used for open-ended evaluation with an LLM judge.
Full experimental details, the scoring rubric, and per-model results are provided in Appendix~\ref{app:generation-eval}.

\section{Limitations}

EHRNote-ChatQA has several limitations. 
First, the benchmark is constructed from data from a single hospital system (MIMIC-IV), and therefore may not fully capture variation across institutions or EHR systems. 
Second, we focus on discharge summaries, which provide the overall clinical context of each admission, but do not cover other EHR sources such as progress notes, nursing notes, radiology reports, or structured EHR tables. 
Extending evidence-grounded multi-turn QA to these heterogeneous data sources may be an important direction for future work.
Finally, although every QA sample is reviewed and revised by medical experts, each sample is reviewed by a single expert rather than multiple independent annotators. 
This trade-off was necessary given the large scale of the dataset (16{,}072 QA pairs) and the limited availability of medical experts.

\section{Conclusion}

We introduced \textbf{EHRNote-ChatQA}, the first evidence-grounded, multi-turn, multi-note clinical QA benchmark over patient discharge summaries.
The benchmark contains 967 patient-level multi-turn samples and 16{,}072 medical-expert-verified QA pairs across eight clinical categories, constructed through a medical-expert-informed pipeline and comprehensive expert review.

Benchmarking 24 LLMs shows that current models still struggle more with evidence grounding than content answering, that errors compound across turns, and that single-turn clinical QA performance does not reliably transfer to this setting.
These results highlight the need for benchmarks that evaluate multi-turn QA over longitudinal patient contexts with verifiable evidence grounding, and EHRNote-ChatQA provides a rigorous evaluation benchmark for evaluating clinical QA systems.

\begin{ack}
This work was supported by the Institute of Information \& Communications Technology Planning \& Evaluation (IITP) grants (No.RS-2019-II190075, No.RS-2025-02304967), the Korea Health Industry Development Institute (KHIDI) grant (No.RS-2025-02213750), and the National Research Foundation of Korea (NRF) grant (NRF-2020H1D3A2A03100945, RS-2026-25484088), and the Medical Scientist Training Program funded by the Korean government (MSIT, MOHW).
\end{ack}

\clearpage

{\small
\bibliographystyle{plainnat}
\bibliography{references}
}

\clearpage
\appendix

\etocdepthtag.toc{appendix}
\begingroup
\etocsettagdepth{mainmatter}{none}
\etocsettagdepth{appendix}{subsection}
\etocsettocstyle{\section*{Supplementary Contents}}{}
\tableofcontents
\endgroup
\clearpage

\section{EHRNote-ChatQA Dataset Samples}\label{app:example}

We provide three EHRNote-ChatQA samples below, spanning different question categories (Figures~\ref{fig:example-466}, \ref{fig:example-854}, and~\ref{fig:example-572}). 
Each turn is a 5-way multiple-choice question in chat format, with the correct option shown in \textbf{bold}. 
Odd-numbered turns are content questions; each is immediately followed by an evidence-grounding (source-location) question (even-numbered) that asks for the minimal set of notes and section headers supporting the preceding content answer. 
The text is reproduced from EHRNote-ChatQA; the only modification is that each note's chartdate is replaced by a placeholder ([chartdate$N$] for Note \#$N$) to avoid disclosing patient-identifying dates. In the released dataset (under PhysioNet credentialed access) the questions and answer choices contain the actual chartdates.

\begin{tcolorbox}[
  enhanced,
  breakable,
  colback=gray!3,
  colframe=gray!55,
  boxrule=0.6pt,
  arc=2mm,
  left=10pt, right=10pt, top=10pt, bottom=10pt,
  before upper={\setlength{\parskip}{10pt}\setlength{\parindent}{0pt}},
]
\footnotesize
\textbf{Category: Diagnosis}

\textbf{Q1.} During her first admission in [chartdate1], what was the primary diagnosis, and what underlying conditions were identified?\\
A. The primary diagnosis was a right thalamic hemorrhage, which was found to be caused by Moyamoya syndrome and a long-segment dissection of the right internal carotid artery.\\
B. The primary diagnosis was acute left-sided weakness, which workup revealed was caused by a right thalamic hemorrhage secondary to underlying Moyamoya syndrome.\\
\textbf{C. The primary diagnosis was a right thalamic hemorrhage, attributed to underlying Moyamoya syndrome and two small aneurysms found on angiography.}\\
D. The patient's primary diagnosis was a right thalamic hemorrhage, attributed to a hypertensive emergency from previously undiagnosed and uncontrolled chronic hypertension.\\
E. The patient was admitted for an evaluation of known Moyamoya disease, which was found to have caused a secondary right thalamic hemorrhage and the formation of two small aneurysms.\\
\textit{Correct answer: C.}

\textbf{Q2.} What are the minimal exact sources from the patient's discharge summaries that fully contain information for your previous answer?\\
A. Note \#2 Chartdate: [chartdate2] Headers: Discharge Diagnosis\\
B. Note \#1 Chartdate: [chartdate1] Headers: Past Medical History, Social History\\
C. Note \#1 Chartdate: [chartdate1] Headers: History of Present Illness\\
D. Note \#1 Chartdate: [chartdate1] Headers: Discharge Diagnosis\\
\textbf{E. Note \#1 Chartdate: [chartdate1] Headers: Brief Hospital Course}\\
\textit{Correct answer: E.}

\textbf{Q3.} Did the patient have any subsequent hospitalizations for a similar problem?\\
A. Yes, approximately nine years later, in [chartdate3], she presented with a week of headache and nausea and was diagnosed with a new left occipital intracerebral hemorrhage.\\
B. Yes, in [chartdate2], she was readmitted with severe headache and fever and was diagnosed with EVD-associated meningitis after having ventricular drains placed.\\
\textbf{C. Yes, approximately three years later, in [chartdate2], she was admitted with a severe headache and vomiting and was diagnosed with an intraventricular hemorrhage.}\\
D. No, her underlying vascular issues were stable after the initial event. Her next admission was a planned hospitalization for an elective revascularization procedure to treat her Moyamoya disease.\\
E. Yes, she was readmitted approximately two years later for an acute ischemic stroke, which presented with new-onset aphasia and right-sided weakness, as had been a concern during her first admission.\\
\textit{Correct answer: C.}

\textbf{Q4.} What are the minimal exact sources from the patient's discharge summaries that fully contain information for your previous answer?\\
A. Note \#2 Chartdate: [chartdate2] Headers: Discharge Diagnosis\\
\textbf{B. Note \#2 Chartdate: [chartdate2] Headers: History of Present Illness}\\
C. Note \#2 Chartdate: [chartdate2] Headers: Chief Complaint\\
D. Note \#2 Chartdate: [chartdate2] Headers:Family History\\
E. Note \#1 Chartdate: [chartdate1] Headers: History of Present Illness\\
\textit{Correct answer: B.}

\textbf{Q5.} Were there any complications during that admission?\\
\textbf{A. Yes, her hospitalization was complicated by obstructive hydrocephalus requiring bilateral EVDs, an Enterococcus UTI, and EVD-associated meningitis.}\\
B. Yes, the ischemic stroke was complicated by hemorrhagic conversion, which led to brain swelling and required an emergent decompressive craniectomy.\\
C. Yes, the main complications were metabolic, including severe hyponatremia down to 129 and delirium, which was attributed to sleep deprivation in the ICU.\\
D. Yes, the admission was complicated by obstructive hydrocephalus requiring EVDs, and she also developed hospital-acquired pneumonia that prolonged her ICU stay.\\
E. Yes, during that admission she developed severe depression and anxiety with suicidal ideation, which required a psychiatry consultation and initiation of mirtazapine.\\
\textit{Correct answer: A.}

\textbf{Q6.} What are the minimal exact sources from the patient's discharge summaries that fully contain information for your previous answer?\\
A. Note \#1 Chartdate: [chartdate1] Headers: Brief Hospital Course\\
B. Note \#2 Chartdate: [chartdate2] Headers: Major Surgical or Invasive Procedure\\
C. Note \#2 Chartdate: [chartdate2] Headers: Discharge Diagnosis, Major Surgical or Invasive Procedure\\
\textbf{D. Note \#2 Chartdate: [chartdate2] Headers: Brief Hospital Course}\\
E. Note \#2 Chartdate: [chartdate2] Headers: Discharge Diagnosis\\
\textit{Correct answer: D.}

\textbf{Q7.} What organisms were found in relation to the infections?\\
A. The UTI was caused by vancomycin-sensitive Enterococcus, but the meningitis was caused by a heavy growth of methicillin-resistant Staphylococcus aureus (MRSA).\\
B. The UTI was caused by E. coli, a common pathogen, while the meningitis was caused by both Staphylococcus epidermidis and a fungal infection with Candida.\\
C. The urinary tract infection was caused by Staphylococcus epidermidis, and the meningitis was caused by Enterococcus sp.\\
\textbf{D. The UTI was caused by Enterococcus sp. and the meningitis was polymicrobial, involving Staphylococcus epidermidis, Corynebacterium, and Enterococcus.}\\
E. The urinary tract infection was caused by Enterococcus sp.; subsequent CSF cultures showed no growth and were considered contaminants after initial concerns.\\
\textit{Correct answer: D.}

\textbf{Q8.} What are the minimal exact sources from the patient's discharge summaries that fully contain information for your previous answer?\\
A. Note \#2 Chartdate: [chartdate2] Headers: Discharge Diagnosis\\
B. Note \#2 Chartdate: [chartdate2] Headers: Brief Hospital Course\\
C. Note \#2 Chartdate: [chartdate2] Headers: Physical Exam, Discharge Instructions\\
\textbf{D. Note \#2 Chartdate: [chartdate2] Headers: Pertinent Results}\\
E. Note \#1 Chartdate: [chartdate1] Headers: Pertinent Results\\
\textit{Correct answer: D.}

\textbf{Q9.} How did the management of the central nervous system infection evolve?\\
A. Given the diagnosis of enterococcal meningitis, the patient was treated with a synergistic combination of high-dose Ampicillin and Gentamicin after the EVD was replaced.\\
B. The patient was empirically treated with Vancomycin and Meropenem. This was narrowed to Vancomycin based on culture results, but was later switched to Linezolid after follow-up cultures showed the development of vancomycin resistance.\\
C. In addition to systemic IV Vancomycin, the patient also received intraventricular Vancomycin administered directly into the EVD to achieve higher drug concentrations at the site of infection.\\
D. The patient was started on Linezolid for meningitis. When she continued to spike fevers, her antibiotics were broadened to Vancomycin and Meropenem, and her EVD was replaced for source control.\\
\textbf{E. Initially, she received empiric Vancomycin and Meropenem, and her EVD was replaced. This was narrowed to Vancomycin per cultures, then switched to Linezolid for suspected drug fever.}\\
\textit{Correct answer: E.}

\textbf{Q10.} What are the minimal exact sources from the patient's discharge summaries that fully contain information for your previous answer?\\
A. Note \#2 Chartdate: [chartdate2] Headers: Discharge Diagnosis, Past Medical History\\
\textbf{B. Note \#2 Chartdate: [chartdate2] Headers: Brief Hospital Course}\\
C. Note \#2 Chartdate: [chartdate2] Headers: Pertinent Results\\
D. Note \#2 Chartdate: [chartdate2] Headers: Discharge Medications\\
E. Note \#3 Chartdate: [chartdate3] Headers: Brief Hospital Course\\
\textit{Correct answer: B.}

\textbf{Q11.} What was the patient's clinical course after that hospitalization?\\
A. Approximately five and a half years later, in [chartdate3], she was admitted again after a week of headache and was found to have an ischemic stroke in the left occipital lobe.\\
B. Her condition remained stable for several years. Her next known medical event was a scheduled admission for the right-sided EDAS bypass surgery that had been planned after her first hemorrhage.\\
\textbf{C. About five and a half years later, in [chartdate3], she was readmitted for a new left occipital intracerebral hemorrhage with intraventricular extension.}\\
D. After recovering from meningitis, she was lost to follow-up until she presented with another severe headache and was found to have a ruptured aneurysm, which was then successfully treated with endovascular coiling.\\
E. Her next hospitalization was for management of severe depression with suicidal ideation, which had become her primary medical issue after her prolonged and complicated recovery.\\
\textit{Correct answer: C.}

\textbf{Q12.} What are the minimal exact sources from the patient's discharge summaries that fully contain information for your previous answer?\\
A. Note \#1 Chartdate: [chartdate1] Headers: Discharge Diagnosis\\
B. Note \#3 Chartdate: [chartdate3] Headers: Discharge Diagnosis\\
\textbf{C. Note \#3 Chartdate: [chartdate3] Headers: History of Present Illness}\\
D. Note \#3 Chartdate: [chartdate3] Headers: Physical Exam, Family History\\
E. Note \#2 Chartdate: [chartdate2] Headers: History of Present Illness\\
\textit{Correct answer: C.}

\textbf{Q13.} What did the evaluation of her underlying disease show during that latest admission?\\
A. An MRI and MRA of the brain were performed, which showed the new hemorrhage and confirmed stability of the previously diagnosed Moyamoya disease without any new vascular changes.\\
B. A conventional angiogram was performed which identified two 3mm aneurysms in the basal ganglia, which were felt to be the cause of the hemorrhage.\\
C. A diagnostic angiogram re-demonstrated her bilateral Moyamoya disease and also revealed a failure of her previous right-sided EDAS bypass graft, which likely caused the new hemorrhage.\\
\textbf{D. A diagnostic angiogram was performed which re-demonstrated her bilateral Moyamoya disease.}\\
E. A diagnostic angiogram was performed which showed progression of her Moyamoya disease, with two new aneurysms having formed near the site of the new hemorrhage.\\
\textit{Correct answer: D.}

\textbf{Q14.} What are the minimal exact sources from the patient's discharge summaries that fully contain information for your previous answer?\\
A. Note \#3 Chartdate: [chartdate3] Headers: Discharge Medications, Discharge Instructions\\
B. Note \#1 Chartdate: [chartdate1] Headers: Pertinent Results\\
C. Note \#3 Chartdate: [chartdate3] Headers: Major Surgical or Invasive Procedure\\
D. Note \#1 Chartdate: [chartdate1] Headers: Brief Hospital Course\\
\textbf{E. Note \#3 Chartdate: [chartdate3] Headers: Brief Hospital Course}\\
\textit{Correct answer: E.}

\textbf{Q15.} Was a new long-term medication started to manage that condition?\\
A. No, given her recent hemorrhage, the decision was made to hold all new antithrombotic or anticoagulant medications to minimize the risk of re-bleeding.\\
\textbf{B. Yes, the neurology service cleared her to start aspirin 81 mg daily.}\\
C. Yes, she was started on Mirtazapine 7.5 mg at bedtime for long-term management.\\
D. Yes, she was started on the antiplatelet agent Clopidogrel 75 mg daily instead of aspirin.\\
E. Yes, she was started on Linezolid to be taken orally for an extended course as an outpatient.\\
\textit{Correct answer: B.}

\textbf{Q16.} What are the minimal exact sources from the patient's discharge summaries that fully contain information for your previous answer?\\
A. Note \#3 Chartdate: [chartdate3] Headers: Discharge Instructions\\
B. Note \#3 Chartdate: [chartdate3] Headers: Discharge Medications\\
\textbf{C. Note \#3 Chartdate: [chartdate3] Headers: Brief Hospital Course}\\
D. Note \#1 Chartdate: [chartdate1] Headers: Discharge Medications\\
E. Note \#3 Chartdate: [chartdate3] Headers: Past Medical History, Physical Exam\\
\textit{Correct answer: C.}

\textbf{Q17.} What was the likely clinical reasoning for initiating this new treatment?\\
A. The aspirin was started to maintain the patency of her right-sided EDAS bypass graft, which is standard practice after revascularization surgery.\\
\textbf{B. The goal of antiplatelet therapy in Moyamoya disease is to reduce the risk of future ischemic strokes.}\\
C. The Mirtazapine was initiated to help manage the patient's depression, anxiety, poor sleep, and nausea, as recommended by the psychiatry service.\\
D. The aspirin was started to prevent another hemorrhagic stroke, as its anti-inflammatory effects are thought to stabilize the fragile Moyamoya vessels.\\
E. The likely reason was for long-term prevention of deep vein thrombosis (DVT), as an oral alternative to the heparin the patient had received in her prior admission.\\
\textit{Correct answer: B.}

\textbf{Q18.} What are the minimal exact sources from the patient's discharge summaries that fully contain information for your previous answer?\\
A. Note \#3 Chartdate: [chartdate3] Headers: Discharge Instructions\\
B. Note \#1 Chartdate: [chartdate1] Headers: Brief Hospital Course\\
C. Note \#3 Chartdate: [chartdate3] Headers: Brief Hospital Course\\
D. Note \#2 Chartdate: [chartdate2] Headers: Brief Hospital Course\\
\textbf{E. Note \#1 Chartdate: [chartdate1] Headers: Brief Hospital Course, Note \#3 Chartdate: [chartdate3] Headers: Brief Hospital Course}\\
\textit{Correct answer: E.}
\end{tcolorbox}
\noindent\begin{minipage}{\linewidth}
\captionsetup{type=figure,hypcap=false}
\captionof{figure}{Example multi-turn QA sample from the Diagnosis category.}
\label{fig:example-466}
\end{minipage}

\clearpage

\begin{tcolorbox}[
  enhanced,
  breakable,
  colback=gray!3,
  colframe=gray!55,
  boxrule=0.6pt,
  arc=2mm,
  left=10pt, right=10pt, top=10pt, bottom=10pt,
  before upper={\setlength{\parskip}{10pt}\setlength{\parindent}{0pt}},
]
\footnotesize
\textbf{Category: Symptom}

\textbf{Q1.} What were the main presenting symptoms for the admission on [chartdate1]?\\
\textbf{A. Abdominal distention and hyponatremia.}\\
B. A clogged nasojejunal feeding tube that was placed during a prior admission.\\
C. Worsening abdominal distention and severe malnutrition.\\
D. Confusion and somnolence consistent with an episode of hepatic encephalopathy.\\
E. Abdominal distention and hyperkalemia.\\
\textit{Correct answer: A.}

\textbf{Q2.} What are the minimal exact sources from the patient's discharge summaries that fully contain information for your previous answer?\\
A. Note \#1 Chartdate: [chartdate1] Headers: Physical Exam\\
B. Note \#1 Chartdate: [chartdate1] Headers: Pertinent Results\\
\textbf{C. Note \#1 Chartdate: [chartdate1] Headers: Chief Complaint}\\
D. Note \#1 Chartdate: [chartdate1] Headers: History of Present Illness, Discharge Diagnosis\\
E. Note \#1 Chartdate: [chartdate1] Headers: Major Surgical or Invasive Procedure, Discharge Condition\\
\textit{Correct answer: C.}

\textbf{Q3.} What was identified as the cause of those symptoms?\\
A. Severe tricuspid regurgitation leading to congestive hepatopathy, based on initial TTE findings.\\
B. Hepatorenal syndrome, which caused both fluid retention and impaired potassium excretion leading to hyperkalemia.\\
C. The patient's severe hyponatremia was identified as the primary cause of the diuretic-refractory ascites.\\
\textbf{D. Decompensated alcoholic cirrhosis resulting in diuretic-refractory ascites.}\\
E. Mechanical obstruction of the nasojejunal feeding tube due to inadequate flushing and crushed medications.\\
\textit{Correct answer: D.}

\textbf{Q4.} What are the minimal exact sources from the patient's discharge summaries that fully contain information for your previous answer?\\
A. Note \#1 Chartdate: [chartdate1] Headers: Discharge Diagnosis, Discharge Instructions\\
B. Note \#1 Chartdate: [chartdate1] Headers: History of Present Illness\\
C. Note \#1 Chartdate: [chartdate1] Headers: Past Medical History, Major Surgical or Invasive Procedure\\
D. Note \#1 Chartdate: [chartdate1] Headers: Discharge Diagnosis\\
\textbf{E. Note \#1 Chartdate: [chartdate1] Headers: Brief Hospital Course}\\
\textit{Correct answer: E.}

\textbf{Q5.} How was the primary symptom managed during that admission?\\
A. As the ascites was diuretic-refractory, the patient was evaluated for and scheduled for a transjugular intrahepatic portosystemic shunt (TIPS) procedure.\\
B. The severe malnutrition was managed via the endoscopic placement of a Dobhoff feeding tube to provide enteral nutrition.\\
C. The patient was managed with aggressive intravenous diuretics and a cardiology consultation to evaluate for tricuspid valve surgery.\\
\textbf{D. With a large-volume paracentesis that removed 9.75 L of ascitic fluid, as diuretics could not be used.}\\
E. With a large-volume paracentesis that removed 6.0 L of ascitic fluid, based on the volume removed in a recent ED visit.\\
\textit{Correct answer: D.}

\textbf{Q6.} What are the minimal exact sources from the patient's discharge summaries that fully contain information for your previous answer?\\
A. Note \#1 Chartdate: [chartdate1] Headers: Brief Hospital Course\\
\textbf{B. Note \#1 Chartdate: [chartdate1] Headers: Brief Hospital Course, Pertinent Results}\\
C. Note \#1 Chartdate: [chartdate1] Headers: Major Surgical or Invasive Procedure\\
D. Note \#1 Chartdate: [chartdate1] Headers: Major Surgical or Invasive Procedure, Discharge Instructions\\
E. Note \#1 Chartdate: [chartdate1] Headers: Pertinent Results\\
\textit{Correct answer: B.}

\textbf{Q7.} In the following admissions, is there evidence that this symptom burden continued?\\
A. Yes, his hyponatremia persisted, with admission sodium levels of 121 mEq/L documented during both subsequent hospitalizations in [chartdate2]/[chartdate3].\\
B. Yes, the symptom burden of malnutrition continued, as he was readmitted twice in [chartdate2]/[chartdate3] for a clogged Dobhoff tube.\\
C. No, his ascites was successfully managed as an outpatient, and his subsequent admissions in [chartdate2]/[chartdate3] were for the unrelated problem of a clogged feeding tube.\\
D. Yes, he required an 8 L therapeutic paracentesis during his admission on [chartdate2] and a 9 L paracentesis on [chartdate3].\\
\textbf{E. Yes, he required a 9 L therapeutic paracentesis during his admission on [chartdate2] and an 8 L paracentesis on [chartdate3].}\\
\textit{Correct answer: E.}

\textbf{Q8.} What are the minimal exact sources from the patient's discharge summaries that fully contain information for your previous answer?\\
A. Note \#1 Chartdate: [chartdate1] Headers: Brief Hospital Course, Note \#2 Chartdate: [chartdate2] Headers: Brief Hospital Course\\
\textbf{B. Note \#2 Chartdate: [chartdate2] Headers: Brief Hospital Course, Note \#3 Chartdate: [chartdate3] Headers: Brief Hospital Course}\\
C. Note \#2 Chartdate: [chartdate2] Headers: Major Surgical or Invasive Procedure, Note \#3 Chartdate: [chartdate3] Headers: Major Surgical or Invasive Procedure\\
D. Note \#2 Chartdate: [chartdate2] Headers: Brief Hospital Course\\
E. Note \#3 Chartdate: [chartdate3] Headers: Pertinent Results\\
\textit{Correct answer: B.}

\textbf{Q9.} What was the documented reasoning for the persistence of this symptom and the need for frequent interventions?\\
\textbf{A. The patient's ascites was documented as being diuretic-refractory because diuretic use was limited by persistent severe hyponatremia.}\\
B. The reasoning was the patient's severe malnutrition, which could not be supported by oral intake alone, necessitating a feeding tube that was prone to clogging.\\
C. The notes suggest the development of hepatorenal syndrome (HRS), which made his kidneys unable to excrete sodium and water, rendering diuretics ineffective.\\
D. The documented reason was progressive right-sided heart failure due to severe tricuspid regurgitation, which was refractory to medical management.\\
E. The recurrent clogging of the feeding tube was attributed to the formula, so it was changed to Osmolite 1.5, and a stricter flushing schedule was ordered.\\
\textit{Correct answer: A.}

\textbf{Q10.} What are the minimal exact sources from the patient's discharge summaries that fully contain information for your previous answer?\\
\textbf{A. Note \#1 Chartdate: [chartdate1] Headers: Brief Hospital Course}\\
B. Note \#1 Chartdate: [chartdate1] Headers: Major Surgical or Invasive Procedure\\
C. Note \#2 Chartdate: [chartdate2] Headers: History of Present Illness\\
D. Note \#1 Chartdate: [chartdate1] Headers: Discharge Diagnosis\\
E. Note \#1 Chartdate: [chartdate1] Headers: Medications on Admission, Pertinent Results\\
\textit{Correct answer: A.}

\textbf{Q11.} Was a different type of intervention performed in a later admission to manage this persistent symptom?\\
\textbf{A. Yes, during the admission on [chartdate4], the patient underwent a planned transjugular intrahepatic portosystemic shunt (TIPS) procedure.}\\
B. Yes, the Dobhoff tube was replaced and advanced post-pyloric by interventional radiology in an attempt to prevent future clogging.\\
C. Yes, a permanent indwelling peritoneal catheter was placed to allow for intermittent drainage of ascitic fluid at home by the patient's caregiver.\\
D. Yes, in a later admission, he underwent a right heart catheterization to obtain definitive pressures and was referred for consideration of tricuspid valve surgery.\\
E. Yes, he underwent a TIPS revision with balloon dilatation of a previously placed stent during the admission on [chartdate4].\\
\textit{Correct answer: A.}

\textbf{Q12.} What are the minimal exact sources from the patient's discharge summaries that fully contain information for your previous answer?\\
A. Note \#4 Chartdate: [chartdate4] Headers: Discharge Diagnosis, Brief Hospital Course\\
B. Note \#3 Chartdate: [chartdate3] Headers: Brief Hospital Course\\
C. Note \#4 Chartdate: [chartdate4] Headers: Major Surgical or Invasive Procedure\\
\textbf{D. Note \#4 Chartdate: [chartdate4] Headers: Chief Complaint, Major Surgical or Invasive Procedure}\\
E. Note \#4 Chartdate: [chartdate4] Headers: History of Present Illness\\
\textit{Correct answer: D.}

\textbf{Q13.} What was the initial outcome of that intervention?\\
A. The right heart catheterization revealed a normal gradient across the tricuspid valve, definitively ruling out significant tricuspid regurgitation.\\
B. The immediate outcome was the development of worsening hepatic encephalopathy, with confusion and asterixis, requiring an increase in his lactulose dose.\\
C. The initial outcome was a reduction in paracentesis volume from 8 liters to 4 liters, but with an ongoing need for weekly procedures.\\
D. The procedure was successful, with a decrease in the right atrial pressure from 28 mmHg to 23 mmHg, although 8 L of ascites were still drained.\\
\textbf{E. The procedure was successful, with the portosystemic gradient decreasing from 10 to 5 mmHg and 8 L of ascites drained.}\\
\textit{Correct answer: E.}

\textbf{Q14.} What are the minimal exact sources from the patient's discharge summaries that fully contain information for your previous answer?\\
A. Note \#4 Chartdate: [chartdate4] Headers: Discharge Condition, Discharge Diagnosis\\
\textbf{B. Note \#4 Chartdate: [chartdate4] Headers: Pertinent Results}\\
C. Note \#4 Chartdate: [chartdate4] Headers: Discharge Diagnosis\\
D. Note \#4 Chartdate: [chartdate4] Headers: Major Surgical or Invasive Procedure\\
E. Note \#4 Chartdate: [chartdate4] Headers: History of Present Illness, Discharge Instructions\\
\textit{Correct answer: B.}

\textbf{Q15.} Did that intervention resolve the need for paracentesis?\\
A. No, and the development of severe post-TIPS encephalopathy made him a poor candidate for further interventions, so weekly 8 L paracentesis continued.\\
B. It is not documented if the intervention for urinary retention was fully successful long-term, though a single dose of Flomax did resolve the acute episode.\\
C. Yes, the TIPS procedure was fully successful and completely resolved his ascites, with no further fluid reaccumulation noted in subsequent follow-up.\\
\textbf{D. No, it only partially resolved the need; paracentesis volume was reduced from 8 L to 4 L, but he still required weekly procedures.}\\
E. No, the need for paracentesis continued, and in fact the required volume of fluid removal increased from 4 liters to 8 liters weekly.\\
\textit{Correct answer: D.}

\textbf{Q16.} What are the minimal exact sources from the patient's discharge summaries that fully contain information for your previous answer?\\
A. Note \#5 Chartdate: [chartdate5] Headers: Major Surgical or Invasive Procedure, Discharge Diagnosis\\
B. Note \#4 Chartdate: [chartdate4] Headers: Brief Hospital Course\\
C. Note \#5 Chartdate: [chartdate5] Headers: Brief Hospital Course\\
\textbf{D. Note \#5 Chartdate: [chartdate5] Headers: History of Present Illness}\\
E. Note \#4 Chartdate: [chartdate4] Headers: Pertinent Results\\
\textit{Correct answer: D.}

\textbf{Q17.} Because the symptom was not fully resolved, what further intervention was performed?\\
A. The patient was referred for expedited liver transplant evaluation as the definitive treatment for his end-stage liver disease and refractory ascites.\\
B. The patient underwent an emergency revision of the thrombosed TIPS shunt, during which the clot was removed and the stent was dilated to restore flow.\\
\textbf{C. The patient underwent placement of a second, parallel transjugular intrahepatic portosystemic shunt (TIPS).}\\
D. No further interventions for ascites were needed; the next procedure was a planned balloon dilatation of the existing, well-functioning TIPS stent as routine maintenance.\\
E. The patient was started on high-dose diuretic therapy with spironolactone 200 mg and furosemide 40 mg daily to augment the effect of the initial TIPS.\\
\textit{Correct answer: C.}

\textbf{Q18.} What are the minimal exact sources from the patient's discharge summaries that fully contain information for your previous answer?\\
A. Note \#5 Chartdate: [chartdate5] Headers: Major Surgical or Invasive Procedure\\
B. Note \#4 Chartdate: [chartdate4] Headers: Major Surgical or Invasive Procedure\\
C. Note \#5 Chartdate: [chartdate5] Headers: Major Surgical or Invasive Procedure, Discharge Instructions\\
\textbf{D. Note \#5 Chartdate: [chartdate5] Headers: History of Present Illness}\\
E. Note \#5 Chartdate: [chartdate5] Headers: Discharge Instructions\\
\textit{Correct answer: D.}

\textbf{Q19.} Did the patient develop any new symptoms immediately following the first of those major procedures?\\
A. No, following the parallel TIPS procedure in the final admission, the patient reported feeling well with no new pain or other complaints documented.\\
B. Yes, he developed severe hypokalemia with a potassium of 2.4 due to fluid shifts after the procedure, requiring aggressive intravenous potassium repletion.\\
C. Yes, immediately after starting high-dose diuretics, he developed worsening kidney function and hyperkalemia, requiring the diuretics to be held.\\
\textbf{D. Yes, he developed post-procedural urinary retention that required multiple straight catheterizations.}\\
E. Yes, after the procedure he developed significant confusion and asterixis, consistent with worsening hepatic encephalopathy that required increasing his lactulose.\\
\textit{Correct answer: D.}

\textbf{Q20.} What are the minimal exact sources from the patient's discharge summaries that fully contain information for your previous answer?\\
\textbf{A. Note \#4 Chartdate: [chartdate4] Headers: Brief Hospital Course}\\
B. Note \#4 Chartdate: [chartdate4] Headers: Discharge Diagnosis\\
C. Note \#5 Chartdate: [chartdate5] Headers: Brief Hospital Course\\
D. Note \#4 Chartdate: [chartdate4] Headers: Pertinent Results\\
E. Note \#4 Chartdate: [chartdate4] Headers: Transitional Issues\\
\textit{Correct answer: A.}
\end{tcolorbox}
\noindent\begin{minipage}{\linewidth}
\captionsetup{type=figure,hypcap=false}
\captionof{figure}{Example multi-turn QA sample from the Symptom category.}
\label{fig:example-854}
\end{minipage}

\clearpage

\begin{tcolorbox}[
  enhanced,
  breakable,
  colback=gray!3,
  colframe=gray!55,
  boxrule=0.6pt,
  arc=2mm,
  left=10pt, right=10pt, top=10pt, bottom=10pt,
  before upper={\setlength{\parskip}{10pt}\setlength{\parindent}{0pt}},
]
\footnotesize
\textbf{Category: Procedure}

\textbf{Q1.} Was any major surgical procedure performed for the patient's left knee pain during the admission on [chartdate2]?\\
A. Yes, the patient underwent a left unicompartmental knee arthroplasty after failing conservative treatments.\\
B. No, a major procedure was not performed; the admission was for medical management of her knee pain, which included the use of a patient-controlled analgesia (PCA) pump.\\
\textbf{C. Yes, a left total knee arthroplasty was performed for management of her left knee pain.}\\
D. Yes, a right total knee arthroplasty was performed for management of her knee pain.\\
E. Yes, a left total knee arthroplasty was performed to repair a pathologic fracture of the knee that occurred as a result of her underlying osteoporosis.\\
\textit{Correct answer: C.}

\textbf{Q2.} What are the minimal exact sources from the patient's discharge summaries that fully contain information for your previous answer?\\
A. Note \#2 Chartdate: [chartdate2] Headers: Social History, Family History\\
B. Note \#2 Chartdate: [chartdate2] Headers: Major Surgical or Invasive Procedure\\
C. Note \#2 Chartdate: [chartdate2] Headers: Chief Complaint, Major Surgical or Invasive Procedure, Discharge Condition\\
D. Note \#1 Chartdate: [chartdate1] Headers: Major Surgical or Invasive Procedure\\
\textbf{E. Note \#2 Chartdate: [chartdate2] Headers: Chief Complaint, Major Surgical or Invasive Procedure}\\
\textit{Correct answer: E.}

\textbf{Q3.} What was the underlying reason for that operation?\\
A. The procedure was indicated for severe rheumatoid arthritis, which had destroyed the left knee joint and was unresponsive to medical management.\\
B. The operation was necessary to treat a pathologic fracture of the left knee caused by the patient's severe osteoporosis.\\
C. The operation was an arthroscopic debridement for a meniscal tear and to remove loose cartilage causing mechanical symptoms in the left knee.\\
D. The operation was a lumpectomy performed for Stage I infiltrating ductal carcinoma of the right breast.\\
\textbf{E. The procedure was performed for osteoarthritis of the left knee following the failure of conservative treatments.}\\
\textit{Correct answer: E.}

\textbf{Q4.} What are the minimal exact sources from the patient's discharge summaries that fully contain information for your previous answer?\\
A. Note \#1 Chartdate: [chartdate1] Headers: History of Present Illness\\
\textbf{B. Note \#2 Chartdate: [chartdate2] Headers: History of Present Illness}\\
C. Note \#2 Chartdate: [chartdate2] Headers: Medications on Admission, Social History\\
D. Note \#2 Chartdate: [chartdate2] Headers: Chief Complaint\\
E. Note \#2 Chartdate: [chartdate2] Headers: Discharge Diagnosis\\
\textit{Correct answer: B.}

\textbf{Q5.} What prophylactic treatments were administered after the procedure?\\
\textbf{A. She received perioperative IV antibiotics, and Lovenox for DVT prophylaxis for three weeks, to be followed by aspirin for an additional three weeks.}\\
B. She received Lovenox for DVT prophylaxis and was instructed to continue her home regimen of Fosamax and calcium to promote healing of the osteoporotic fracture.\\
C. She received perioperative IV antibiotics, and aspirin for DVT prophylaxis for three weeks, to be followed by Lovenox for an additional three weeks.\\
D. For DVT prophylaxis, she was started on the oral anticoagulant rivaroxaban 10 mg daily for three weeks and also received perioperative IV antibiotics.\\
E. She was given perioperative IV antibiotics and started on warfarin for DVT prophylaxis, with instructions to monitor her INR until it was therapeutic.\\
\textit{Correct answer: A.}

\textbf{Q6.} What are the minimal exact sources from the patient's discharge summaries that fully contain information for your previous answer?\\
A. Note \#2 Chartdate: [chartdate2] Headers: Past Medical History, Physical Exam\\
B. Note \#2 Chartdate: [chartdate2] Headers: Discharge Medications\\
C. Note \#1 Chartdate: [chartdate1] Headers: Discharge Medications\\
\textbf{D. Note \#2 Chartdate: [chartdate2] Headers: Brief Hospital Course, Discharge Medications}\\
E. Note \#2 Chartdate: [chartdate2] Headers: Brief Hospital Course\\
\textit{Correct answer: D.}

\textbf{Q7.} What was the patient's post-operative course and discharge disposition?\\
A. Her post-operative course was uncomplicated with the incision healing well and pain adequately controlled; she was discharged in stable condition to her home with home health services.\\
\textbf{B. Her post-operative course was uncomplicated with the incision healing well and pain adequately controlled; she was discharged in stable condition to an extended care facility.}\\
C. The post-operative course was complicated by episodes of tachycardia and hypotension, requiring IV fluid resuscitation. After she was stabilized, she was discharged to an extended care facility.\\
D. The course was uncomplicated, and her INR became therapeutic on post-operative day 4 after starting warfarin. She was then deemed stable for discharge to an extended care facility for rehabilitation.\\
E. The course was complicated by a superficial wound infection which required a course of IV antibiotics. After showing improvement, she was discharged to an extended care facility to complete a course of oral antibiotics.\\
\textit{Correct answer: B.}

\textbf{Q8.} What are the minimal exact sources from the patient's discharge summaries that fully contain information for your previous answer?\\
A. Note \#2 Chartdate: [chartdate2] Headers: Brief Hospital Course\\
B. Note \#4 Chartdate: [chartdate4] Headers: Brief Hospital Course, Discharge Disposition\\
\textbf{C. Note \#2 Chartdate: [chartdate2] Headers: Brief Hospital Course, Discharge Disposition}\\
D. Note \#2 Chartdate: [chartdate2] Headers: Physical Exam, Discharge Condition\\
E. Note \#2 Chartdate: [chartdate2] Headers: Past Medical History, Medications on Admission\\
\textit{Correct answer: C.}

\textbf{Q9.} In a subsequent admission on [chartdate3], was a new condition identified for which a procedure was recommended?\\
A. Yes, she was newly diagnosed with chronic kidney disease of unclear etiology, and it was recommended that she undergo a renal biopsy as an outpatient.\\
B. Yes, she was diagnosed with AV Nodal Reentrant Tachycardia (AVNRT), and the electrophysiology team recommended implantation of a permanent pacemaker to prevent future episodes.\\
\textbf{C. Yes, she was diagnosed with AV Nodal Reentrant Tachycardia (AVNRT) and an outpatient ablation was recommended.}\\
D. Yes, a 2.2 cm right thyroid nodule was identified on a CT scan, and a fine-needle aspiration biopsy was recommended for further evaluation.\\
E. Yes, she was diagnosed with new-onset atrial fibrillation with rapid ventricular response, for which an electrical cardioversion was recommended by the cardiology team.\\
\textit{Correct answer: C.}

\textbf{Q10.} What are the minimal exact sources from the patient's discharge summaries that fully contain information for your previous answer?\\
A. Note \#3 Chartdate: [chartdate3] Headers: Discharge Diagnosis\\
B. Note \#3 Chartdate: [chartdate3] Headers: Medications on Admission, Discharge Instructions\\
\textbf{C. Note \#3 Chartdate: [chartdate3] Headers: Brief Hospital Course}\\
D. Note \#3 Chartdate: [chartdate3] Headers: Past Medical History, Social History\\
E. Note \#2 Chartdate: [chartdate2] Headers: Brief Hospital Course\\
\textit{Correct answer: C.}

\textbf{Q11.} Was that recommended intervention ever performed?\\
A. No, the notes indicate the patient opted for continued medical management with a beta-blocker and chose to defer the invasive procedure.\\
B. Yes, an EP study was performed, but it failed to induce the tachycardia, so the ablation part of the procedure was aborted.\\
\textbf{C. Yes, the history in the subsequent admission note from [chartdate4] mentions the patient had a ``recent EP study and ablation.''}\\
D. No, the subsequent admission summary does not contain any record of a fine-needle aspiration of her thyroid nodule being performed.\\
E. No, the subsequent admission note does not mention a cardioversion being performed, and instead notes she had an ablation, suggesting a change in the treatment plan.\\
\textit{Correct answer: C.}

\textbf{Q12.} What are the minimal exact sources from the patient's discharge summaries that fully contain information for your previous answer?\\
A. Note \#4 Chartdate: [chartdate4] Headers: Discharge Diagnosis, Discharge Disposition\\
\textbf{B. Note \#4 Chartdate: [chartdate4] Headers: History of Present Illness}\\
C. Note \#2 Chartdate: [chartdate2] Headers: Major Surgical or Invasive Procedure\\
D. Note \#3 Chartdate: [chartdate3] Headers: Brief Hospital Course\\
E. Note \#4 Chartdate: [chartdate4] Headers: Major Surgical or Invasive Procedure\\
\textit{Correct answer: B.}

\textbf{Q13.} After that procedure, did the patient experience any recurrence of related symptoms?\\
A. Yes, because the ablation was not completed, she presented again with a sudden episode of supraventricular tachycardia with a heart rate of 200.\\
\textbf{B. Yes, on the admission following the procedure, she was tachycardic with a heart rate up to 104.}\\
C. Yes, she continued to experience left knee pain, for which she took tramadol as needed according to her medication list in the subsequent admission.\\
D. No, the ablation was successful and she had no further documented cardiac symptoms; her subsequent admission was for an unrelated GI illness.\\
E. Yes, she was tachycardic, and this was determined to be caused by post-ablation pericarditis, a known complication of the procedure.\\
\textit{Correct answer: B.}

\textbf{Q14.} What are the minimal exact sources from the patient's discharge summaries that fully contain information for your previous answer?\\
A. Note \#3 Chartdate: [chartdate3] Headers: History of Present Illness\\
B. Note \#4 Chartdate: [chartdate4] Headers: Discharge Diagnosis, Past Medical History\\
C. Note \#4 Chartdate: [chartdate4] Headers: Physical Exam\\
D. Note \#4 Chartdate: [chartdate4] Headers: Brief Hospital Course\\
\textbf{E. Note \#4 Chartdate: [chartdate4] Headers: Brief Hospital Course, Physical Exam}\\
\textit{Correct answer: E.}

\textbf{Q15.} How was that episode of tachycardia explained and managed?\\
A. The recurrent SVT was managed with 6 mg of IV adenosine, and an electrophysiology consult was placed to discuss a repeat ablation procedure.\\
B. The tachycardia was managed by restarting her on the beta-blocker metoprolol 25 mg daily, which had been previously discontinued.\\
C. The episode was believed to be due to post-ablation atrial irritation, and she was started on a short course of colchicine for presumed pericarditis.\\
\textbf{D. The tachycardia was thought to be secondary to her viral gastroenteritis and it normalized with IV fluids.}\\
E. The episode was identified as AVNRT and was managed with 6 mg of IV adenosine, which converted her to normal sinus rhythm.\\
\textit{Correct answer: D.}

\textbf{Q16.} What are the minimal exact sources from the patient's discharge summaries that fully contain information for your previous answer?\\
A. Note \#3 Chartdate: [chartdate3] Headers: Brief Hospital Course\\
\textbf{B. Note \#4 Chartdate: [chartdate4] Headers: Brief Hospital Course}\\
C. Note \#4 Chartdate: [chartdate4] Headers: Discharge Diagnosis\\
D. Note \#4 Chartdate: [chartdate4] Headers: Past Medical History, Medications on Admission\\
E. Note \#4 Chartdate: [chartdate4] Headers: Discharge Diagnosis, Discharge Instructions\\
\textit{Correct answer: B.}

\textbf{Q17.} Was the patient's cardiac medication regimen changed after the intervention?\\
A. Yes, metoprolol 25 mg daily was restarted to manage her tachycardia, and her lisinopril was temporarily held to avoid hypotension.\\
B. Yes, due to the recurrence of her arrhythmia, her metoprolol was restarted at a higher dose of 50 mg daily and she was referred back to electrophysiology.\\
\textbf{C. Yes, the metoprolol started before the procedure was stopped, as a subsequent note indicates she was no longer on a beta-blocker.}\\
D. No, her cardiac medication regimen was unchanged; she was continued on the metoprolol 25 mg daily that was initiated prior to the ablation.\\
E. Yes, her lisinopril dose was decreased from 40 mg to 20 mg, and furosemide and nifedipine were stopped due to hypotension observed after the procedure.\\
\textit{Correct answer: C.}

\textbf{Q18.} What are the minimal exact sources from the patient's discharge summaries that fully contain information for your previous answer?\\
A. Note \#3 Chartdate: [chartdate3] Headers: History of Present Illness, Note \#4 Chartdate: [chartdate4] Headers: History of Present Illness\\
B. Note \#3 Chartdate: [chartdate3] Headers: Transitional Issues, Note \#4 Chartdate: [chartdate4] Headers: Transitional Issues\\
\textbf{C. Note \#3 Chartdate: [chartdate3] Headers: Brief Hospital Course, Note \#4 Chartdate: [chartdate4] Headers: Brief Hospital Course}\\
D. Note \#3 Chartdate: [chartdate3] Headers: Brief Hospital Course\\
E. Note \#4 Chartdate: [chartdate4] Headers: Brief Hospital Course\\
\textit{Correct answer: C.}
\end{tcolorbox}
\noindent\begin{minipage}{\linewidth}
\captionsetup{type=figure,hypcap=false}
\captionof{figure}{Example multi-turn QA sample from the Procedure category}
\label{fig:example-572}
\end{minipage}

\clearpage
\section{EHRNote-ChatQA vs.\ Prior Patient-Record QA Benchmarks}\label{app:dataset-comparison}

\newcommand{\cmark}{\textcolor{teal}{\checkmark}}
\newcommand{\xmark}{\textcolor{red!70}{\ensuremath{\times}}}
\newcommand{\pmark}{\ensuremath{\sim}}

\begin{table}[!htbp]
\centering
\caption{Comparison with prior patient-record QA benchmarks: EHRNote-ChatQA is the first to combine multi-turn questions, multi-note grounding, and evidence-grounding questions.}
\label{tab:dataset-comparison}
\footnotesize
\setlength{\tabcolsep}{4pt}
\resizebox{\linewidth}{!}{%
\begin{tabular}{l r r r l c c c c l}
\toprule
Benchmark & Patients & Notes & Questions & Source & \shortstack{Multi-\\Note} & \shortstack{Multi-\\Turn} & \shortstack{Evidence\\Grounding} & \shortstack{Expert\\Verified} & \shortstack[l]{Answer\\Format} \\
\midrule
emrQA~\cite{pampari2018emrqa} & --- & --- & 455{,}837$^\dagger$ & Clinical Notes (i2b2/n2c2) & \xmark & \xmark & \pmark & \xmark & Text Span \\
DiSCQ~\cite{lehman2022learning} & 114 & 114 & 2{,}029 & Discharge Summaries (MIMIC-III) & \xmark & \xmark & \xmark & \cmark & No Answer \\
RadQA~\cite{soni2022radqa} & --- & --- & 3{,}074 & Radiology Reports (MIMIC-III) & \xmark & \xmark & \pmark & \cmark & Text Span \\
MedAlign~\cite{fleming2024medalign} & 276 & --- & 983 & Full EHR (Stanford EHR) & \cmark & \xmark & \xmark & \cmark & Free Text \\
EHRNoteQA~\cite{kweon2024ehrnoteqa} & 962 & --- & 962 & Discharge Summaries (MIMIC-IV) & \cmark & \xmark & \xmark & \cmark & Multiple Choice \\
ArchEHR-QA~\cite{soni2025overview} & 134 & 134 & 134 & Discharge Summaries (MIMIC-III/IV) & \xmark & \xmark & \cmark & \cmark & Free Text \\
\midrule
\textbf{EHRNote-ChatQA (Ours)} & \textbf{967} & \textbf{2{,}852}$^\ddagger$ & \textbf{16{,}072} & Discharge Summaries (MIMIC-IV) & \cmark & \cmark & \cmark & \cmark & Multiple Choice \\
\bottomrule
\end{tabular}%
}
\\[2pt]
{\footnotesize $^\dagger$Auto-generated questions using rule-based template slot-filling. $^\ddagger$Total discharge summaries across 967 patients. \pmark{}~(Evidence Grounding): the benchmarks provide extractive text spans as answers, but do not pair complete content answers with separate evidence-grounding questions that ask models to identify the supporting source for each answer.}
\end{table}

\clearpage
\section{Discharge-Summary Structuring Schema}\label{app:schema}

\begin{figure}[!htbp]
\centering
\fbox{\begin{minipage}{0.97\linewidth}
\footnotesize
\textbf{Entity types and per-entity attributes (19 entity types):}
\begin{itemize}\setlength\itemsep{-0.15em}\setlength\leftmargini{1.2em}
  \item \texttt{Diagnosis} --- is\_main, status, finding\_site, laterality, assertion, time
  \item \texttt{Finding} --- is\_main, status, finding\_site, laterality, assertion, time
  \item \texttt{Symptom} --- is\_main, status, finding\_site, laterality, assertion, time
  \item \texttt{Procedure} --- is\_main, status, finding\_site, laterality, assertion, time
  \item \texttt{Outcome} --- status, assertion, time
  \item \texttt{Vital\_Sign} --- status, value, assertion, time
  \item \texttt{Physical\_Exam} --- status, finding, finding\_site, laterality, assertion, time
  \item \texttt{Lab\_Test} --- value, abnormal\_flag, status, assertion, time
  \item \texttt{Diagnostic\_Imaging\_Test} --- status, result, finding\_site, laterality, assertion, time
  \item \texttt{Specimen} --- source, assertion, time
  \item \texttt{Microbiology\_Test} --- organism\_growth, result, assertion, time
  \item \texttt{Microbiology\_Organism} --- growth, assertion, time
  \item \texttt{Microbiology\_Antibiotics} --- dilution, sensitivity, assertion, time
  \item \texttt{Activity} --- status, assertion, time
  \item \texttt{Medication} --- strength, dosage, form\_description, form, route, instruction, disp, refills, assertion, time
  \item \texttt{Event} --- status, assertion, time
  \item \texttt{Allergy} --- status, assertion, time
  \item \texttt{Medical\_Device} --- status, assertion, time
  \item \texttt{Instruction} --- instruction\_text, assertion, time
\end{itemize}
\vspace{0.2em}
\rule{\linewidth}{0.4pt}\\[0.2em]
\textbf{Relationships (23 relationship types):}
\begin{itemize}\setlength\itemsep{-0.15em}\setlength\leftmargini{1.2em}
  \item \texttt{Procedure/Medication} $\rightarrow$ improves $\rightarrow$ \texttt{Diagnosis/Symptom/Finding}
  \item \texttt{Procedure/Medication} $\rightarrow$ worsens $\rightarrow$ \texttt{Diagnosis/Symptom/Finding}
  \item \texttt{Procedure/Medication} $\rightarrow$ causes $\rightarrow$ \texttt{Diagnosis/Symptom/Finding}
  \item \texttt{Procedure/Medication} $\rightarrow$ administered\_for $\rightarrow$ \texttt{Diagnosis/Symptom/Finding}
  \item \texttt{Procedure/Medication} $\rightarrow$ reveals $\rightarrow$ \texttt{Diagnosis/Symptom/Finding}
  \item \texttt{Lab\_Test/Diagnostic\_Imaging\_Test/Physical\_Exam/Vital\_Sign} $\rightarrow$ reveals $\rightarrow$ \texttt{Diagnosis/Finding}
  \item \texttt{Specimen} $\rightarrow$ has\_test $\rightarrow$ \texttt{Microbiology\_Test/Lab\_Test}
  \item \texttt{Microbiology\_Test} $\rightarrow$ detected $\rightarrow$ \texttt{Microbiology\_Organism}
  \item \texttt{Microbiology\_Organism} $\rightarrow$ tested\_against $\rightarrow$ \texttt{Microbiology\_Antibiotics}
  \item \texttt{Microbiology\_Organism} $\rightarrow$ reveals $\rightarrow$ \texttt{Diagnosis/Finding}
  \item \texttt{Diagnosis/Symptom/Finding} $\rightarrow$ causes $\rightarrow$ \texttt{Diagnosis/Symptom/Finding}
  \item \texttt{Diagnosis/Symptom/Finding} $\rightarrow$ causes $\rightarrow$ \texttt{Event}
  \item \texttt{Diagnosis/Symptom/Finding} $\rightarrow$ resulted\_in $\rightarrow$ \texttt{Outcome/Finding/Activity}
  \item \texttt{Outcome/Finding} $\rightarrow$ after $\rightarrow$ \texttt{Diagnosis/Symptom/Finding}
  \item \texttt{Outcome/Finding} $\rightarrow$ after $\rightarrow$ \texttt{Procedure/Medication}
  \item \texttt{Lab\_Test/Diagnostic\_Imaging\_Test/Physical\_Exam/Vital\_Sign} $\rightarrow$ resulted\_in $\rightarrow$ \texttt{Procedure/Medication}
  \item \texttt{Procedure/Medication} $\rightarrow$ causes $\rightarrow$ \texttt{Procedure/Medication}
  \item \texttt{Procedure/Medication} $\rightarrow$ switched\_to $\rightarrow$ \texttt{Procedure/Medication}
  \item \texttt{Procedure/Medication} $\rightarrow$ causes $\rightarrow$ \texttt{Event}
  \item \texttt{Procedure/Medication} $\rightarrow$ resulted\_in $\rightarrow$ \texttt{Outcome/Finding}
  \item \texttt{Event} $\rightarrow$ causes $\rightarrow$ \texttt{Diagnosis/Symptom/Finding}
  \item \texttt{Event} $\rightarrow$ causes $\rightarrow$ \texttt{Procedure/Medication}
  \item (any entity) $\rightarrow$ has\_instruction $\rightarrow$ \texttt{Instruction}
\end{itemize}
\end{minipage}}
\caption{Discharge-summary schema: 19 clinical entity types with per-entity attributes (top) and 23 relationship types linking entities (bottom). The \texttt{is\_main} attribute marks principal diagnoses, symptoms, findings, and procedures of the admission. \texttt{causes} and \texttt{resulted\_in} are distinguished as follows: \texttt{causes} denotes adverse effects or complications, whereas \texttt{resulted\_in} denotes desired or expected outcomes.}
\label{fig:schema}
\end{figure}

\clearpage
\section{Multi-turn QA Templates}\label{app:template}

The full set of multi-turn QA templates, defined in collaboration with medical experts, is shown below. The templates span all eight question categories in both single-note and multi-note settings, yielding 16 template sets of five templates each (80 templates in total), and build on the discharge-summary structuring schema (the comprehensive entity, attribute, and relationship definitions in Appendix~\ref{app:schema}). For each template, every answer \textbf{A\#} specifies the schema entities, attributes, and relationships that ground the answer to the paired question \textbf{Q\#} in the source notes; attributes are written in dot notation after the entity (e.g., \texttt{Medication.dosage}), and relationships as $\langle$entity$\rangle$ $\rightarrow$ relationship $\rightarrow$ $\langle$entity$\rangle$ triples.

{\small

\subsection{Diagnosis}

\subsubsection{Single-note setting}
\textbf{Template 1 --- Treatment Response and Diagnostic Outcome}\\[0.15em]
\textbf{Q1.} What diagnosis was treated during admission?\\
\textbf{A1.} (Diagnosis.is\_main, Diagnosis.assertion)\\
\textbf{Q2.} What medications or procedures were administered for the condition?\\
\textbf{A2.} (Procedure/Medication $\rightarrow$ administered\_for $\rightarrow$ Diagnosis, Medication.dosage, route, time)\\
\textbf{Q3.} What was the documented clinical reasoning for choosing that particular treatment approach over alternatives?\\
\textbf{A3.} (Procedure/Medication $\rightarrow$ administered\_for $\rightarrow$ Diagnosis; clinical reasoning from Brief Hospital Course linking findings to treatment decision)\\
\textbf{Q4.} Did the condition improve, worsen, or remain unchanged after treatment?\\
\textbf{A4.} (Procedure/Medication $\rightarrow$ improves/worsens $\rightarrow$ Diagnosis, Diagnosis.status over time, Outcome $\rightarrow$ after $\rightarrow$ Procedure/Medication, Procedure/Medication $\rightarrow$ resulted\_in $\rightarrow$ Outcome)\\
\textbf{Q5.} Were there any treatment side effects or complications affecting the condition?\\
\textbf{A5.} (Procedure/Medication $\rightarrow$ causes $\rightarrow$ Diagnosis/Symptom/Finding/Event)\\
\textbf{Q6.} What was the final clinical outcome related to the condition at discharge?\\
\textbf{A6.} (Diagnosis $\rightarrow$ resulted\_in $\rightarrow$ Outcome, Outcome.status, time = discharge)

\textbf{Template 2 --- Comorbidity Interaction and Diagnostic Interrelationships}\\[0.15em]
\textbf{Q1.} Were any comorbid diagnoses present during admission?\\
\textbf{A1.} (Diagnosis.is\_main = false, Diagnosis.assertion)\\
\textbf{Q2.} Did any of those conditions contribute to or worsen another?\\
\textbf{A2.} (Diagnosis $\rightarrow$ causes $\rightarrow$ Diagnosis, Diagnosis $\rightarrow$ worsens $\rightarrow$ Diagnosis)\\
\textbf{Q3.} Was there any clinical evidence documented in the notes that supported that interaction between those conditions?\\
\textbf{A3.} (Lab\_Test/Diagnostic\_Imaging\_Test/Physical\_Exam $\rightarrow$ reveals $\rightarrow$ Diagnosis; clinical reasoning connecting documented findings to the interaction)\\
\textbf{Q4.} Did the interaction result in any complication or adverse event?\\
\textbf{A4.} (Diagnosis $\rightarrow$ causes $\rightarrow$ Event, Diagnosis $\rightarrow$ resulted\_in $\rightarrow$ Outcome)\\
\textbf{Q5.} At discharge, which of those conditions were stabilized, and which required ongoing management?\\
\textbf{A5.} (Diagnosis.status at discharge, Diagnosis $\rightarrow$ has\_instruction $\rightarrow$ Instruction)

\textbf{Template 3 --- Primary Diagnosis Identification and Clinical Basis}\\[0.15em]
\textbf{Q1.} What is the primary diagnosis for the patient?\\
\textbf{A1.} (Diagnosis.is\_main=true, Diagnosis.assertion, time=admission)\\
\textbf{Q2.} Were there any symptoms or findings on presentation that led to the identification of that condition?\\
\textbf{A2.} (Symptom/Finding $\rightarrow$ causes $\rightarrow$ Diagnosis; Symptom.time=admission; Physical\_Exam.finding, Vital\_Sign $\rightarrow$ reveals $\rightarrow$ Diagnosis)\\
\textbf{Q3.} How did the documented evaluation narrow the differential to that condition?\\
\textbf{A3.} (Lab\_Test/Diagnostic\_Imaging\_Test/Microbiology\_Test $\rightarrow$ reveals $\rightarrow$ Diagnosis; Lab\_Test.abnormal\_flag; Diagnostic\_Imaging\_Test.result; clinical reasoning linking findings to diagnosis confirmation)\\
\textbf{Q4.} What treatments were given specifically for the condition during the hospitalization?\\
\textbf{A4.} (Procedure/Medication $\rightarrow$ administered\_for $\rightarrow$ Diagnosis; Medication.dosage, Medication.route, Medication.frequency; Procedure.status)\\
\textbf{Q5.} What was the status of the condition at discharge?\\
\textbf{A5.} (Diagnosis.status at discharge; Outcome.status; Diagnosis $\rightarrow$ resulted\_in $\rightarrow$ Outcome)

\textbf{Template 4 --- Differential Diagnosis Workup and Resolution}\\[0.15em]
\textbf{Q1.} Was any diagnosis initially suspected or considered as a working diagnosis when the patient was first evaluated?\\
\textbf{A1.} (Diagnosis.assertion=suspected/possible, Diagnosis.is\_main, time=early admission)\\
\textbf{Q2.} Were any other conditions considered as part of the differential during the workup?\\
\textbf{A2.} (Diagnosis.assertion=possible/suspected/ruled\_out; all non-primary Diagnosis entities, time=admission)\\
\textbf{Q3.} What tests were performed to differentiate between those possibilities, and what were the results?\\
\textbf{A3.} (Lab\_Test/Diagnostic\_Imaging\_Test/Microbiology\_Test $\rightarrow$ reveals $\rightarrow$ Diagnosis/Finding; Lab\_Test.abnormal\_flag; Specimen $\rightarrow$ has\_test $\rightarrow$ Lab\_Test/Microbiology\_Test; test result values)\\
\textbf{Q4.} Based on those results, were any conditions ruled out, and on what basis?\\
\textbf{A4.} (Diagnosis.assertion=ruled\_out; Lab\_Test.result=negative/normal; Diagnostic\_Imaging\_Test $\rightarrow$ reveals $\rightarrow$ Finding that excludes diagnosis)\\
\textbf{Q5.} Was any evidence documented that confirmed the final diagnosis?\\
\textbf{A5.} (Diagnosis.status=confirmed; Lab\_Test/Diagnostic\_Imaging\_Test/Physical\_Exam $\rightarrow$ reveals $\rightarrow$ Diagnosis; key positive findings supporting confirmation)\\
\textbf{Q6.} Were any incidental or secondary diagnoses identified during the workup?\\
\textbf{A6.} (Diagnosis.is\_main=false; Diagnostic\_Imaging\_Test $\rightarrow$ reveals $\rightarrow$ secondary Diagnosis/Finding; secondary Diagnosis.assertion, time relative to primary)

\textbf{Template 5 --- Diagnosis Etiology, Complications, and Causal Chain}\\[0.15em]
\textbf{Q1.} What diagnosis was the patient admitted for, and what was it attributed to?\\
\textbf{A1.} (Diagnosis.is\_main, Diagnosis.assertion; Event/Diagnosis/Finding $\rightarrow$ causes $\rightarrow$ Diagnosis; time=admission)\\
\textbf{Q2.} Were there any pre-existing conditions, prior medications, or procedures that may have precipitated the condition?\\
\textbf{A2.} (Diagnosis/Symptom/Finding $\rightarrow$ causes $\rightarrow$ Diagnosis; Procedure/Medication $\rightarrow$ causes $\rightarrow$ Diagnosis; prior Diagnosis.status=active/chronic)\\
\textbf{Q3.} How did the documented clinical course demonstrate the causal relationship between those contributing factors and the condition?\\
\textbf{A3.} (Lab\_Test/Diagnostic\_Imaging\_Test/Physical\_Exam $\rightarrow$ reveals $\rightarrow$ Diagnosis; temporal and causal evidence from notes linking precipitants to the diagnosis)\\
\textbf{Q4.} Did the condition lead to any complications or secondary diagnoses during the hospitalization?\\
\textbf{A4.} (Diagnosis $\rightarrow$ causes $\rightarrow$ Diagnosis/Symptom/Finding/Event; secondary Diagnosis.assertion; time=during admission)\\
\textbf{Q5.} How were those complications managed, and what was their status at discharge?\\
\textbf{A5.} (Procedure/Medication $\rightarrow$ administered\_for $\rightarrow$ secondary Diagnosis; Diagnosis.status at discharge; Outcome $\rightarrow$ after $\rightarrow$ Procedure/Medication)\\
\textbf{Q6.} Were any instructions given to prevent recurrence or manage the underlying cause after discharge?\\
\textbf{A6.} (Diagnosis $\rightarrow$ has\_instruction $\rightarrow$ Instruction; Instruction.instruction\_text for prevention, risk modification, or monitoring)

\subsubsection{Multi-note setting}
\textbf{Template 1 --- Longitudinal Problem Timeline and Status Transitions}\\[0.15em]
\textbf{Q1.} Across all admissions, what diagnoses did the patient have, and when did each first appear?\\
\textbf{A1.} (Diagnosis.assertion; Diagnosis.time across notes; first-occurrence time)\\
\textbf{Q2.} For those conditions, what was their clinical status at the time of each discharge?\\
\textbf{A2.} (Diagnosis.status over time; Outcome.status; time = discharge)\\
\textbf{Q3.} Among those conditions, did any recur or lead to re-hospitalization, and what was the interval between episodes?\\
\textbf{A3.} (Diagnosis.time comparisons; Event.time related to recurrence)\\
\textbf{Q4.} Was there any clinical evidence supporting recurrence or persistence of those conditions during subsequent admissions?\\
\textbf{A4.} (Lab\_Test/Diagnostic\_Imaging\_Test/Physical\_Exam/Vital\_Sign $\rightarrow$ reveals $\rightarrow$ Diagnosis; time = admission)\\
\textbf{Q5.} How did the treatment approach evolve across admissions to address the persistence of those conditions?\\
\textbf{A5.} (Medication/Procedure changes across admissions $\rightarrow$ administered\_for $\rightarrow$ Diagnosis; clinical reasoning from notes linking prior treatment failure to subsequent treatment modification)\\
\textbf{Q6.} By the most recent discharge, were any of those conditions still active, and what was the follow-up plan for them?\\
\textbf{A6.} (Diagnosis.status = active; Diagnosis $\rightarrow$ has\_instruction $\rightarrow$ Instruction)

\textbf{Template 2 --- Diagnostic Reasoning and Differential Resolution}\\[0.15em]
\textbf{Q1.} What was the primary condition being evaluated in the most recent admission?\\
\textbf{A1.} (Diagnosis.is\_main; Diagnosis.assertion; time = admission)\\
\textbf{Q2.} Were there any symptoms, findings, or clinical changes that prompted evaluation for that condition?\\
\textbf{A2.} (Diagnosis/Symptom/Finding $\rightarrow$ causes $\rightarrow$ Diagnosis; time = admission)\\
\textbf{Q3.} What diagnostic tests were obtained to evaluate the condition, and were there any abnormal or positive results?\\
\textbf{A3.} (Lab\_Test/Diagnostic\_Imaging\_Test/Vital\_Sign/Physical\_Exam $\rightarrow$ reveals $\rightarrow$ Diagnosis; Lab\_Test.abnormal\_flag; Diagnostic\_Imaging\_Test.result)\\
\textbf{Q4.} Compared with the prior admission, were those findings new, worse, or unchanged?\\
\textbf{A4.} (Diagnosis.status comparison; Lab\_Test value trends; Diagnostic\_Imaging\_Test comparisons; Vital\_Sign trends; Physical\_Exam trends)\\
\textbf{Q5.} What was the documented clinical reasoning that explained the change in the condition's status from the prior admission?\\
\textbf{A5.} (clinical reasoning from notes linking cross-admission finding trends to the current condition status; causal factors documented)\\
\textbf{Q6.} By discharge, was the condition confirmed, ruled out, or left as suspected?\\
\textbf{A6.} (Diagnosis.status at discharge; Outcome $\rightarrow$ after $\rightarrow$ Diagnosis)\\
\textbf{Q7.} Were any discharge instructions provided for monitoring or follow-up of the condition?\\
\textbf{A7.} (Diagnosis $\rightarrow$ has\_instruction $\rightarrow$ Instruction)

\textbf{Template 3 --- Diagnosis Course and Treatment Response Across Admissions}\\[0.15em]
\textbf{Q1.} What was the primary diagnosis at admission $\langle$chartdate or admission\_id$\rangle$?\\
\textbf{A1.} (Diagnosis.is\_main; Diagnosis.assertion; time = $\langle$admission$\rangle$)\\
\textbf{Q2.} Were there any symptoms, findings, or clinical events that led clinicians to identify the condition?\\
\textbf{A2.} (Symptom/Finding $\rightarrow$ reveals $\rightarrow$ Diagnosis; Event $\rightarrow$ causes $\rightarrow$ Diagnosis)\\
\textbf{Q3.} Were any diagnostic tests performed that supported or confirmed the condition?\\
\textbf{A3.} (Lab\_Test/Diagnostic\_Imaging\_Test/Physical\_Exam $\rightarrow$ reveals $\rightarrow$ Diagnosis; abnormal findings)\\
\textbf{Q4.} What treatments were administered to manage the condition?\\
\textbf{A4.} (Procedure / Medication $\rightarrow$ administered\_for $\rightarrow$ Diagnosis; Medication.dosage; Medication.route)\\
\textbf{Q5.} During hospitalization, how did the condition respond to those treatments?\\
\textbf{A5.} (Medication/Procedure $\rightarrow$ improves or worsens $\rightarrow$ Diagnosis; Diagnosis.status trend)\\
\textbf{Q6.} What was the final status of the condition at discharge?\\
\textbf{A6.} (Diagnosis.status at discharge; Outcome.status)

\textbf{Template 4 --- Differential Diagnosis Reasoning and Confirmation}\\[0.15em]
\textbf{Q1.} Was any diagnosis initially suspected on presentation for admission $\langle$chartdate or admission\_id$\rangle$?\\
\textbf{A1.} (Diagnosis.is\_main at admission; Diagnosis.assertion = suspected)\\
\textbf{Q2.} During the evaluation of that suspected condition, were any other diagnostic possibilities considered?\\
\textbf{A2.} (Diagnosis entities with assertion = possible/suspected/ruled out)\\
\textbf{Q3.} Were any diagnostic tests ordered to differentiate between those possibilities?\\
\textbf{A3.} (Lab\_Test / Diagnostic\_Imaging\_Test / Microbiology\_Test ordered; Specimen $\rightarrow$ has\_test $\rightarrow$ Lab\_Test)\\
\textbf{Q4.} Based on those test results, were any possibilities ruled out, and why?\\
\textbf{A4.} (Diagnosis.assertion = ruled out; Lab\_Test.result; Diagnostic\_Imaging\_Test findings excluding diagnosis)\\
\textbf{Q5.} Were any findings that ultimately confirmed the final diagnosis documented?\\
\textbf{A5.} (Lab\_Test / Diagnostic\_Imaging\_Test / Physical\_Exam / Vital\_Sign $\rightarrow$ reveals $\rightarrow$ Diagnosis)\\
\textbf{Q6.} During the workup for that condition, were any additional incidental findings or secondary diagnoses discovered?\\
\textbf{A6.} (Diagnosis.is\_main = false; Diagnostic\_Imaging\_Test $\rightarrow$ reveals $\rightarrow$ Finding/Diagnosis)

\textbf{Template 5 --- Etiology, Complications, and Prevention of the Admission Diagnosis}\\[0.15em]
\textbf{Q1.} What diagnosis was the patient admitted for during admission $\langle$chartdate or admission\_id$\rangle$?\\
\textbf{A1.} (Diagnosis.is\_main; Diagnosis.assertion; time = most recent admission)\\
\textbf{Q2.} Were there any underlying conditions, events, or findings that contributed to the development of the condition?\\
\textbf{A2.} (Diagnosis/Symptom/Finding $\rightarrow$ causes $\rightarrow$ Diagnosis; Event $\rightarrow$ causes $\rightarrow$ Diagnosis)\\
\textbf{Q3.} Were there any procedures or medications from prior admissions that may have contributed to the condition?\\
\textbf{A3.} (Procedure/Medication $\rightarrow$ causes $\rightarrow$ Diagnosis; time from prior notes)\\
\textbf{Q4.} What was the documented clinical reasoning that connected those contributing factors from the current and prior admissions to the current condition?\\
\textbf{A4.} (clinical reasoning from notes linking cross-admission contributing factors to the diagnosis; Lab\_Test/Diagnostic\_Imaging\_Test/Physical\_Exam/Microbiology\_Test $\rightarrow$ reveals $\rightarrow$ Diagnosis)\\
\textbf{Q5.} Did the condition lead to any complications, secondary diagnoses, or clinical events?\\
\textbf{A5.} (Diagnosis $\rightarrow$ causes $\rightarrow$ Diagnosis/Symptom/Finding/Event; Diagnosis $\rightarrow$ resulted\_in $\rightarrow$ Outcome)\\
\textbf{Q6.} Were any treatments administered for those complications, and if so, how were they addressed?\\
\textbf{A6.} (Procedure / Medication $\rightarrow$ administered\_for $\rightarrow$ causal Diagnosis; Procedure/Medication $\rightarrow$ improves/worsens $\rightarrow$ condition)\\
\textbf{Q7.} Were any preventive measures or follow-up instructions provided to reduce the risk of recurrence?\\
\textbf{A7.} (Diagnosis $\rightarrow$ has\_instruction $\rightarrow$ Instruction)

\subsection{Symptom}

\subsubsection{Single-note setting}
\textbf{Template 1 --- Symptom Progression and Secondary Complication During Hospitalization}\\[0.15em]
\textbf{Q1.} What was the initial presenting symptom at admission?\\
\textbf{A1.} (Symptom.is\_main, Symptom.time=admission, Symptom.assertion)\\
\textbf{Q2.} Did the initial symptom or its underlying condition lead to additional symptoms or complications during the hospital stay?\\
\textbf{A2.} (Symptom $\rightarrow$ causes $\rightarrow$ Symptom/Diagnosis; Diagnosis $\rightarrow$ causes $\rightarrow$ Symptom; Symptom.time progression during admission)\\
\textbf{Q3.} How were those secondary or evolving symptoms evaluated?\\
\textbf{A3.} (Lab\_Test/Diagnostic\_Imaging\_Test/Physical\_Exam/Vital\_Sign $\rightarrow$ reveals $\rightarrow$ Diagnosis/Finding related to secondary Symptom)\\
\textbf{Q4.} What was the documented clinical reasoning that linked those evaluation findings to the secondary symptoms?\\
\textbf{A4.} (clinical reasoning from notes connecting documented evaluation results to the secondary symptom etiology; Diagnosis/Finding $\rightarrow$ causes $\rightarrow$ Symptom)\\
\textbf{Q5.} Were any treatments directed at the secondary symptoms, and were they distinct from treatments for the primary symptom?\\
\textbf{A5.} (Medication/Procedure $\rightarrow$ administered\_for $\rightarrow$ secondary Symptom; compare with primary Symptom treatments)\\
\textbf{Q6.} Did any treatments for the primary symptom worsen or cause the secondary symptoms?\\
\textbf{A6.} (Medication/Procedure $\rightarrow$ causes $\rightarrow$ secondary Symptom; Medication/Procedure $\rightarrow$ worsens $\rightarrow$ Symptom)\\
\textbf{Q7.} What was the status of all symptoms at the time of discharge?\\
\textbf{A7.} (Symptom.status at discharge for each Symptom; Outcome $\rightarrow$ after $\rightarrow$ Symptom)

\textbf{Template 2 --- Symptom Burden, Functional Impact, and Discharge Planning}\\[0.15em]
\textbf{Q1.} Were there any symptoms that impaired the patient's functional status or activities of daily living during admission?\\
\textbf{A1.} (Symptom.is\_main, Symptom.status; Symptom $\rightarrow$ resulted\_in $\rightarrow$ Activity limitation; Activity.status)\\
\textbf{Q2.} How did those symptoms affect the patient's ability to ambulate, self-care, or perform basic tasks?\\
\textbf{A2.} (Symptom $\rightarrow$ resulted\_in $\rightarrow$ Activity; Physical\_Exam.finding related to functional performance)\\
\textbf{Q3.} Were any interventions used to support functional recovery?\\
\textbf{A3.} (Procedure/Medication $\rightarrow$ administered\_for $\rightarrow$ Symptom; Medical\_Device $\rightarrow$ improves $\rightarrow$ Symptom/Activity; Activity.status post-intervention)\\
\textbf{Q4.} How did the care team determine the discharge disposition based on those functional limitations?\\
\textbf{A4.} (Symptom $\rightarrow$ resulted\_in $\rightarrow$ Outcome; clinical reasoning from notes linking the symptom burden and functional status to the disposition decision)\\
\textbf{Q5.} Were any activity restrictions or precautions related to the patient's symptoms specified in the discharge instructions?\\
\textbf{A5.} (Symptom/Activity $\rightarrow$ has\_instruction $\rightarrow$ Instruction; restrictions, duration, escalation criteria)\\
\textbf{Q6.} Were any symptoms or warning signs documented that should prompt the patient to seek urgent care after discharge?\\
\textbf{A6.} (Symptom $\rightarrow$ has\_instruction $\rightarrow$ Instruction; threshold criteria for return to ED/clinic)

\textbf{Template 3 --- Presenting Symptom, Diagnostic Workup, and Treatment Response}\\[0.15em]
\textbf{Q1.} What was the primary presenting symptom at admission, and when did it begin?\\
\textbf{A1.} (Symptom.is\_main = true; Symptom.time = onset relative to admission; Symptom.assertion = present)\\
\textbf{Q2.} Was any condition identified as the cause of the symptom?\\
\textbf{A2.} (Diagnosis/Finding $\rightarrow$ causes $\rightarrow$ Symptom)\\
\textbf{Q3.} Were any laboratory tests, imaging studies, vital signs, or physical exam findings used to evaluate the symptom, and were any abnormal findings documented?\\
\textbf{A3.} (Lab\_Test/Diagnostic\_Imaging\_Test/Physical\_Exam/Vital\_Sign $\rightarrow$ reveals $\rightarrow$ Diagnosis/Finding, Lab\_Test.abnormal\_flag; Diagnostic\_Imaging\_Test.result)\\
\textbf{Q4.} Were any treatments administered specifically for the symptom?\\
\textbf{A4.} (Medication/Procedure $\rightarrow$ administered\_for $\rightarrow$ Symptom)\\
\textbf{Q5.} Did the symptom improve, persist, or worsen during hospitalization?\\
\textbf{A5.} (Medication/Procedure $\rightarrow$ improves/worsens $\rightarrow$ Symptom; Symptom.status trend)\\
\textbf{Q6.} What was the status of the symptom at discharge?\\
\textbf{A6.} (Symptom.status at discharge; Outcome.status)\\
\textbf{Q7.} Were any discharge instructions provided regarding monitoring or recurrence of the symptom?\\
\textbf{A7.} (Symptom $\rightarrow$ has\_instruction $\rightarrow$ Instruction)

\textbf{Template 4 --- Chronic Symptom Exacerbation During Single Admission}\\[0.15em]
\textbf{Q1.} Did the patient have a chronic symptom prior to admission, and if so, what was its baseline severity and duration?\\
\textbf{A1.} (Symptom.status = chronic; Symptom.time duration; Symptom.assertion)\\
\textbf{Q2.} What changes in the symptom prompted admission?\\
\textbf{A2.} (Symptom.status = worsened/exacerbated; Symptom.time = pre-admission period)\\
\textbf{Q3.} Were any factors documented as contributing to the worsening of the symptom?\\
\textbf{A3.} (Event $\rightarrow$ causes $\rightarrow$ Symptom, Medication.status = non-compliance/discontinued, Diagnosis $\rightarrow$ causes $\rightarrow$ Symptom)\\
\textbf{Q4.} What was the documented clinical reasoning that connected those contributing factors to the exacerbation?\\
\textbf{A4.} (clinical reasoning from notes linking the documented triggers to the symptom worsening; temporal and causal evidence)\\
\textbf{Q5.} Were any interventions implemented to manage the exacerbation?\\
\textbf{A5.} (Medication/Procedure $\rightarrow$ administered\_for $\rightarrow$ Symptom)\\
\textbf{Q6.} How did the symptom respond to those interventions?\\
\textbf{A6.} (Medication/Procedure $\rightarrow$ improves/worsens $\rightarrow$ Symptom; Symptom.status trend)\\
\textbf{Q7.} Was long-term symptom management adjusted at discharge?\\
\textbf{A7.} (Medication.time = discharge; Medication.dosage/route changes; Symptom $\rightarrow$ has\_instruction $\rightarrow$ Instruction)

\textbf{Template 5 --- New Symptom Development During Hospitalization}\\[0.15em]
\textbf{Q1.} Did the patient develop any new symptoms during hospitalization? If so, what and when?\\
\textbf{A1.} (Symptom.time = during admission; Symptom.assertion = new)\\
\textbf{Q2.} Was any cause suspected for the new symptom?\\
\textbf{A2.} (Medication/Procedure $\rightarrow$ causes $\rightarrow$ Symptom, Diagnosis $\rightarrow$ causes $\rightarrow$ Symptom, Event $\rightarrow$ causes $\rightarrow$ Symptom)\\
\textbf{Q3.} Was any diagnostic evaluation performed for the new symptom?\\
\textbf{A3.} (Lab\_Test/Diagnostic\_Imaging\_Test/Physical\_Exam/Vital\_Sign $\rightarrow$ reveals $\rightarrow$ Diagnosis/Finding)\\
\textbf{Q4.} Did the documented findings confirm or refute the suspected cause based on those evaluation results?\\
\textbf{A4.} (clinical reasoning from notes linking evaluation results to the suspected etiology; Diagnosis/Finding confirmation or exclusion)\\
\textbf{Q5.} Were any management changes made in response to the symptom?\\
\textbf{A5.} (Medication.status = held/discontinued, Medication $\rightarrow$ switched\_to $\rightarrow$ Medication, Procedure $\rightarrow$ administered\_for $\rightarrow$ Symptom)\\
\textbf{Q6.} Did the new symptom resolve before discharge?\\
\textbf{A6.} (Symptom.status = resolved/persistent; Medication/Procedure $\rightarrow$ improves $\rightarrow$ Symptom)\\
\textbf{Q7.} Were any precautions given to prevent recurrence of the symptom?\\
\textbf{A7.} (Symptom/Medication $\rightarrow$ has\_instruction $\rightarrow$ Instruction)

\subsubsection{Multi-note setting}
\textbf{Template 1 --- Readmission Trigger, Evaluation, and Treatment Response}\\[0.15em]
\textbf{Q1.} After discharge on admission $\langle$chartdate or admission\_id$\rangle$, was there any symptom that prompted the patient to return to the hospital, and when did it begin?\\
\textbf{A1.} (Symptom.is\_main, Symptom.time relative to last discharge)\\
\textbf{Q2.} Was any condition identified as the cause of the symptom?\\
\textbf{A2.} (Diagnosis/Finding $\rightarrow$ causes $\rightarrow$ Symptom)\\
\textbf{Q3.} Were any laboratory tests, imaging studies, physical examination findings, or vital sign abnormalities used to evaluate the condition?\\
\textbf{A3.} (Lab\_Test/Diagnostic\_Imaging\_Test/Physical\_Exam/Vital\_Sign $\rightarrow$ reveals $\rightarrow$ Diagnosis/Finding; Lab\_Test.abnormal\_flag; Diagnostic\_Imaging\_Test.result)\\
\textbf{Q4.} Was any treatment or intervention administered for the symptom or its underlying condition?\\
\textbf{A4.} (Medication/Procedure $\rightarrow$ administered\_for $\rightarrow$ Symptom/Diagnosis/Finding)\\
\textbf{Q5.} After the treatment was started, did the symptom improve, persist, or worsen, and was there any clinical evidence supporting the assessment?\\
\textbf{A5.} (Medication/Procedure $\rightarrow$ improves/worsens $\rightarrow$ Symptom; Symptom.status; relevant Lab\_Test/Vital\_Sign/Physical\_Exam changes)\\
\textbf{Q6.} Were any discharge instructions or return precautions provided regarding the symptom or its underlying condition?\\
\textbf{A6.} (Symptom $\rightarrow$ has\_instruction $\rightarrow$ Instruction; Diagnosis $\rightarrow$ has\_instruction $\rightarrow$ Instruction)

\textbf{Template 2 --- In-Hospital or Treatment-Related Symptom Development and Management}\\[0.15em]
\textbf{Q1.} During the hospitalization on $\langle$chartdate$\rangle$, did the patient develop any new symptom after a medication or procedure was initiated, and when did that symptom appear?\\
\textbf{A1.} (Symptom.time during admission; Medication/Procedure.time)\\
\textbf{Q2.} Was any medication or procedure suspected to have caused that symptom?\\
\textbf{A2.} (Medication/Procedure $\rightarrow$ causes $\rightarrow$ Symptom/Finding; Symptom $\rightarrow$ causes $\rightarrow$ Diagnosis/Symptom/Finding)\\
\textbf{Q3.} Were any diagnostic evaluations performed to assess the suspected adverse effect or to rule out complications?\\
\textbf{A3.} (Lab\_Test/Diagnostic\_Imaging\_Test/Vital\_Sign/Physical\_Exam $\rightarrow$ reveals $\rightarrow$ Diagnosis/Symptom/Finding; Lab\_Test.abnormal\_flag)\\
\textbf{Q4.} What was the documented clinical reasoning that confirmed or refuted the relationship between the treatment and the new symptom?\\
\textbf{A4.} (clinical reasoning from notes linking the temporal relationship and evaluation findings to the suspected causal mechanism; Medication/Procedure $\rightarrow$ causes $\rightarrow$ Symptom)\\
\textbf{Q5.} Were any management changes made to address the symptom?\\
\textbf{A5.} (Medication.status = held/discontinued; Medication.dosage; Medication $\rightarrow$ switched\_to $\rightarrow$ Medication; Medication/Procedure $\rightarrow$ improves/worsens $\rightarrow$ Symptom)\\
\textbf{Q6.} Were any monitoring or precautionary instructions given at discharge to prevent recurrence of the symptom?\\
\textbf{A6.} (Symptom $\rightarrow$ has\_instruction $\rightarrow$ Instruction; Medication $\rightarrow$ has\_instruction $\rightarrow$ Instruction)

\textbf{Template 3 --- Recurrent Symptom Pattern Across Admissions}\\[0.15em]
\textbf{Q1.} Was there any symptom that occurred repeatedly across multiple hospitalizations for this patient?\\
\textbf{A1.} (Symptom appears in >=2 notes; Symptom.time)\\
\textbf{Q2.} Were there any events, medication changes, or clinical triggers that occurred before episodes of the symptom?\\
\textbf{A2.} (Event $\rightarrow$ causes $\rightarrow$ Symptom; Medication.status changes; Activity/environmental triggers across notes)\\
\textbf{Q3.} How did the severity or clinical presentation of the symptom differ between hospitalizations?\\
\textbf{A3.} (compare Symptom.status; Physical\_Exam.finding; Vital\_Sign.value associated Symptoms)\\
\textbf{Q4.} Were any conditions associated with the symptom across admissions?\\
\textbf{A4.} (Diagnosis/Finding/Symptom $\rightarrow$ causes $\rightarrow$ Symptom)\\
\textbf{Q5.} How did the treatment approach evolve over time across admissions?\\
\textbf{A5.} (Medication/Procedure $\rightarrow$ administered\_for $\rightarrow$ Symptom; clinical reasoning from notes explaining how prior treatment outcomes informed subsequent treatment choices)\\
\textbf{Q6.} Among those treatments, were any that improved the symptom, and were any that were ineffective or worsened it?\\
\textbf{A6.} (Medication/Procedure $\rightarrow$ improves $\rightarrow$ Symptom; Medication/Procedure $\rightarrow$ worsens $\rightarrow$ Symptom)\\
\textbf{Q7.} Were there hospitalizations where that symptom was absent or well controlled?\\
\textbf{A7.} (Symptom.status across notes; Symptom $\rightarrow$ resulted\_in $\rightarrow$ Outcome/Finding)\\
\textbf{Q8.} Were any preventive strategies or maintenance therapies recommended to prevent recurrence of the symptom?\\
\textbf{A8.} (Medication.time = discharge; Symptom $\rightarrow$ has\_instruction $\rightarrow$ Instruction)

\textbf{Template 4 --- Symptom Progression and Secondary Symptom Development}\\[0.15em]
\textbf{Q1.} What was the initial presenting complaint at admission $\langle$chartdate or admission\_id$\rangle$?\\
\textbf{A1.} (Symptom.is\_main; Symptom.time = admission)\\
\textbf{Q2.} Was any condition identified as the cause of the symptom?\\
\textbf{A2.} (Diagnosis/Symptom/Finding $\rightarrow$ causes $\rightarrow$ Symptom)\\
\textbf{Q3.} Did the condition or symptom lead to any additional symptoms during the hospitalization?\\
\textbf{A3.} (Diagnosis/Symptom/Finding $\rightarrow$ causes $\rightarrow$ Symptom)\\
\textbf{Q4.} Did any treatment administered for the initial symptom lead to new symptoms?\\
\textbf{A4.} (Medication/Procedure $\rightarrow$ administered\_for $\rightarrow$ Symptom; Medication/Procedure $\rightarrow$ causes $\rightarrow$ Symptom)\\
\textbf{Q5.} Were any treatments administered to manage the secondary or treatment-related symptoms?\\
\textbf{A5.} (Medication/Procedure $\rightarrow$ administered\_for $\rightarrow$ Symptom)\\
\textbf{Q6.} Did those symptoms lead to any complications or clinical events affecting the hospital course?\\
\textbf{A6.} (Symptom $\rightarrow$ causes $\rightarrow$ Event/Diagnosis/Symptom/Finding; Outcome $\rightarrow$ after $\rightarrow$ Symptom)\\
\textbf{Q7.} By discharge, which symptoms had resolved and which remained active?\\
\textbf{A7.} (Symptom.status at discharge; Outcome.status)\\
\textbf{Q8.} Were any monitoring or follow-up instructions provided for those persistent symptoms?\\
\textbf{A8.} (Symptom $\rightarrow$ has\_instruction $\rightarrow$ Instruction)

\textbf{Template 5 --- Chronic Symptom Burden and Functional Impact Across Admissions}\\[0.15em]
\textbf{Q1.} Was there any chronic or persistent symptom that the patient experienced across multiple admissions?\\
\textbf{A1.} (Symptom present in >=2 notes; Symptom.status = chronic/persistent)\\
\textbf{Q2.} How did the severity or frequency of that symptom change over time?\\
\textbf{A2.} (Symptom.status trend; Symptom.time)\\
\textbf{Q3.} Did the symptom affect the patient's functional status or daily activities?\\
\textbf{A3.} (Symptom $\rightarrow$ resulted\_in $\rightarrow$ Activity; Activity.status)\\
\textbf{Q4.} Were any treatments or interventions used to improve functioning despite the symptom?\\
\textbf{A4.} (Medication/Procedure $\rightarrow$ administered\_for $\rightarrow$ Symptom; Activity.status)\\
\textbf{Q5.} What was the documented clinical reasoning that connected the symptom trajectory to the care plan at the most recent discharge?\\
\textbf{A5.} (clinical reasoning from notes linking the chronic symptom pattern to the discharge care plan; Symptom $\rightarrow$ resulted\_in $\rightarrow$ Outcome)\\
\textbf{Q6.} Did the symptom result in changes in the patient's care needs or living situation?\\
\textbf{A6.} (Symptom $\rightarrow$ resulted\_in $\rightarrow$ Outcome)\\
\textbf{Q7.} Was any long-term management or monitoring plan recommended at the most recent discharge for the symptom?\\
\textbf{A7.} (Symptom $\rightarrow$ has\_instruction $\rightarrow$ Instruction)

\subsection{Procedure}

\subsubsection{Single-note setting}
\textbf{Template 1 --- Procedure Indication, Supporting Evidence, and Discharge Planning}\\[0.15em]
\textbf{Q1.} Did the patient undergo any procedures during admission, and on what date or hospital day were they performed?\\
\textbf{A1.} (Procedure.assertion; Procedure.time; Procedure.is\_main; Procedure.status)\\
\textbf{Q2.} What was the primary indication that led to the procedure?\\
\textbf{A2.} (Procedure $\rightarrow$ administered\_for $\rightarrow$ Diagnosis/Symptom/Finding; Diagnosis.status; Diagnosis.assertion)\\
\textbf{Q3.} Were there any pre-procedural findings that supported the decision to perform the procedure, and were any abnormalities documented?\\
\textbf{A3.} (Lab\_Test/Diagnostic\_Imaging\_Test/Physical\_Exam/Vital\_Sign $\rightarrow$ resulted\_in $\rightarrow$ Procedure; Lab\_Test.value; Lab\_Test.abnormal\_flag; Diagnostic\_Imaging\_Test.result; Vital\_Sign.value)\\
\textbf{Q4.} What was the documented clinical reasoning that connected those abnormalities to the decision to proceed with the intervention?\\
\textbf{A4.} (clinical reasoning from notes linking pre-procedural findings to the procedure decision; Procedure $\rightarrow$ administered\_for $\rightarrow$ Diagnosis/Symptom/Finding)\\
\textbf{Q5.} Were there any intraoperative or immediate post-procedural complications, and if so, how were they identified?\\
\textbf{A5.} (Procedure $\rightarrow$ causes $\rightarrow$ Event/Diagnosis/Symptom/Finding; Lab\_Test/Imaging/Exam/Vitals $\rightarrow$ reveals $\rightarrow$ Diagnosis/Symptom/Finding)\\
\textbf{Q6.} What was the patient's clinical outcome by discharge related to the procedure?\\
\textbf{A6.} (Procedure $\rightarrow$ resulted\_in $\rightarrow$ Outcome/Finding; Outcome.status; Diagnosis/Symptom/Finding.status at discharge)\\
\textbf{Q7.} Were any discharge instructions or follow-up plans provided that were specifically related to the procedure?\\
\textbf{A7.} (Procedure $\rightarrow$ has\_instruction $\rightarrow$ Instruction; time=discharge)

\textbf{Template 2 --- Emergent or Urgent Procedure Decision-Making}\\[0.15em]
\textbf{Q1.} Were there any acute clinical changes or events during admission that led to an urgent or emergent procedure?\\
\textbf{A1.} (Event $\rightarrow$ causes $\rightarrow$ Procedure; Diagnosis/Symptom/Finding $\rightarrow$ causes $\rightarrow$ Procedure; acute Vital\_Sign/Lab\_Test abnormality)\\
\textbf{Q2.} Were any diagnostic findings documented that supported proceeding with the urgent intervention?\\
\textbf{A2.} (Lab\_Test/Imaging/Physical\_Exam $\rightarrow$ resulted\_in $\rightarrow$ Procedure; Lab\_Test.abnormal\_flag; Imaging.result)\\
\textbf{Q3.} What procedure was performed, and when during the admission course?\\
\textbf{A3.} (Procedure.assertion; Procedure.time; Procedure.is\_main)\\
\textbf{Q4.} Were there any intraoperative or immediate postoperative complications?\\
\textbf{A4.} (Procedure $\rightarrow$ causes $\rightarrow$ Diagnosis/Symptom/Finding/Event; Vital\_Sign changes; Lab\_Test changes)\\
\textbf{Q5.} How did the patient's clinical status change after the procedure?\\
\textbf{A5.} (Procedure $\rightarrow$ resulted\_in $\rightarrow$ Outcome/Finding; Vital\_Sign.value; Lab\_Test.value trend; Physical\_Exam.finding changes)\\
\textbf{Q6.} Was any monitoring, rehabilitation, or follow-up planned due to the urgency of the procedure?\\
\textbf{A6.} (Procedure $\rightarrow$ has\_instruction $\rightarrow$ Instruction; Activity.status; time=discharge)

\textbf{Template 3 --- Surgical Approach, Perioperative Care, and Recovery Trajectory}\\[0.15em]
\textbf{Q1.} What surgical procedure was performed, and what were the operative approach or components?\\
\textbf{A1.} (Procedure.assertion; Procedure.time; Procedure.is\_main; Procedure.status)\\
\textbf{Q2.} What condition was the surgery intended to address?\\
\textbf{A2.} (Procedure $\rightarrow$ administered\_for $\rightarrow$ Diagnosis/Symptom/Finding)\\
\textbf{Q3.} What perioperative management strategies were implemented?\\
\textbf{A3.} (Procedure/Medication $\rightarrow$ administered\_for $\rightarrow$ Diagnosis/Symptom/Finding; Medical\_Device.status; Medication.route; Medication.time)\\
\textbf{Q4.} Did the procedure result in any expected postoperative findings or outcomes?\\
\textbf{A4.} (Procedure $\rightarrow$ resulted\_in $\rightarrow$ Outcome/Finding; Outcome $\rightarrow$ after $\rightarrow$ Procedure)\\
\textbf{Q5.} Did any postoperative complication occur, and what was done in response?\\
\textbf{A5.} (Procedure $\rightarrow$ causes $\rightarrow$ Diagnosis/Symptom/Finding/Event; Medication/Procedure $\rightarrow$ administered\_for $\rightarrow$ Diagnosis/Symptom/Finding)\\
\textbf{Q6.} How was the recovery assessed at discharge based on the response to complication management?\\
\textbf{A6.} (Activity.status; Outcome.status; Diagnosis/Symptom/Finding.status at discharge; clinical reasoning linking complication management to discharge status)\\
\textbf{Q7.} What were the patient's wound care, rehabilitation, or procedural precautions at discharge?\\
\textbf{A7.} (Procedure $\rightarrow$ has\_instruction $\rightarrow$ Instruction; Activity $\rightarrow$ has\_instruction $\rightarrow$ Instruction)

\textbf{Template 4 --- Procedure-Related Complication and Management Escalation}\\[0.15em]
\textbf{Q1.} Did the patient undergo a procedure during admission, and was there any sign that a complication had occurred afterward?\\
\textbf{A1.} (Procedure.assertion, Procedure.time; Procedure $\rightarrow$ causes $\rightarrow$ Diagnosis/Symptom/Finding/Event; earliest time evidence)\\
\textbf{Q2.} Was any diagnostic workup done to evaluate the complication, and what did it reveal?\\
\textbf{A2.} (Lab\_Test/Diagnostic\_Imaging\_Test/Physical\_Exam/Vital\_Sign $\rightarrow$ reveals $\rightarrow$ Diagnosis/Finding; Lab\_Test.value, Lab\_Test.abnormal\_flag, Diagnostic\_Imaging\_Test.result)\\
\textbf{Q3.} What was the initial treatment approach for the complication, and was it effective?\\
\textbf{A3.} (Medication/Procedure $\rightarrow$ administered\_for $\rightarrow$ Diagnosis/Symptom/Finding; Medication/Procedure $\rightarrow$ improves/worsens $\rightarrow$ Diagnosis/Symptom/Finding)\\
\textbf{Q4.} If the initial approach was not fully effective, was the management escalated, and what clinical reasoning drove that decision?\\
\textbf{A4.} (Procedure/Medication $\rightarrow$ switched\_to $\rightarrow$ Procedure/Medication; clinical reasoning from notes linking inadequate response to the escalation decision)\\
\textbf{Q5.} What was the final status of the complication at discharge, and were any additional follow-up steps recommended?\\
\textbf{A5.} (Diagnosis/Symptom/Finding.status at discharge; Outcome.status; Procedure/Medication $\rightarrow$ has\_instruction $\rightarrow$ Instruction)

\textbf{Template 5 --- Procedure Outcome, Functional Status Change, and Discharge Planning}\\[0.15em]
\textbf{Q1.} What was the patient's functional or clinical status prior to the procedure that made intervention necessary?\\
\textbf{A1.} (Diagnosis/Symptom/Finding.status pre-procedure; Physical\_Exam.finding; Activity.status pre-admission)\\
\textbf{Q2.} What procedure was ultimately performed, and what were the expected goals?\\
\textbf{A2.} (Procedure.assertion, Procedure.is\_main; Procedure $\rightarrow$ administered\_for $\rightarrow$ Diagnosis/Symptom/Finding; Procedure $\rightarrow$ resulted\_in $\rightarrow$ Outcome/Finding expected)\\
\textbf{Q3.} How did objective measures such as labs, imaging, vitals, or exam change after the procedure?\\
\textbf{A3.} (Lab\_Test.value post-procedure; Vital\_Sign.value; Physical\_Exam.finding; Diagnostic\_Imaging\_Test.result post-procedure)\\
\textbf{Q4.} What was the patient's functional and clinical status at discharge compared to admission?\\
\textbf{A4.} (Outcome.status; Activity.status at discharge; Diagnosis/Symptom/Finding.status resolved/improved/persistent)\\
\textbf{Q5.} What discharge disposition was chosen, and were any ongoing procedure-related instructions or follow-up given?\\
\textbf{A5.} (Procedure $\rightarrow$ has\_instruction $\rightarrow$ Instruction; Activity $\rightarrow$ has\_instruction $\rightarrow$ Instruction; Outcome $\rightarrow$ after $\rightarrow$ Procedure; discharge disposition)

\subsubsection{Multi-note setting}
\textbf{Template 1 --- Procedure Indication, Course, and Discharge Follow-up}\\[0.15em]
\textbf{Q1.} Was any procedure performed during the most recent admission, and if so, when was it performed?\\
\textbf{A1.} (Procedure.assertion, Procedure.time, Procedure.is\_main)\\
\textbf{Q2.} Was there any condition that led clinicians to perform the procedure?\\
\textbf{A2.} (Procedure $\rightarrow$ administered\_for $\rightarrow$ Diagnosis/Symptom/Finding)\\
\textbf{Q3.} Were there any laboratory tests, imaging, physical examination findings, or vital sign abnormalities that supported performing the procedure?\\
\textbf{A3.} (Lab\_Test/Diagnostic\_Imaging\_Test/Physical\_Exam/Vital\_Sign $\rightarrow$ resulted\_in $\rightarrow$ Procedure; Lab\_Test.value; Lab\_Test.abnormal\_flag; Diagnostic\_Imaging\_Test.result; Physical\_Exam.finding; Vital\_Sign.value)\\
\textbf{Q4.} What was the documented clinical reasoning that linked those abnormalities to the decision to proceed with the intervention?\\
\textbf{A4.} (clinical reasoning from notes connecting pre-procedural findings to the procedure decision; Procedure $\rightarrow$ administered\_for $\rightarrow$ Diagnosis/Symptom/Finding)\\
\textbf{Q5.} After the procedure was performed, did it lead to any complication, event, or new condition, and if so, how was that recognized?\\
\textbf{A5.} (Procedure $\rightarrow$ causes $\rightarrow$ Event/Diagnosis/Symptom/Finding; Lab\_Test/Diagnostic\_Imaging\_Test/Vital\_Sign/Physical\_Exam $\rightarrow$ reveals $\rightarrow$ Diagnosis/Symptom/Finding)\\
\textbf{Q6.} Was any treatment or intervention administered to manage the complication?\\
\textbf{A6.} (Procedure/Medication $\rightarrow$ administered\_for $\rightarrow$ Diagnosis/Symptom/Finding; Procedure/Medication $\rightarrow$ resulted\_in $\rightarrow$ Outcome/Finding)\\
\textbf{Q7.} Were any discharge instructions or follow-up plans documented for the procedure or the condition related to it?\\
\textbf{A7.} (Procedure $\rightarrow$ has\_instruction $\rightarrow$ Instruction; Diagnosis/Symptom/Finding $\rightarrow$ has\_instruction $\rightarrow$ Instruction; time=discharge)

\textbf{Template 2 --- Procedure-Related Complication Cascade and Escalation of Care}\\[0.15em]
\textbf{Q1.} At admission $\langle$chartdate or admission\_id$\rangle$, was any procedure performed, and were any complications, events, or abnormal findings observed afterward?\\
\textbf{A1.} (Procedure.time; Procedure $\rightarrow$ causes $\rightarrow$ Event/Diagnosis/Symptom/Finding)\\
\textbf{Q2.} Were any diagnostic evaluations performed to further assess the complication or abnormal finding?\\
\textbf{A2.} (Lab\_Test/Imaging/Physical\_Exam/Vital\_Sign $\rightarrow$ reveals $\rightarrow$ Diagnosis/Symptom/Finding; Lab\_Test.abnormal\_flag; Diagnostic\_Imaging\_Test.result)\\
\textbf{Q3.} Was any initial treatment or management strategy used to address the complication, and did it improve the condition?\\
\textbf{A3.} (Medication/Procedure $\rightarrow$ administered\_for $\rightarrow$ Diagnosis/Symptom/Finding; Medication/Procedure $\rightarrow$ improves/worsens $\rightarrow$ Diagnosis/Symptom/Finding)\\
\textbf{Q4.} If the initial management was not fully effective, was the care escalated, and what clinical reasoning drove that decision?\\
\textbf{A4.} (Procedure/Medication $\rightarrow$ switched\_to $\rightarrow$ Procedure/Medication; Event.status; clinical reasoning from notes linking inadequate response to escalation)\\
\textbf{Q5.} What was the final outcome or clinical status after management of the complication?\\
\textbf{A5.} (Outcome/Finding status; Instruction)

\textbf{Template 3 --- Emergent Procedure Decision-Making and Execution}\\[0.15em]
\textbf{Q1.} Were there any clinical changes or acute events at admission $\langle$chartdate or admission\_id$\rangle$ that necessitated an emergent or urgent procedure?\\
\textbf{A1.} (Event $\rightarrow$ causes $\rightarrow$ Procedure, Diagnosis/Symptom/Finding $\rightarrow$ causes $\rightarrow$ Procedure, acute changes in Vital\_Sign/Lab\_Test)\\
\textbf{Q2.} Were any diagnostic findings documented that supported performing the procedure?\\
\textbf{A2.} (Lab\_Test/Diagnostic\_Imaging\_Test/Physical\_Exam $\rightarrow$ resulted\_in $\rightarrow$ Procedure; Lab\_Test.abnormal\_flag; Diagnostic\_Imaging\_Test.finding)\\
\textbf{Q3.} What procedure was performed to address the clinical change, and when during the hospitalization did it occur?\\
\textbf{A3.} (Procedure.assertion, Procedure.time, Procedure.is\_main)\\
\textbf{Q4.} After the procedure was performed, were any intraoperative or immediate post-procedural complications observed?\\
\textbf{A4.} (Procedure $\rightarrow$ causes $\rightarrow$ Diagnosis/Symptom/Finding/Event)\\
\textbf{Q5.} Were any interventions performed to manage the complications following the procedure?\\
\textbf{A5.} (Medication/Procedure $\rightarrow$ administered\_for $\rightarrow$ Diagnosis/Symptom/Finding)\\
\textbf{Q6.} How did the patient's clinical status change after the procedure?\\
\textbf{A6.} (Procedure $\rightarrow$ resulted\_in $\rightarrow$ Outcome/Finding)\\
\textbf{Q7.} Were any monitoring or follow-up instructions provided for the procedure?\\
\textbf{A7.} (Procedure $\rightarrow$ has\_instruction $\rightarrow$ Instruction)

\textbf{Template 4 --- Repeat Procedures and Recurrence-Driven Reintervention Across Admissions}\\[0.15em]
\textbf{Q1.} During the first admission, was any procedure performed for $\langle$Diagnosis/Symptom/Finding$\rangle$, and what was the immediate post-procedure outcome at discharge?\\
\textbf{A1.} (Procedure.time=admission; Procedure $\rightarrow$ administered\_for $\rightarrow$ Diagnosis/Symptom/Finding; Outcome.status at discharge)\\
\textbf{Q2.} After that procedure, did the condition recur in a later admission?\\
\textbf{A2.} (Diagnosis/Symptom/Finding.status over time)\\
\textbf{Q3.} Were there any tests, examinations, imaging studies, or vital signs that supported the decision to perform another procedure for the recurring condition?\\
\textbf{A3.} (Lab\_Test/Diagnostic\_Imaging\_Test/Physical\_Exam/Vital\_Sign $\rightarrow$ resulted\_in $\rightarrow$ Procedure)\\
\textbf{Q4.} What was the documented clinical reasoning that explained why the condition returned despite the prior intervention?\\
\textbf{A4.} (clinical reasoning from notes linking factors such as incomplete resolution, non-adherence, or disease progression across admissions to the recurrence)\\
\textbf{Q5.} Was a repeat procedure performed, or was a different procedure chosen?\\
\textbf{A5.} (Procedure $\rightarrow$ switched\_to $\rightarrow$ Procedure)\\
\textbf{Q6.} After the procedure, did the patient's clinical outcome improve compared with the outcome after the previous procedure?\\
\textbf{A6.} (Procedure $\rightarrow$ resulted\_in $\rightarrow$ Outcome/Finding; Outcome $\rightarrow$ after $\rightarrow$ Procedure)

\textbf{Template 5 --- Surgical Approach, Anesthesia Management, and Post-Surgical Outcomes}\\[0.15em]
\textbf{Q1.} Was any surgical procedure performed during admission $\langle$chartdate or admission\_id$\rangle$, and what operative approach was used?\\
\textbf{A1.} (Procedure.assertion, Procedure.time, Procedure.is\_main)\\
\textbf{Q2.} Was there any condition that led clinicians to perform the procedure?\\
\textbf{A2.} (Procedure $\rightarrow$ administered\_for $\rightarrow$ Diagnosis/Symptom/Finding)\\
\textbf{Q3.} What was the anesthesia type or anesthesia-related considerations for the surgical procedure?\\
\textbf{A3.} (Event/Diagnosis/Finding related to anesthesia; Vital\_Sign instability)\\
\textbf{Q4.} During the surgery, were there any intraoperative findings or events?\\
\textbf{A4.} (Procedure $\rightarrow$ resulted\_in $\rightarrow$ Outcome/Finding during Procedure; Procedure $\rightarrow$ causes $\rightarrow$ Event/Diagnosis/Finding)\\
\textbf{Q5.} What was the immediate postoperative clinical status after the surgical procedure?\\
\textbf{A5.} (Procedure $\rightarrow$ resulted\_in $\rightarrow$ Outcome/Finding)\\
\textbf{Q6.} Did the surgical procedure result in any postoperative complications?\\
\textbf{A6.} (Procedure $\rightarrow$ causes $\rightarrow$ Diagnosis/Symptom/Finding/Event)\\
\textbf{Q7.} By discharge, how had the patient's condition changed after the procedure?\\
\textbf{A7.} (Outcome.status; Diagnosis/Symptom/Finding.status)\\
\textbf{Q8.} Were any discharge instructions or follow-up plans provided for the procedure?\\
\textbf{A8.} (Procedure $\rightarrow$ has\_instruction $\rightarrow$ Instruction)

\subsection{Medication}

\subsubsection{Single-note setting}
\textbf{Template 1 --- Inpatient Initiation, Dose Adjustment, and Tolerability}\\[0.15em]
\textbf{Q1.} Were any medications initiated during admission to treat $\langle$Diagnosis/Symptom/Finding$\rangle$?\\
\textbf{A1.} (Medication.time = inpatient, Medication $\rightarrow$ administered\_for $\rightarrow$ Diagnosis/Symptom/Finding)\\
\textbf{Q2.} What were the initial dose, route, and frequency?\\
\textbf{A2.} (Medication.strength, dosage, route, frequency, form)\\
\textbf{Q3.} Did the patient develop any side effects or intolerance after initiation?\\
\textbf{A3.} (Medication $\rightarrow$ causes $\rightarrow$ Symptom/Finding/Event, Symptom.status; Lab\_Test.abnormal\_flag if applicable)\\
\textbf{Q4.} Was the regimen modified in response to those adverse effects?\\
\textbf{A4.} (Medication.dosage change with time; Medication $\rightarrow$ switched\_to $\rightarrow$ Medication; clinical reasoning from notes linking the adverse effect to the dose adjustment)\\
\textbf{Q5.} Did the adjusted regimen improve the medical condition?\\
\textbf{A5.} (Medication $\rightarrow$ improves $\rightarrow$ Diagnosis/Symptom/Finding, Medication $\rightarrow$ resulted\_in $\rightarrow$ Outcome)\\
\textbf{Q6.} What was the final discharge medication regimen?\\
\textbf{A6.} (Medication.time = discharge, Medication.strength, dosage, route, frequency, disp, refills)

\textbf{Template 2 --- Diagnostic Findings Driving Medication Adjustment}\\[0.15em]
\textbf{Q1.} What medications was the patient taking on admission?\\
\textbf{A1.} (Medication.time = admission)\\
\textbf{Q2.} Were there any abnormal lab tests, imaging findings, or vital sign changes that prompted medication changes?\\
\textbf{A2.} (Lab\_Test/Diagnostic\_Imaging\_Test/Vital\_Sign $\rightarrow$ resulted\_in $\rightarrow$ Medication change)\\
\textbf{Q3.} What was the documented reasoning for the particular adjustment chosen in response to those abnormal findings?\\
\textbf{A3.} (clinical reasoning from notes linking the specific abnormal values to the treatment decision; Medication.dosage/route/frequency change rationale)\\
\textbf{Q4.} After those medication adjustments were made, did the relevant clinical parameters improve?\\
\textbf{A4.} (Medication $\rightarrow$ improves $\rightarrow$ Diagnosis/Symptom/Finding, Improved Lab\_Test/Vital\_Sign trends)\\
\textbf{Q5.} Were any time-limited medications prescribed?\\
\textbf{A5.} (Medication.instruction including duration, stop date)\\
\textbf{Q6.} Were any monitoring instructions provided at discharge?\\
\textbf{A6.} (Medication $\rightarrow$ has\_instruction $\rightarrow$ Instruction)

\textbf{Template 3 --- Medication Safety, Allergies, and Adverse Effects}\\[0.15em]
\textbf{Q1.} Does the patient have any medication allergies?\\
\textbf{A1.} (Allergy.status, assertion)\\
\textbf{Q2.} Were any medications avoided or modified due to the allergy history?\\
\textbf{A2.} (Medication.assertion with avoidance)\\
\textbf{Q3.} Did the patient experience any adverse medication-related complications during the stay?\\
\textbf{A3.} (Medication $\rightarrow$ causes $\rightarrow$ Diagnosis/Symptom/Event/Finding)\\
\textbf{Q4.} Was a replacement medication selected, and what was the documented reasoning for that choice?\\
\textbf{A4.} (Medication $\rightarrow$ switched\_to $\rightarrow$ Medication; clinical reasoning from notes explaining why the alternative was chosen given the adverse effect and allergy profile)\\
\textbf{Q5.} Did the alternative medication lead to clinical stabilization?\\
\textbf{A5.} (Medication $\rightarrow$ improves $\rightarrow$ Diagnosis/Symptom/Finding, Medication $\rightarrow$ resulted\_in $\rightarrow$ Outcome)\\
\textbf{Q6.} Were any safety instructions provided regarding medication use?\\
\textbf{A6.} (Medication $\rightarrow$ has\_instruction $\rightarrow$ Instruction)

\textbf{Template 4 --- Discharge Medication Reconciliation and Patient Instructions}\\[0.15em]
\textbf{Q1.} What medications were prescribed at discharge?\\
\textbf{A1.} (Medication.time = discharge; Medication.dosage, Medication.route, Medication.frequency, Medication.disp, Medication.refills)\\
\textbf{Q2.} Which of those medications were newly started during admission, and which were continued from before?\\
\textbf{A2.} (Medication.assertion; Medication.time comparison)\\
\textbf{Q3.} What condition was each newly started medication intended to treat?\\
\textbf{A3.} (Medication $\rightarrow$ administered\_for $\rightarrow$ Diagnosis/Symptom/Finding)\\
\textbf{Q4.} Were any pre-admission medications discontinued at discharge, and if so, why?\\
\textbf{A4.} (Medication.assertion = discontinued; Medication $\rightarrow$ causes $\rightarrow$ Diagnosis/Symptom/Finding; Allergy.status)\\
\textbf{Q5.} Were any instructions given to the patient regarding the discharge medications?\\
\textbf{A5.} (Medication $\rightarrow$ has\_instruction $\rightarrow$ Instruction)

\textbf{Template 5 --- Medication for Complication Management During Hospitalization}\\[0.15em]
\textbf{Q1.} Did the patient develop any complications or new conditions during the hospital stay?\\
\textbf{A1.} (Diagnosis/Symptom/Finding.time = inpatient; Procedure/Medication $\rightarrow$ causes $\rightarrow$ Diagnosis/Symptom/Finding)\\
\textbf{Q2.} Were any medications started to manage those complications?\\
\textbf{A2.} (Medication.time; Medication $\rightarrow$ administered\_for $\rightarrow$ Diagnosis/Symptom/Finding)\\
\textbf{Q3.} What was the documented reasoning for choosing that treatment approach for the complication?\\
\textbf{A3.} (clinical reasoning from notes linking the complication characteristics to the treatment choice; Medication $\rightarrow$ administered\_for $\rightarrow$ Diagnosis/Symptom/Finding)\\
\textbf{Q4.} Did the treatment improve the patient's condition?\\
\textbf{A4.} (Medication $\rightarrow$ improves $\rightarrow$ Diagnosis/Symptom/Finding; Medication $\rightarrow$ resulted\_in $\rightarrow$ Outcome)\\
\textbf{Q5.} Were the complication-related medications continued or stopped at discharge?\\
\textbf{A5.} (Medication.time = discharge; Medication.assertion)

\subsubsection{Multi-note setting}
\textbf{Template 1 --- Medication Intolerance, Adverse Effects, and Regimen Switching}\\[0.15em]
\textbf{Q1.} Was any medication initiated to treat $\langle$Diagnosis/Symptom/Finding$\rangle$ during admission $\langle$chartdate or admission\_id$\rangle$?\\
\textbf{A1.} (Medication.time; Medication $\rightarrow$ administered\_for $\rightarrow$ Diagnosis/Symptom/Finding)\\
\textbf{Q2.} After that medication was started, did the patient develop any adverse symptoms, complications, or clinical events?\\
\textbf{A2.} (Medication $\rightarrow$ causes $\rightarrow$ Diagnosis/Symptom/Event/Finding)\\
\textbf{Q3.} Were any clinical findings or diagnostic abnormalities documented that suggested the adverse effect was related to the medication?\\
\textbf{A3.} (Symptom.status; Lab\_Test.abnormal\_flag = abnormal; Medication $\rightarrow$ causes $\rightarrow$ Event)\\
\textbf{Q4.} Was the medication switched in response, and if so, what was the documented reasoning for choosing the replacement?\\
\textbf{A4.} (Medication.status; Medication $\rightarrow$ switched\_to $\rightarrow$ Medication/Procedure; clinical reasoning from notes explaining the rationale for the alternative selection)\\
\textbf{Q5.} After the alternative treatment was started, did the patient's condition improve without recurrence of the adverse effect?\\
\textbf{A5.} (Medication/Procedure $\rightarrow$ improves $\rightarrow$ Diagnosis/Symptom/Finding; Medication/Procedure $\rightarrow$ resulted\_in $\rightarrow$ Outcome)\\
\textbf{Q6.} Which medication regimen was continued at discharge for management of the patient's condition?\\
\textbf{A6.} (Medication.time = discharge)

\textbf{Template 2 --- Diagnostic Evidence Driving Medication Adjustment}\\[0.15em]
\textbf{Q1.} What medications was the patient taking at the time of admission $\langle$chartdate or admission\_id$\rangle$?\\
\textbf{A1.} (Medication.time = admission)\\
\textbf{Q2.} While the patient was receiving those medications, were there any abnormal laboratory results, imaging findings, physical examination findings, or vital sign changes that prompted clinicians to adjust treatment?\\
\textbf{A2.} (Lab\_Test/Diagnostic\_Imaging\_Test/Vital\_Sign/Physical\_Exam $\rightarrow$ resulted\_in $\rightarrow$ Medication change)\\
\textbf{Q3.} Were the dose, frequency, or route modified in response to those abnormal findings?\\
\textbf{A3.} (Medication.dosage change; Medication.frequency change; Medication.route change)\\
\textbf{Q4.} After those medication adjustments were made, did the relevant clinical parameters improve?\\
\textbf{A4.} (Medication $\rightarrow$ improves $\rightarrow$ Diagnosis/Symptom/Finding; Lab\_Test/Vital\_Sign trend improvement)\\
\textbf{Q5.} At discharge, what medications and dosing regimens were prescribed following those adjustments?\\
\textbf{A5.} (Medication.time = discharge; Medication.dosage/frequency)\\
\textbf{Q6.} Were any monitoring instructions provided related to those medications?\\
\textbf{A6.} (Medication $\rightarrow$ has\_instruction $\rightarrow$ Instruction)

\textbf{Template 3 --- Discharge Medication Reconciliation and Allergy Safety Decisions}\\[0.15em]
\textbf{Q1.} What medications were prescribed at discharge during admission $\langle$chartdate or admission\_id$\rangle$?\\
\textbf{A1.} (Medication.time = discharge)\\
\textbf{Q2.} At the next admission discharge, were any of those medications added, discontinued, or changed?\\
\textbf{A2.} (Medication.time = discharge comparison; Medication.status changes)\\
\textbf{Q3.} Were there any conditions or clinical findings that prompted clinicians to add those new medications?\\
\textbf{A3.} (Medication $\rightarrow$ administered\_for $\rightarrow$ Diagnosis/Symptom/Finding)\\
\textbf{Q4.} What was the documented reasoning that explained the evolution of the medication regimen from one discharge to the next?\\
\textbf{A4.} (clinical reasoning from notes linking cross-admission clinical changes to the medication reconciliation decisions; Allergy.status; Medication $\rightarrow$ causes $\rightarrow$ Event)\\
\textbf{Q5.} Were the doses or frequencies of those medications modified to improve safety or adherence?\\
\textbf{A5.} (Medication.dosage comparison; Medication.frequency comparison)\\
\textbf{Q6.} At the final discharge, were any medication continuation or monitoring instructions provided?\\
\textbf{A6.} (Medication $\rightarrow$ has\_instruction $\rightarrow$ Instruction)

\textbf{Template 4 --- Medication Adherence, Outpatient Failure, and Readmission Risk}\\[0.15em]
\textbf{Q1.} What medications was the patient discharged on during admission $\langle$chartdate or admission\_id$\rangle$?\\
\textbf{A1.} (Medication.time = discharge)\\
\textbf{Q2.} Before the next hospitalization, did the patient stop taking or inconsistently take any of those medications?\\
\textbf{A2.} (Medication.status = non-adherence)\\
\textbf{Q3.} Did that medication non-adherence precede worsening symptoms or acute clinical events?\\
\textbf{A3.} (Medication.status = non-adherence; Medication $\rightarrow$ causes $\rightarrow$ Event/Diagnosis/Symptom/Finding)\\
\textbf{Q4.} How did clinicians address the adherence issue during readmission?\\
\textbf{A4.} (Medication.time; Medication $\rightarrow$ switched\_to $\rightarrow$ Medication; clinical reasoning from notes linking non-adherence pattern to the revised treatment approach)\\
\textbf{Q5.} After medication adherence was restored or treatment was adjusted, did the patient's clinical status improve?\\
\textbf{A5.} (Medication $\rightarrow$ improves $\rightarrow$ Diagnosis/Symptom/Finding; Outcome.status)\\
\textbf{Q6.} Were any adherence counseling or follow-up instructions provided at discharge to prevent recurrence?\\
\textbf{A6.} (Medication $\rightarrow$ has\_instruction $\rightarrow$ Instruction)

\textbf{Template 5 --- Inpatient Medication Adjustment and Discharge Treatment Planning}\\[0.15em]
\textbf{Q1.} What medications was the patient taking on admission $\langle$chartdate or admission\_id$\rangle$?\\
\textbf{A1.} (Medication.time = admission; Medication.assertion)\\
\textbf{Q2.} During hospitalization, were there any changes to the dose, route, or frequency of those medications?\\
\textbf{A2.} (Medication.dosage change; Medication.route change; Medication.frequency change)\\
\textbf{Q3.} Were there any diagnostic findings or vital sign abnormalities that prompted clinicians to make those medication changes?\\
\textbf{A3.} (Lab\_Test/Diagnostic\_Imaging\_Test/Physical\_Exam/Vital\_Sign $\rightarrow$ resulted\_in $\rightarrow$ Medication)\\
\textbf{Q4.} After those medication adjustments were implemented, how did the patient clinically respond?\\
\textbf{A4.} (Medication $\rightarrow$ improves/worsens $\rightarrow$ Diagnosis/Symptom/Finding; Medication $\rightarrow$ resulted\_in $\rightarrow$ Outcome/Finding)\\
\textbf{Q5.} Based on that response, what medications were ultimately prescribed at discharge?\\
\textbf{A5.} (Medication.time = discharge)\\
\textbf{Q6.} For the medications prescribed at discharge, what conditions were they intended to treat?\\
\textbf{A6.} (Medication $\rightarrow$ administered\_for $\rightarrow$ Diagnosis/Symptom/Finding)\\
\textbf{Q7.} Were any discharge instructions given regarding those medications?\\
\textbf{A7.} (Medication $\rightarrow$ has\_instruction $\rightarrow$ Instruction)

\subsection{Microbiology}

\subsubsection{Single-note setting}
\textbf{Template 1 --- Empiric Antibiotic Initiation for Suspected Infection}\\[0.15em]
\textbf{Q1.} Was any infection suspected or diagnosed during admission?\\
\textbf{A1.} (Diagnosis.assertion, Diagnosis.status, Diagnosis.time)\\
\textbf{Q2.} Were any specimens collected to work up the suspected infection?\\
\textbf{A2.} (Specimen.source, Specimen.assertion; Specimen $\rightarrow$ has\_test $\rightarrow$ Microbiology\_Test/Lab\_Test)\\
\textbf{Q3.} Was any empiric therapy started, and for what indication?\\
\textbf{A3.} (Medication.assertion, Medication.time; Medication $\rightarrow$ administered\_for $\rightarrow$ Diagnosis)\\
\textbf{Q4.} What was the documented reasoning for choosing that particular empiric regimen?\\
\textbf{A4.} (clinical reasoning from notes linking the suspected infection type, patient risk factors, and documented rationale to the empiric therapy selection)\\
\textbf{Q5.} Was the empiric treatment continued, changed, or stopped by discharge?\\
\textbf{A5.} (Medication.status at discharge; Medication $\rightarrow$ switched\_to $\rightarrow$ Medication)\\
\textbf{Q6.} What treatment regimen was the patient discharged on, and for how long?\\
\textbf{A6.} (Medication.time, Medication.instruction, Medication.duration at discharge)

\textbf{Template 2 --- Culture Result Driving Therapy Modification}\\[0.15em]
\textbf{Q1.} Were any specimens sent for microbiologic testing during admission, and if so, when?\\
\textbf{A1.} (Specimen.source, Specimen.time; Specimen $\rightarrow$ has\_test $\rightarrow$ Microbiology\_Test)\\
\textbf{Q2.} Were any organisms identified from those cultures?\\
\textbf{A2.} (Microbiology\_Test $\rightarrow$ detected $\rightarrow$ Microbiology\_Organism; Microbiology\_Organism.growth, Microbiology\_Test.result)\\
\textbf{Q3.} Were any susceptibility or resistance patterns reported for the organism?\\
\textbf{A3.} (Microbiology\_Organism $\rightarrow$ tested\_against $\rightarrow$ Microbiology\_Antibiotics; Microbiology\_Antibiotics.sensitivity, Microbiology\_Antibiotics.dilution)\\
\textbf{Q4.} Was the treatment regimen adjusted based on those susceptibility results?\\
\textbf{A4.} (Medication $\rightarrow$ switched\_to $\rightarrow$ Medication; Medication.assertion, Medication.time; clinical reasoning from notes linking susceptibility findings to therapy change)\\
\textbf{Q5.} Did the patient's infection-related signs and symptoms improve after the adjustment?\\
\textbf{A5.} (Medication $\rightarrow$ improves $\rightarrow$ Diagnosis/Symptom/Finding; Vital\_Sign/Lab\_Test trends; Outcome.status)

\textbf{Template 3 --- Culture-Negative Infection Workup and Clinical Decision-Making}\\[0.15em]
\textbf{Q1.} Was an infection suspected during admission, and what was the clinical basis?\\
\textbf{A1.} (Diagnosis.assertion; Symptom.assertion; Physical\_Exam/Vital\_Sign/Lab\_Test $\rightarrow$ reveals $\rightarrow$ Diagnosis/Finding)\\
\textbf{Q2.} What microbiology tests were performed, and what were the results?\\
\textbf{A2.} (Specimen $\rightarrow$ has\_test $\rightarrow$ Microbiology\_Test; Microbiology\_Test.result, Microbiology\_Organism.growth)\\
\textbf{Q3.} Since those cultures did not identify an organism, was there other clinical evidence supporting or refuting the infection?\\
\textbf{A3.} (Lab\_Test/Diagnostic\_Imaging\_Test/Vital\_Sign $\rightarrow$ reveals $\rightarrow$ Diagnosis/Finding; Lab\_Test.value, Lab\_Test.abnormal\_flag)\\
\textbf{Q4.} What was the documented clinical reasoning for the decision to continue or discontinue treatment in the setting of those inconclusive results?\\
\textbf{A4.} (Medication.status, Medication.assertion; clinical reasoning from notes explaining the treatment decision in the setting of negative cultures)\\
\textbf{Q5.} What was the patient's clinical status at discharge, and was any follow-up for the suspected infection recommended?\\
\textbf{A5.} (Outcome.status; Diagnosis/Medication $\rightarrow$ has\_instruction $\rightarrow$ Instruction)

\textbf{Template 4 --- Infection as a Complication of Primary Admission}\\[0.15em]
\textbf{Q1.} What was the primary reason for this patient's admission?\\
\textbf{A1.} (Diagnosis.is\_main, Diagnosis.assertion; Event.status)\\
\textbf{Q2.} Did the patient develop a new infection during the hospitalization as a complication?\\
\textbf{A2.} (Diagnosis.assertion, Diagnosis.time; Event $\rightarrow$ causes $\rightarrow$ Diagnosis; Procedure $\rightarrow$ causes $\rightarrow$ Diagnosis)\\
\textbf{Q3.} Were any specimens collected and what microbiologic testing was performed for the new infection?\\
\textbf{A3.} (Specimen.source, Specimen.time; Specimen $\rightarrow$ has\_test $\rightarrow$ Microbiology\_Test; Microbiology\_Test.result)\\
\textbf{Q4.} Was any treatment started for the infection, and how did it affect the patient's clinical course?\\
\textbf{A4.} (Medication $\rightarrow$ administered\_for $\rightarrow$ Diagnosis; Medication $\rightarrow$ improves $\rightarrow$ Diagnosis/Symptom/Finding; Outcome.status)\\
\textbf{Q5.} Was any follow-up imaging, laboratory testing, or repeat culture recommended at discharge to confirm resolution?\\
\textbf{A5.} (Diagnostic\_Imaging\_Test/Lab\_Test/Microbiology\_Test $\rightarrow$ has\_instruction $\rightarrow$ Instruction; Instruction.instruction\_text)

\textbf{Template 5 --- Monitoring Microbiologic Clearance}\\[0.15em]
\textbf{Q1.} Was any infection identified during admission, and was an organism detected?\\
\textbf{A1.} (Diagnosis.assertion; Microbiology\_Test $\rightarrow$ detected $\rightarrow$ Microbiology\_Organism)\\
\textbf{Q2.} Was any treatment started, and when?\\
\textbf{A2.} (Medication.assertion/time; Medication $\rightarrow$ administered\_for $\rightarrow$ Diagnosis)\\
\textbf{Q3.} Were any follow-up microbiology tests performed?\\
\textbf{A3.} (Microbiology\_Test.time after initial test)\\
\textbf{Q4.} Did follow-up testing show persistent organism growth or microbiologic clearance?\\
\textbf{A4.} (Microbiology\_Test.organism\_growth/result; Diagnosis.status)\\
\textbf{Q5.} How were the duration and continuation of therapy determined based on those follow-up results?\\
\textbf{A5.} (Medication.duration/status; clinical reasoning from notes linking clearance/persistence findings to the treatment duration decision)\\
\textbf{Q6.} What was the infection status and treatment plan at discharge?\\
\textbf{A6.} (Outcome.status; Medication.time = discharge; Diagnosis/Medication $\rightarrow$ has\_instruction $\rightarrow$ Instruction)

\subsubsection{Multi-note setting}
\textbf{Template 1 --- Infection Workup from Suspicion to Targeted Therapy}\\[0.15em]
\textbf{Q1.} During admission $\langle$chartdate or admission\_id$\rangle$, was any infection suspected or diagnosed?\\
\textbf{A1.} (Diagnosis.assertion, Diagnosis.time; Event.status)\\
\textbf{Q2.} Was any specimen collected for microbiologic testing to evaluate the infection?\\
\textbf{A2.} (Specimen.source; Specimen $\rightarrow$ has\_test $\rightarrow$ Microbiology\_Test)\\
\textbf{Q3.} From testing of the specimen, was any organism identified?\\
\textbf{A3.} (Microbiology\_Test $\rightarrow$ detected $\rightarrow$ Microbiology\_Organism; Microbiology\_Test.organism\_growth/result)\\
\textbf{Q4.} Before identification of the organism, was any empiric therapy started for the infection?\\
\textbf{A4.} (Medication.assertion; Medication.time; Medication $\rightarrow$ administered\_for $\rightarrow$ Diagnosis)\\
\textbf{Q5.} After susceptibility results for the organism were available, was the treatment changed or narrowed?\\
\textbf{A5.} (Microbiology\_Organism $\rightarrow$ tested\_against $\rightarrow$ Microbiology\_Antibiotics; Medication $\rightarrow$ switched\_to $\rightarrow$ Medication)\\
\textbf{Q6.} What was the documented reasoning for selecting the targeted regimen based on those susceptibility results?\\
\textbf{A6.} (clinical reasoning from notes linking the susceptibility pattern to the targeted therapy choice; Medication $\rightarrow$ administered\_for $\rightarrow$ Diagnosis)\\
\textbf{Q7.} By discharge, did the infection improve based on clinical or laboratory markers?\\
\textbf{A7.} (Medication $\rightarrow$ improves $\rightarrow$ Diagnosis/Symptom/Finding; Lab\_Test trends; Vital\_Sign trends; Outcome.status)

\textbf{Template 2 --- Culture-Negative vs Culture-Positive Infection Resolution}\\[0.15em]
\textbf{Q1.} During admission $\langle$chartdate or admission\_id$\rangle$, was any infection clinically suspected even before microbiologic confirmation?\\
\textbf{A1.} (Diagnosis.assertion; clinical suspicion)\\
\textbf{Q2.} Were any microbiologic tests performed to evaluate the infection, and what were their results?\\
\textbf{A2.} (Specimen $\rightarrow$ has\_test $\rightarrow$ Microbiology\_Test; Microbiology\_Test.result)\\
\textbf{Q3.} If those tests did not confirm an organism, was there other clinical evidence supporting the infection?\\
\textbf{A3.} (Lab\_Test/Vital\_Sign/Diagnostic\_Imaging\_Test/Physical\_Exam $\rightarrow$ reveals $\rightarrow$ Diagnosis/Finding)\\
\textbf{Q4.} Based on the diagnostic evaluation, was the treatment continued, stopped, or modified?\\
\textbf{A4.} (Medication.status; Medication $\rightarrow$ switched\_to $\rightarrow$ Medication; Medication $\rightarrow$ administered\_for $\rightarrow$ Diagnosis)\\
\textbf{Q5.} After the management decision, did the patient's clinical status improve?\\
\textbf{A5.} (Outcome.status; Symptom.status; Vital\_Sign/Lab improvement)\\
\textbf{Q6.} Was any follow-up or monitoring plan documented at discharge for the infection?\\
\textbf{A6.} (Diagnosis $\rightarrow$ has\_instruction $\rightarrow$ Instruction; Medication $\rightarrow$ has\_instruction $\rightarrow$ Instruction)

\textbf{Template 3 --- Antibiotic Resistance-Driven Therapy Escalation}\\[0.15em]
\textbf{Q1.} During admission $\langle$chartdate or admission\_id$\rangle$, was any organism identified from microbiologic testing?\\
\textbf{A1.} (Microbiology\_Test $\rightarrow$ detected $\rightarrow$ Microbiology\_Organism)\\
\textbf{Q2.} For the infection caused by the organism, was any treatment initially started?\\
\textbf{A2.} (Medication $\rightarrow$ administered\_for $\rightarrow$ Diagnosis; Medication.time)\\
\textbf{Q3.} Were any susceptibility or resistance findings reported for the organism?\\
\textbf{A3.} (Microbiology\_Organism $\rightarrow$ tested\_against $\rightarrow$ Microbiology\_Antibiotics; Microbiology\_Antibiotics.sensitivity; Microbiology\_Antibiotics.dilution)\\
\textbf{Q4.} Was the treatment modified or escalated based on those susceptibility findings?\\
\textbf{A4.} (Medication $\rightarrow$ switched\_to $\rightarrow$ Medication; Medication.status; clinical reasoning from notes linking resistance findings to the therapy escalation)\\
\textbf{Q5.} After the therapy adjustment, did the infection markers or symptoms improve?\\
\textbf{A5.} (Medication $\rightarrow$ improves $\rightarrow$ Diagnosis/Symptom/Finding; Outcome.status)\\
\textbf{Q6.} What treatment regimen was continued at discharge for the infection?\\
\textbf{A6.} (Medication.time = discharge)

\textbf{Template 4 --- Recurrent Infection and Organism Persistence Across Admissions}\\[0.15em]
\textbf{Q1.} During the first admission, was any infection diagnosed?\\
\textbf{A1.} (Diagnosis.assertion; Diagnosis.time = admission\_1)\\
\textbf{Q2.} For the infection, was any organism identified from microbiologic testing?\\
\textbf{A2.} (Microbiology\_Test $\rightarrow$ detected $\rightarrow$ Microbiology\_Organism)\\
\textbf{Q3.} After treatment of the infection, did the same condition recur during a later admission?\\
\textbf{A3.} (Diagnosis.status recurrence; Diagnosis.time comparison)\\
\textbf{Q4.} During the later episode of the infection, was the same organism detected again or was a different organism identified?\\
\textbf{A4.} (Microbiology\_Test $\rightarrow$ detected $\rightarrow$ Microbiology\_Organism organism comparison)\\
\textbf{Q5.} How did the treatment approach change compared with the earlier episode?\\
\textbf{A5.} (Medication/Procedure $\rightarrow$ administered\_for $\rightarrow$ Diagnosis; Medication $\rightarrow$ switched\_to $\rightarrow$ Medication; clinical reasoning from notes linking prior treatment outcome and organism persistence to the revised approach)\\
\textbf{Q6.} What was the clinical outcome after management of the recurrent infection?\\
\textbf{A6.} (Outcome.status)

\textbf{Template 5 --- Monitoring Microbiologic Clearance and Treatment Duration Decisions}\\[0.15em]
\textbf{Q1.} During admission $\langle$chartdate or admission\_id$\rangle$, was any infection identified, and was any organism detected?\\
\textbf{A1.} (Diagnosis.assertion; Microbiology\_Test $\rightarrow$ detected $\rightarrow$ Microbiology\_Organism)\\
\textbf{Q2.} For treatment of the infection, what therapy was started, and when?\\
\textbf{A2.} (Medication.assertion; Medication.time; Medication $\rightarrow$ administered\_for $\rightarrow$ Diagnosis)\\
\textbf{Q3.} After the therapy was initiated, were any follow-up microbiologic tests performed to evaluate treatment response?\\
\textbf{A3.} (Microbiology\_Test.time after Medication)\\
\textbf{Q4.} Did results of those follow-up tests indicate persistence of infection or microbiologic clearance?\\
\textbf{A4.} (Microbiology\_Test.result; Diagnosis.status)\\
\textbf{Q5.} Did those follow-up findings affect continuation or duration of therapy?\\
\textbf{A5.} (Medication.duration; Medication.status)\\
\textbf{Q6.} Was any discharge plan or outpatient monitoring recommended to ensure resolution of the infection?\\
\textbf{A6.} (Diagnosis $\rightarrow$ has\_instruction $\rightarrow$ Instruction; Medication $\rightarrow$ has\_instruction $\rightarrow$ Instruction)

\subsection{Clinical Assessment}

\subsubsection{Single-note setting}
\textbf{Template 1 --- Clinical Assessment Used to Rule Out Competing Diagnoses}\\[0.15em]
\textbf{Q1.} Were any diagnoses initially suspected based on presenting clinical assessments?\\
\textbf{A1.} (Diagnosis.status; Diagnosis.assertion; time = admission)\\
\textbf{Q2.} What assessments were performed to evaluate those suspected conditions?\\
\textbf{A2.} (Lab\_Test; Diagnostic\_Imaging\_Test; Physical\_Exam; Vital\_Sign; time)\\
\textbf{Q3.} Did any of those assessment findings rule out a suspected condition?\\
\textbf{A3.} (Lab\_Test / Diagnostic\_Imaging\_Test / Vital\_Sign / Physical\_Exam $\rightarrow$ reveals $\rightarrow$ Diagnosis with negative assertion)\\
\textbf{Q4.} After those exclusionary results, were any findings documented that confirmed the final diagnosis?\\
\textbf{A4.} (Lab\_Test / Diagnostic\_Imaging\_Test / Vital\_Sign / Physical\_Exam $\rightarrow$ reveals $\rightarrow$ Diagnosis; clinical reasoning from notes linking the positive findings to the confirmed diagnosis)\\
\textbf{Q5.} Was any treatment decision made based on the clarified diagnosis?\\
\textbf{A5.} (Procedure/Medication $\rightarrow$ administered\_for $\rightarrow$ Diagnosis)\\
\textbf{Q6.} Were any assessment findings at discharge that supported clinical stability?\\
\textbf{A6.} (Lab\_Test.value; Vital\_Sign.status; Physical\_Exam.finding; Diagnostic\_Imaging\_Test.result; Outcome.status; time = discharge)

\textbf{Template 2 --- Assessment-Guided Monitoring and Treatment Adjustment}\\[0.15em]
\textbf{Q1.} Were there any assessment findings that prompted initiation or adjustment of treatment during admission?\\
\textbf{A1.} (Lab\_Test / Vital\_Sign / Physical\_Exam / Diagnostic\_Imaging\_Test $\rightarrow$ resulted\_in $\rightarrow$ Medication/Procedure)\\
\textbf{Q2.} How frequently were those assessments repeated?\\
\textbf{A2.} (Lab\_Test.frequency; Vital\_Sign.frequency; Physical\_Exam.frequency; time intervals during admission)\\
\textbf{Q3.} Did repeat assessments show improvement after treatment?\\
\textbf{A3.} (Medication/Procedure $\rightarrow$ resulted\_in $\rightarrow$ Outcome/Finding; Lab\_Test.value normalization; Vital\_Sign.status improvement)\\
\textbf{Q4.} Were there any assessment findings that worsened despite treatment?\\
\textbf{A4.} (Medication/Procedure $\rightarrow$ worsens $\rightarrow$ Diagnosis/Symptom/Finding OR Outcome/Finding $\rightarrow$ after $\rightarrow$ Medication/Procedure)\\
\textbf{Q5.} Did those findings improve, and if not, was the treatment plan modified in response?\\
\textbf{A5.} (Medication/Procedure $\rightarrow$ switched\_to $\rightarrow$ Medication/Procedure; clinical reasoning from notes linking the worsening findings to the specific treatment modification)\\
\textbf{Q6.} What were the clinical assessment values at discharge?\\
\textbf{A6.} (Lab\_Test.value; Vital\_Sign.value/status; Physical\_Exam.finding; Diagnostic\_Imaging\_Test.result; Outcome.status; time = discharge)

\textbf{Template 3 --- Complication Detection Through Clinical Assessment}\\[0.15em]
\textbf{Q1.} Were there any new abnormal clinical findings during hospitalization?\\
\textbf{A1.} (Lab\_Test.abnormal\_flag; Vital\_Sign.status; Physical\_Exam.finding; Diagnostic\_Imaging\_Test.result; time)\\
\textbf{Q2.} Did any procedure or medication cause the new abnormal finding?\\
\textbf{A2.} (Procedure/Medication $\rightarrow$ causes $\rightarrow$ Diagnosis/Symptom/Finding)\\
\textbf{Q3.} Were any assessments used to confirm the complication?\\
\textbf{A3.} (Lab\_Test / Diagnostic\_Imaging\_Test / Vital\_Sign / Physical\_Exam $\rightarrow$ reveals $\rightarrow$ Diagnosis/Finding)\\
\textbf{Q4.} Was any treatment initiated to manage the complication?\\
\textbf{A4.} (Procedure/Medication $\rightarrow$ administered\_for $\rightarrow$ Diagnosis/Symptom/Finding)\\
\textbf{Q5.} Did follow-up assessments show resolution of the complication?\\
\textbf{A5.} (Medication/Procedure $\rightarrow$ resulted\_in $\rightarrow$ Outcome/Finding; Lab\_Test.value trend; Vital\_Sign improvement)\\
\textbf{Q6.} What was the patient's final assessment status at discharge?\\
\textbf{A6.} (Outcome.status; Lab\_Test.value; Vital\_Sign.status; Physical\_Exam.finding; Diagnostic\_Imaging\_Test.result)

\textbf{Template 4 --- Discharge Assessment and Follow-Up Planning Based on Residual Abnormalities}\\[0.15em]
\textbf{Q1.} What were the patient's clinical assessment findings on the day of discharge?\\
\textbf{A1.} (Lab\_Test.value; Lab\_Test.abnormal\_flag; Vital\_Sign.status/value; Physical\_Exam.finding; Diagnostic\_Imaging\_Test.result; time = discharge)\\
\textbf{Q2.} Were any abnormalities still present at discharge?\\
\textbf{A2.} (Lab\_Test.abnormal\_flag; Vital\_Sign.status; Physical\_Exam.finding; Diagnostic\_Imaging\_Test.result)\\
\textbf{Q3.} What was the documented clinical reasoning for managing those persistent abnormalities as outpatient?\\
\textbf{A3.} (clinical reasoning from notes explaining why persistent findings did not preclude discharge; Diagnosis/Symptom/Finding $\rightarrow$ has\_instruction $\rightarrow$ Instruction)\\
\textbf{Q4.} Were any follow-up tests or imaging recommended based on those assessment findings?\\
\textbf{A4.} (Diagnosis/Symptom/Finding $\rightarrow$ resulted\_in $\rightarrow$ Lab\_Test/Diagnostic\_Imaging\_Test)\\
\textbf{Q5.} What was the overall clinical outcome based on the final assessments?\\
\textbf{A5.} (Outcome.status; Outcome/Finding $\rightarrow$ after $\rightarrow$ Procedure/Medication)

\textbf{Template 5 --- Treatment Adjustment Guided by Reassessment}\\[0.15em]
\textbf{Q1.} What was the patient's initial clinical status at admission based on exam findings and lab/imaging results?\\
\textbf{A1.} (Physical\_Exam.finding, Lab\_Test.value/abnormal\_flag, Vital\_Sign.value/status, Diagnostic\_Imaging\_Test.result; time = admission)\\
\textbf{Q2.} Was any treatment started initially, and what was it targeting?\\
\textbf{A2.} (Medication/Procedure $\rightarrow$ administered\_for $\rightarrow$ Diagnosis/Symptom/Finding)\\
\textbf{Q3.} Were there any assessment findings during the hospital course that indicated the initial treatment was insufficient or causing adverse effects?\\
\textbf{A3.} (Medication/Procedure $\rightarrow$ worsens $\rightarrow$ Diagnosis/Symptom/Finding; Medication/Procedure $\rightarrow$ causes $\rightarrow$ Diagnosis/Symptom/Finding; Lab\_Test/Vital\_Sign trend worsening)\\
\textbf{Q4.} Did the reassessment findings show adequate response, and if not, were any changes to treatment made?\\
\textbf{A4.} (Medication/Procedure $\rightarrow$ switched\_to $\rightarrow$ Medication/Procedure; Lab\_Test/Vital\_Sign/Physical\_Exam $\rightarrow$ resulted\_in $\rightarrow$ Medication/Procedure; clinical reasoning linking reassessment to treatment change)\\
\textbf{Q5.} Did the treatment modification lead to clinical improvement on reassessment?\\
\textbf{A5.} (Medication/Procedure $\rightarrow$ resulted\_in $\rightarrow$ Outcome/Finding; Lab\_Test.value normalization; Vital\_Sign.value improvement)\\
\textbf{Q6.} What were the assessment results at discharge after treatment was optimized?\\
\textbf{A6.} (Lab\_Test.value, Vital\_Sign.value, Physical\_Exam.finding at discharge; Outcome.status)

\subsubsection{Multi-note setting}
\textbf{Template 1 --- Assessment Findings Triggering Treatment Initiation and Adjustment}\\[0.15em]
\textbf{Q1.} Were there any abnormal clinical assessments at admission $\langle$chartdate or admission\_id$\rangle$? If so, what were they?\\
\textbf{A1.} (Lab\_Test.value/abnormal\_flag; Vital\_Sign.value/status; Physical\_Exam.finding; Diagnostic\_Imaging\_Test.result)\\
\textbf{Q2.} Did those findings point to any clinical condition?\\
\textbf{A2.} (Lab\_Test/Diagnostic\_Imaging\_Test/Physical\_Exam/Vital\_Sign $\rightarrow$ reveals $\rightarrow$ Diagnosis/Symptom/Finding)\\
\textbf{Q3.} Was any treatment or procedure started to manage that condition?\\
\textbf{A3.} (Medication/Procedure $\rightarrow$ administered\_for $\rightarrow$ Diagnosis/Symptom/Finding)\\
\textbf{Q4.} After the treatment was initiated, were the assessments repeated, and if so, what trends were observed?\\
\textbf{A4.} (Lab\_Test.value/abnormal\_flag trends; Vital\_Sign.value/status trends; Physical\_Exam.finding trends; Diagnostic\_Imaging\_Test.result trends)\\
\textbf{Q5.} Did those findings improve, and if not, was the treatment modified in response?\\
\textbf{A5.} (Medication/Procedure $\rightarrow$ switched\_to $\rightarrow$ Medication/Procedure; clinical reasoning from notes linking inadequate response to the treatment change)\\
\textbf{Q6.} What were the assessment results by discharge after that management course?\\
\textbf{A6.} (Lab\_Test.value/abnormal\_flag; Vital\_Sign.value/status; Physical\_Exam.finding; Diagnostic\_Imaging\_Test.result at discharge; Outcome.status)

\textbf{Template 2 --- Assessment Findings Used to Rule Out Competing Diagnoses}\\[0.15em]
\textbf{Q1.} Were any conditions suspected at admission $\langle$chartdate or admission\_id$\rangle$ based on the presenting findings?\\
\textbf{A1.} (Diagnosis.assertion; Diagnosis.status)\\
\textbf{Q2.} What clinical assessments were performed to evaluate those possibilities?\\
\textbf{A2.} (Lab\_Test; Diagnostic\_Imaging\_Test; Physical\_Exam; Vital\_Sign)\\
\textbf{Q3.} Did any of those assessment results make some of those possibilities less likely or rule them out?\\
\textbf{A3.} (Lab\_Test/Diagnostic\_Imaging\_Test/Physical\_Exam/Vital\_Sign $\rightarrow$ reveals $\rightarrow$ Diagnosis with negative assertion)\\
\textbf{Q4.} After evaluating those results, what condition was ultimately supported?\\
\textbf{A4.} (Lab\_Test/Diagnostic\_Imaging\_Test/Physical\_Exam/Vital\_Sign $\rightarrow$ reveals $\rightarrow$ Diagnosis)\\
\textbf{Q5.} What was the documented reasoning that connected those findings to the treatment decision?\\
\textbf{A5.} (Procedure/Medication $\rightarrow$ administered\_for $\rightarrow$ Diagnosis; clinical reasoning from notes linking the confirmatory findings to the treatment selection)\\
\textbf{Q6.} By discharge, were there any assessment findings demonstrating improvement of the condition?\\
\textbf{A6.} (Lab\_Test.value/abnormal\_flag; Vital\_Sign.value/status; Physical\_Exam.finding; Diagnostic\_Imaging\_Test.result; Outcome.status)

\textbf{Template 3 --- Assessment-Guided Treatment Adjustment and Monitoring During Admission}\\[0.15em]
\textbf{Q1.} Were there any assessment findings during admission $\langle$chartdate or admission\_id$\rangle$ that led to treatment initiation or change?\\
\textbf{A1.} (Lab\_Test/Diagnostic\_Imaging\_Test/Physical\_Exam/Vital\_Sign $\rightarrow$ resulted\_in $\rightarrow$ Medication/Procedure)\\
\textbf{Q2.} What were the values or findings in those assessments?\\
\textbf{A2.} (Lab\_Test.value; Lab\_Test.abnormal\_flag; Vital\_Sign.value/status; Diagnostic\_Imaging\_Test.result; Physical\_Exam.finding)\\
\textbf{Q3.} After that intervention, were the assessments repeated to monitor response, and if so, how frequently?\\
\textbf{A3.} (Lab\_Test.frequency; Vital\_Sign.frequency; Physical\_Exam.frequency)\\
\textbf{Q4.} Did the repeat assessments show improvement after that treatment was started?\\
\textbf{A4.} (Medication/Procedure $\rightarrow$ resulted\_in $\rightarrow$ Outcome/Finding; Lab\_Test.value normalization; Vital\_Sign.value improvement)\\
\textbf{Q5.} Were there any assessments that worsened despite that treatment?\\
\textbf{A5.} (Outcome.status; Outcome/Finding $\rightarrow$ after $\rightarrow$ Medication/Procedure; Lab\_Test.value trend; Vital\_Sign.value trend)\\
\textbf{Q6.} Did those findings improve adequately, and if not, was the treatment modified?\\
\textbf{A6.} (Medication/Procedure $\rightarrow$ switched\_to $\rightarrow$ Medication/Procedure; clinical reasoning from notes linking the specific worsening findings to the treatment change)\\
\textbf{Q7.} After treatment optimization, what were the assessment values at discharge?\\
\textbf{A7.} (Lab\_Test.value; Vital\_Sign.value; Physical\_Exam.finding; Diagnostic\_Imaging\_Test.result at discharge; Outcome.status)

\textbf{Template 4 --- Clinical Assessment-Driven Diagnosis Confirmation and Treatment Initiation}\\[0.15em]
\textbf{Q1.} What clinical assessments were performed to evaluate $\langle$Diagnosis/Symptom/Finding$\rangle$ at admission $\langle$chartdate or admission\_id$\rangle$?\\
\textbf{A1.} (Lab\_Test; Diagnostic\_Imaging\_Test; Physical\_Exam; Vital\_Sign; time = admission)\\
\textbf{Q2.} Were any abnormalities identified in those assessments?\\
\textbf{A2.} (Lab\_Test.abnormal\_flag; Lab\_Test.value; Vital\_Sign.status; Vital\_Sign.value; Physical\_Exam.finding; Diagnostic\_Imaging\_Test.result)\\
\textbf{Q3.} Did those findings point to any condition?\\
\textbf{A3.} (Lab\_Test/Diagnostic\_Imaging\_Test/Physical\_Exam/Vital\_Sign $\rightarrow$ reveals $\rightarrow$ Diagnosis)\\
\textbf{Q4.} Did those findings lead clinicians to start or change treatment?\\
\textbf{A4.} (Lab\_Test/Diagnostic\_Imaging\_Test/Physical\_Exam/Vital\_Sign $\rightarrow$ resulted\_in $\rightarrow$ Procedure/Medication)\\
\textbf{Q5.} By discharge, how had those assessment findings changed?\\
\textbf{A5.} (Lab\_Test.value at discharge; Vital\_Sign.value at discharge; Physical\_Exam.finding; Diagnostic\_Imaging\_Test.result; Outcome.status)

\textbf{Template 5 --- Cross-Admission Assessment Changes and Evaluation of New or Persistent Abnormalities}\\[0.15em]
\textbf{Q1.} Compared with the previous discharge, were there any new clinical findings or vital sign abnormalities at admission $\langle$chartdate or admission\_id$\rangle$?\\
\textbf{A1.} (Vital\_Sign.status/value; Physical\_Exam.finding; Diagnosis.status comparison over time; time = admission)\\
\textbf{Q2.} Were any diagnostic tests ordered to evaluate those abnormalities?\\
\textbf{A2.} (Symptom/Diagnosis $\rightarrow$ resulted\_in $\rightarrow$ Lab\_Test/Diagnostic\_Imaging\_Test; Lab\_Test/Diagnostic\_Imaging\_Test.time)\\
\textbf{Q3.} Were any results from those tests used to rule out suspected conditions?\\
\textbf{A3.} (Lab\_Test.value/assertion; Diagnostic\_Imaging\_Test.result/assertion; Lab\_Test/Diagnostic\_Imaging\_Test $\rightarrow$ reveals $\rightarrow$ Diagnosis with negative assertion)\\
\textbf{Q4.} Did the documented clinical reasoning interpret those cross-admission findings as progression, a new problem, or a recurrence?\\
\textbf{A4.} (clinical reasoning from notes linking cross-admission finding trends to the clinical interpretation; Lab\_Test/Vital\_Sign/Physical\_Exam $\rightarrow$ resulted\_in $\rightarrow$ Medication/Procedure)\\
\textbf{Q5.} Did any of those assessment trends lead clinicians to change treatment?\\
\textbf{A5.} (Medication $\rightarrow$ switched\_to $\rightarrow$ Medication; Lab\_Test.value trends; Vital\_Sign.status trends)\\
\textbf{Q6.} By discharge, were any abnormalities from those findings still present that required follow-up instructions?\\
\textbf{A6.} (Lab\_Test.abnormal\_flag; Vital\_Sign.status; Physical\_Exam.finding; Diagnosis/Symptom/Finding $\rightarrow$ has\_instruction $\rightarrow$ Instruction; time = discharge)

\subsection{Clinical Outcome}

\subsubsection{Single-note setting}
\textbf{Template 1 --- In-Hospital Complications, Rescue Interventions, and Final Outcome}\\[0.15em]
\textbf{Q1.} Did the patient experience any complications or clinical deterioration during admission?\\
\textbf{A1.} (Event.assertion/time; Diagnosis/Symptom/Finding.status change; Outcome.status change)\\
\textbf{Q2.} Were any clinical findings or test results documented that signaled the complication or deterioration?\\
\textbf{A2.} (Lab\_Test.abnormal\_flag/value; Vital\_Sign.status/value; Physical\_Exam.finding; Diagnostic\_Imaging\_Test.result $\rightarrow$ reveals $\rightarrow$ Diagnosis/Finding)\\
\textbf{Q3.} Was any underlying cause identified for the complication?\\
\textbf{A3.} (Procedure/Medication $\rightarrow$ causes $\rightarrow$ Diagnosis/Symptom/Finding; Diagnosis/Symptom/Finding $\rightarrow$ causes $\rightarrow$ Event)\\
\textbf{Q4.} Were any treatments or procedures initiated in response to the identified cause?\\
\textbf{A4.} (Procedure/Medication $\rightarrow$ administered\_for $\rightarrow$ Diagnosis/Symptom/Finding; Lab\_Test/Vital\_Sign $\rightarrow$ resulted\_in $\rightarrow$ Procedure/Medication)\\
\textbf{Q5.} Did the interventions resolve or improve the complication before discharge?\\
\textbf{A5.} (Procedure/Medication $\rightarrow$ improves $\rightarrow$ Diagnosis/Symptom/Finding; Procedure/Medication $\rightarrow$ resulted\_in $\rightarrow$ Outcome/Finding; Procedure/Medication $\rightarrow$ worsens $\rightarrow$ Diagnosis/Symptom/Finding)\\
\textbf{Q6.} What was the final outcome status at discharge, and were there any residual deficits?\\
\textbf{A6.} (Outcome.status; Diagnosis/Symptom/Finding.status; Activity.status at discharge)

\textbf{Template 2 --- Treatment Goal Attainment}\\[0.15em]
\textbf{Q1.} What was the primary clinical problem driving admission?\\
\textbf{A1.} (Diagnosis.is\_main; Symptom.is\_main; Finding.is\_main)\\
\textbf{Q2.} What interventions were used to treat that problem?\\
\textbf{A2.} (Procedure/Medication $\rightarrow$ administered\_for $\rightarrow$ Diagnosis/Symptom/Finding)\\
\textbf{Q3.} Was the primary problem resolved, improved, or still active at discharge?\\
\textbf{A3.} (Outcome.status; Diagnosis/Symptom/Finding.status at discharge; Outcome $\rightarrow$ after $\rightarrow$ Procedure/Medication)\\
\textbf{Q4.} Was there any evidence supporting whether the treatment goal was achieved?\\
\textbf{A4.} (Lab\_Test.value/abnormal\_flag; Vital\_Sign.value; Physical\_Exam.finding; Diagnostic\_Imaging\_Test.result)\\
\textbf{Q5.} If the goal was not fully achieved, were any barriers, side effects, or complications documented as contributing?\\
\textbf{A5.} (Procedure/Medication $\rightarrow$ causes $\rightarrow$ Diagnosis/Symptom/Finding; Procedure/Medication $\rightarrow$ worsens $\rightarrow$ Diagnosis/Symptom/Finding; Diagnosis/Symptom/Finding $\rightarrow$ causes $\rightarrow$ Event)\\
\textbf{Q6.} How was the follow-up plan designed to address those barriers and the remaining problem?\\
\textbf{A6.} (Any category $\rightarrow$ has\_instruction $\rightarrow$ Instruction; continued/planned Procedure/Medication; clinical reasoning from notes linking residual issues to the follow-up strategy)

\textbf{Template 3 --- Functional Outcome, Disposition, and Post-Discharge Support Needs}\\[0.15em]
\textbf{Q1.} What was the patient's functional status at the time of discharge?\\
\textbf{A1.} (Activity.status at discharge; Physical\_Exam.finding relevant to mobility/function)\\
\textbf{Q2.} Were there any clinical problems or outcomes that limited the patient's function at discharge?\\
\textbf{A2.} (Diagnosis/Symptom/Finding $\rightarrow$ resulted\_in $\rightarrow$ Outcome/Finding; Outcome $\rightarrow$ after $\rightarrow$ Diagnosis/Symptom/Finding)\\
\textbf{Q3.} Did the patient require any assistive devices, home services, or rehabilitation after discharge?\\
\textbf{A3.} (Medical\_Device.status; Activity.status; Outcome.assertion)\\
\textbf{Q4.} What discharge disposition was selected based on those functional limitations and support needs?\\
\textbf{A4.} (Outcome.status/time; Activity.status; Diagnosis/Symptom/Finding.status driving disposition decision)\\
\textbf{Q5.} Were any follow-up appointments, therapy, or instructions arranged to support functional recovery?\\
\textbf{A5.} (Any category $\rightarrow$ has\_instruction $\rightarrow$ Instruction; Procedure/Medication planned/continued post-discharge)

\textbf{Template 4 --- Medication Adjustments During Admission and Their Effect on Outcome}\\[0.15em]
\textbf{Q1.} Were any medications started, adjusted, or discontinued during admission, and why?\\
\textbf{A1.} (Medication.assertion/time; Procedure/Medication $\rightarrow$ administered\_for $\rightarrow$ Diagnosis/Symptom/Finding; Procedure/Medication $\rightarrow$ switched\_to $\rightarrow$ Procedure/Medication)\\
\textbf{Q2.} Did any medication cause adverse effects or complications during the admission?\\
\textbf{A2.} (Procedure/Medication $\rightarrow$ causes $\rightarrow$ Diagnosis/Symptom/Finding; Procedure/Medication $\rightarrow$ worsens $\rightarrow$ Diagnosis/Symptom/Finding)\\
\textbf{Q3.} How were those adverse effects managed, and were medications adjusted as a result?\\
\textbf{A3.} (Procedure/Medication $\rightarrow$ switched\_to $\rightarrow$ Procedure/Medication; Procedure/Medication $\rightarrow$ causes $\rightarrow$ Procedure/Medication; Procedure/Medication $\rightarrow$ administered\_for $\rightarrow$ Diagnosis/Symptom/Finding)\\
\textbf{Q4.} What was the overall effect of the final medication regimen on the patient's primary condition by discharge?\\
\textbf{A4.} (Procedure/Medication $\rightarrow$ improves $\rightarrow$ Diagnosis/Symptom/Finding; Procedure/Medication $\rightarrow$ resulted\_in $\rightarrow$ Outcome/Finding; Outcome $\rightarrow$ after $\rightarrow$ Procedure/Medication)\\
\textbf{Q5.} What medications were prescribed at discharge, and were any instructions given regarding their use?\\
\textbf{A5.} (Medication.strength/dosage/route/instruction at discharge; Any category $\rightarrow$ has\_instruction $\rightarrow$ Instruction)

\textbf{Template 5 --- Goal Attainment and Residual Risk at Discharge}\\[0.15em]
\textbf{Q1.} What was the intended treatment goal during admission?\\
\textbf{A1.} (Diagnosis/Symptom/Finding.is\_main)\\
\textbf{Q2.} Was the treatment goal achieved by discharge?\\
\textbf{A2.} (Outcome.status at discharge, Diagnosis/Symptom/Finding.status at discharge, Outcome $\rightarrow$ after $\rightarrow$ Diagnosis/Symptom/Finding)\\
\textbf{Q3.} Was there any objective data confirming goal attainment?\\
\textbf{A3.} (Lab\_Test.value/abnormal\_flag, Vital\_Sign.status/value, Physical\_Exam.finding, Diagnostic\_Imaging\_Test.result)\\
\textbf{Q4.} If the goal was not fully achieved, were any factors documented as contributing to the incomplete response?\\
\textbf{A4.} (Procedure/Medication $\rightarrow$ causes $\rightarrow$ Diagnosis/Symptom/Finding, Diagnosis/Symptom/Finding $\rightarrow$ causes $\rightarrow$ Event, Procedure/Medication $\rightarrow$ worsens $\rightarrow$ Diagnosis/Symptom/Finding)\\
\textbf{Q5.} What was the documented clinical reasoning for determining which residual conditions posed the highest risk?\\
\textbf{A5.} (clinical reasoning from notes linking residual conditions to specific risk factors; Persistent Diagnosis/Symptom/Finding.status; Outcome.status = partial/stable with deficits)\\
\textbf{Q6.} Was any follow-up plan established to prevent worsening?\\
\textbf{A6.} (Any category $\rightarrow$ has\_instruction $\rightarrow$ Instruction, Planned/continued Medication/Procedure)

\subsubsection{Multi-note setting}
\textbf{Template 1 --- Discharge Readiness: Clinical Stability and Objective Evidence}\\[0.15em]
\textbf{Q1.} What was the patient's overall clinical status at discharge during admission $\langle$chartdate or admission\_id$\rangle$?\\
\textbf{A1.} (Outcome.status; Outcome.time = discharge)\\
\textbf{Q2.} Were there any conditions, findings, or symptoms that were improved, resolved, or still active at discharge?\\
\textbf{A2.} (Diagnosis.status; Symptom.status; Finding.status; time = discharge)\\
\textbf{Q3.} Were there any lab tests, vital signs, physical exams, or diagnostic imaging tests that supported that stability or improvement at discharge?\\
\textbf{A3.} (Lab\_Test.value/abnormal\_flag; Vital\_Sign.value/status; Physical\_Exam.finding; Diagnostic\_Imaging\_Test.result at discharge)\\
\textbf{Q4.} Among those clinical findings, were any abnormalities still present that could pose short-term risk after discharge?\\
\textbf{A4.} (Finding.status; Lab\_Test.abnormal\_flag; Vital\_Sign.status; persistent Diagnosis/Symptom)\\
\textbf{Q5.} What was the documented reasoning for determining the patient was safe for discharge despite those residual abnormalities?\\
\textbf{A5.} (clinical reasoning from notes linking the residual findings to the discharge readiness assessment; risk mitigation strategies documented)\\
\textbf{Q6.} Were any monitoring or contingency instructions provided to address those remaining risks?\\
\textbf{A6.} (Any category $\rightarrow$ has\_instruction $\rightarrow$ Instruction)

\textbf{Template 2 --- In-Hospital Deterioration, Rescue Interventions, and Outcome Shift}\\[0.15em]
\textbf{Q1.} Did the patient clinically deteriorate during admission $\langle$chartdate or admission\_id$\rangle$?\\
\textbf{A1.} (Event.assertion/time; Outcome.status change; Diagnosis/Symptom/Finding.status change)\\
\textbf{Q2.} Were there any clinical findings or assessment results that indicated that deterioration?\\
\textbf{A2.} (Lab\_Test.abnormal\_flag/value; Vital\_Sign.status/value; Physical\_Exam.finding; Diagnostic\_Imaging\_Test.result)\\
\textbf{Q3.} Were any clinical problems identified as the cause of that deterioration?\\
\textbf{A3.} (Diagnosis/Symptom/Finding $\rightarrow$ causes $\rightarrow$ Event; Event $\rightarrow$ causes $\rightarrow$ Diagnosis/Symptom/Finding)\\
\textbf{Q4.} Were any urgent treatments or procedures performed in response?\\
\textbf{A4.} (Procedure/Medication $\rightarrow$ administered\_for $\rightarrow$ Diagnosis/Symptom/Finding)\\
\textbf{Q5.} Did those interventions stabilize or reverse the deterioration before discharge?\\
\textbf{A5.} (Procedure/Medication $\rightarrow$ resulted\_in $\rightarrow$ Outcome/Finding; Procedure/Medication $\rightarrow$ improves $\rightarrow$ Diagnosis/Symptom/Finding; Procedure/Medication $\rightarrow$ worsens $\rightarrow$ Diagnosis/Symptom/Finding)\\
\textbf{Q6.} What was the patient's final outcome status at discharge, and were any deficits still present?\\
\textbf{A6.} (Outcome.status; Activity.status; Diagnosis/Symptom/Finding.status)

\textbf{Template 3 --- Functional Outcome and Care Needs: Disposition and Supports}\\[0.15em]
\textbf{Q1.} What was the patient's functional status at discharge for admission $\langle$chartdate or admission\_id$\rangle$?\\
\textbf{A1.} (Activity.status; time = discharge; Physical\_Exam.finding relevant to function)\\
\textbf{Q2.} Were there any clinical problems that contributed to that functional status at discharge?\\
\textbf{A2.} (Diagnosis/Symptom/Finding $\rightarrow$ resulted\_in $\rightarrow$ Outcome/Finding; Outcome $\rightarrow$ after $\rightarrow$ Diagnosis/Symptom/Finding)\\
\textbf{Q3.} Did the patient require any devices, services, or caregiver support after discharge?\\
\textbf{A3.} (Medical\_Device.status; Activity.status; Outcome.assertion)\\
\textbf{Q4.} What discharge disposition was selected based on that support plan?\\
\textbf{A4.} (Outcome.status/time; Activity.status)\\
\textbf{Q5.} Was any follow-up or rehabilitation plan documented to improve that functional status?\\
\textbf{A5.} (Any category $\rightarrow$ has\_instruction $\rightarrow$ Instruction; planned/continued Procedure or Medication)

\textbf{Template 4 --- Longitudinal Recovery Trajectory Across Admissions}\\[0.15em]
\textbf{Q1.} Across admissions, how did the patient's outcome status change from each discharge to the next admission?\\
\textbf{A1.} (Outcome.status over time; time.category = discharge/admission)\\
\textbf{Q2.} During that longitudinal course, were there clinical problems that repeatedly worsened after discharge or failed to resolve?\\
\textbf{A2.} (Diagnosis/Symptom/Finding.status over time)\\
\textbf{Q3.} Were any factors associated with worsening of those conditions between encounters?\\
\textbf{A3.} (Medication/Procedure $\rightarrow$ worsens $\rightarrow$ Diagnosis/Symptom/Finding; Event $\rightarrow$ causes $\rightarrow$ Diagnosis/Symptom/Finding; Diagnosis/Symptom/Finding $\rightarrow$ causes $\rightarrow$ Event)\\
\textbf{Q4.} How did the treatment approach evolve to address those factors and the pattern of deterioration?\\
\textbf{A4.} (Medication/Procedure $\rightarrow$ improves $\rightarrow$ Diagnosis/Symptom/Finding; Medication/Procedure $\rightarrow$ resulted\_in $\rightarrow$ Outcome/Finding; clinical reasoning from notes linking cross-admission worsening factors to treatment modifications)\\
\textbf{Q5.} By the most recent discharge, were any functional limitations related to those conditions still present, and what plan addressed them?\\
\textbf{A5.} (Activity.status; Diagnosis/Symptom/Finding.status at discharge; Instruction)

\textbf{Template 5 --- Treatment Goal Attainment vs Failure: Why the Outcome Was Achieved (or Not)}\\[0.15em]
\textbf{Q1.} What was the primary treatment goal during admission $\langle$chartdate or admission\_id$\rangle$?\\
\textbf{A1.} (Diagnosis/Symptom/Finding.is\_main; Diagnosis/Symptom/Finding.assertion)\\
\textbf{Q2.} What treatments were administered to manage the condition?\\
\textbf{A2.} (Procedure/Medication $\rightarrow$ administered\_for $\rightarrow$ Diagnosis/Symptom/Finding)\\
\textbf{Q3.} By discharge, did the condition meet the intended treatment goal?\\
\textbf{A3.} (Outcome.status at discharge; Diagnosis/Symptom/Finding.status at discharge; Outcome $\rightarrow$ after $\rightarrow$ Procedure/Medication)\\
\textbf{Q4.} Were there any lab test, vital sign, physical exam, or diagnostic imaging results that supported the assessment of that treatment outcome?\\
\textbf{A4.} (Lab\_Test.value/abnormal\_flag; Vital\_Sign.value/status; Physical\_Exam.finding; Diagnostic\_Imaging\_Test.result)\\
\textbf{Q5.} Did the condition fully improve, and if not, were any complications or barriers documented as contributing to the incomplete response?\\
\textbf{A5.} (Procedure/Medication $\rightarrow$ causes $\rightarrow$ Diagnosis/Symptom/Finding/Event; Diagnosis/Symptom/Finding $\rightarrow$ causes $\rightarrow$ Event)\\
\textbf{Q6.} How was the follow-up plan designed to address those barriers and the remaining problem?\\
\textbf{A6.} (Any category $\rightarrow$ has\_instruction $\rightarrow$ Instruction; continued/planned Procedure or Medication; clinical reasoning from notes linking residual issues to the follow-up strategy)

\subsection{Discharge Plan}

\subsubsection{Single-note setting}
\textbf{Template 1 --- Discharge Medication Reconciliation}\\[0.15em]
\textbf{Q1.} What medications was the patient prescribed at discharge?\\
\textbf{A1.} (Medication.assertion, time = discharge; Medication.strength, dosage, route, frequency)\\
\textbf{Q2.} Which of those medications were newly started during the hospitalization versus continued from admission?\\
\textbf{A2.} (Medication.status comparison: admission vs. discharge; Medication.time)\\
\textbf{Q3.} Were any admission medications stopped or held at discharge, and if so, why?\\
\textbf{A3.} (Medication.status = stopped/held; Medication $\rightarrow$ administered\_for $\rightarrow$ Diagnosis/Symptom/Finding or adverse effect)\\
\textbf{Q4.} For the new or changed medications, what condition was each prescribed for?\\
\textbf{A4.} (Medication $\rightarrow$ administered\_for $\rightarrow$ Diagnosis/Symptom/Finding)\\
\textbf{Q5.} Were any dose adjustments made during the hospitalization before arriving at the final discharge dose, and what drove those changes?\\
\textbf{A5.} (Medication.dosage change; Medication $\rightarrow$ causes $\rightarrow$ Symptom/Finding; Medication $\rightarrow$ worsens $\rightarrow$ Symptom/Finding)

\textbf{Template 2 --- Discharge Disposition and Care Setting Rationale}\\[0.15em]
\textbf{Q1.} Where was the patient discharged to, and what was the stated reason for that disposition?\\
\textbf{A1.} (Outcome.status/assertion = discharge disposition; Activity.status; Diagnosis/Finding $\rightarrow$ resulted\_in $\rightarrow$ Outcome)\\
\textbf{Q2.} What was the patient's clinical condition at the time of discharge that supported the disposition decision?\\
\textbf{A2.} (Vital\_Sign.value, time = discharge; Physical\_Exam.finding, time = discharge; Outcome.assertion)\\
\textbf{Q3.} What was the documented reasoning for determining the appropriate level of post-discharge care based on those findings?\\
\textbf{A3.} (clinical reasoning from notes linking the discharge clinical status to the disposition decision; Activity.status; functional limitations)\\
\textbf{Q4.} Were any follow-up services or care arrangements put in place at the receiving facility or at home?\\
\textbf{A4.} (Instruction.instruction\_text; Event.status = scheduled; Medical\_Device/Activity supports)\\
\textbf{Q5.} Were there any outstanding clinical issues at discharge that the receiving care setting was expected to manage?\\
\textbf{A5.} (Diagnosis/Finding.status = ongoing; Instruction.instruction\_text directed to next provider)

\textbf{Template 3 --- Follow-Up Plan, Outpatient Connections, and Contingency Instructions}\\[0.15em]
\textbf{Q1.} Were any follow-up appointments or outpatient referrals arranged at discharge?\\
\textbf{A1.} (Event.status = scheduled; Instruction.instruction\_text; Instruction.time)\\
\textbf{Q2.} What clinical problems was each follow-up appointment intended to address?\\
\textbf{A2.} (Diagnosis/Symptom/Finding/Outcome $\rightarrow$ has\_instruction $\rightarrow$ Instruction)\\
\textbf{Q3.} Were any instructions given regarding who to contact and under what circumstances before the scheduled follow-up?\\
\textbf{A3.} (Instruction.instruction\_text; Symptom/Finding $\rightarrow$ has\_instruction $\rightarrow$ Instruction with urgency framing)\\
\textbf{Q4.} Were there any explicit instructions about substances, behaviors, or lifestyle factors the patient was told to avoid?\\
\textbf{A4.} (Instruction.instruction\_text; Activity $\rightarrow$ has\_instruction $\rightarrow$ Instruction; Medication/substance warnings)\\
\textbf{Q5.} What should the patient do if they are unable to reach their outpatient provider or feel unsafe?\\
\textbf{A5.} (Instruction.instruction\_text = emergency contingency; Event $\rightarrow$ has\_instruction $\rightarrow$ Instruction)

\textbf{Template 4 --- Symptom Monitoring, Red Flags, and Emergency Planning}\\[0.15em]
\textbf{Q1.} Were any symptoms or warning signs that the patient should monitor after discharge documented?\\
\textbf{A1.} (Symptom/Finding $\rightarrow$ has\_instruction $\rightarrow$ Instruction)\\
\textbf{Q2.} Which conditions or complications were those warning signs intended to detect or prevent?\\
\textbf{A2.} (Diagnosis/Symptom/Finding $\rightarrow$ has\_instruction $\rightarrow$ Instruction)\\
\textbf{Q3.} What was the documented reasoning for emphasizing those specific warning signs based on the patient's hospital course?\\
\textbf{A3.} (clinical reasoning from notes linking the hospital course --- complications, persistent findings, or high-risk conditions --- to the specific warning signs selected)\\
\textbf{Q4.} What actions was the patient instructed to take if those symptoms occur?\\
\textbf{A4.} (Instruction.instruction\_text; may include Event)\\
\textbf{Q5.} Were any objective thresholds provided?\\
\textbf{A5.} (Instruction.instruction\_text; Vital\_Sign.value thresholds if present)

\textbf{Template 5 --- Follow-Up Care, Appointments, and Diagnostic Monitoring}\\[0.15em]
\textbf{Q1.} Were any follow-up appointments or referrals arranged at discharge?\\
\textbf{A1.} (Instruction.instruction\_text; Event.status = scheduled)\\
\textbf{Q2.} What medical or psychiatric issues were those follow-ups meant to address?\\
\textbf{A2.} (Diagnosis/Symptom/Finding/Outcome $\rightarrow$ has\_instruction $\rightarrow$ Instruction)\\
\textbf{Q3.} Were any follow-up tests such as labs, imaging, or microbiology recommended?\\
\textbf{A3.} (Lab\_Test/Diagnostic\_Imaging\_Test/Microbiology\_Test planned; Instruction)\\
\textbf{Q4.} What was the recommended timing or urgency of follow-up?\\
\textbf{A4.} (Instruction.time; Instruction.instruction\_text)\\
\textbf{Q5.} Were any contingency plans described if follow-up could not be completed?\\
\textbf{A5.} (Instruction.instruction\_text; case management/social work if applicable)

\subsubsection{Multi-note setting}
\textbf{Template 1 --- Medication Reconciliation and Post-Discharge Medication Plan}\\[0.15em]
\textbf{Q1.} What medications was the patient instructed to take after discharge at admission $\langle$chartdate or admission\_id$\rangle$?\\
\textbf{A1.} (Medication.assertion; Medication.time = discharge)\\
\textbf{Q2.} Compared with the medications in admission $\langle$chartdate or admission\_id$\rangle$, were any of those medications newly started, stopped, held, or dose-adjusted?\\
\textbf{A2.} (Medication.status/time comparison; Medication.dosage/route/frequency changes)\\
\textbf{Q3.} For those medications, what clinical conditions or symptoms were they intended to treat?\\
\textbf{A3.} (Medication $\rightarrow$ administered\_for $\rightarrow$ Diagnosis/Symptom/Finding)\\
\textbf{Q4.} What was the documented reasoning that explained the evolution of the medication plan from one discharge to the next?\\
\textbf{A4.} (clinical reasoning from notes linking cross-admission clinical changes to the medication adjustments; Medication $\rightarrow$ administered\_for $\rightarrow$ Diagnosis/Symptom/Finding)\\
\textbf{Q5.} Were any monitoring instructions provided for those treatments after discharge?\\
\textbf{A5.} (Medication $\rightarrow$ has\_instruction $\rightarrow$ Instruction)\\
\textbf{Q6.} In the instructions associated with those treatments, were there any warnings about adherence or risks if the medications were stopped or taken incorrectly?\\
\textbf{A6.} (Instruction.instruction\_text; Event risk framing if present)

\textbf{Template 2 --- Red Flags, Return Precautions, and Symptom Monitoring}\\[0.15em]
\textbf{Q1.} At discharge from admission $\langle$chartdate or admission\_id$\rangle$, were any warning symptoms or clinical changes that the patient should watch for at home documented?\\
\textbf{A1.} (Symptom/Finding $\rightarrow$ has\_instruction $\rightarrow$ Instruction)\\
\textbf{Q2.} Which underlying condition or complication were those warning signs intended to detect early or prevent?\\
\textbf{A2.} (Diagnosis/Symptom/Finding $\rightarrow$ has\_instruction $\rightarrow$ Instruction)\\
\textbf{Q3.} What was the documented reasoning for emphasizing those specific warning signs based on the patient's hospital course?\\
\textbf{A3.} (clinical reasoning from notes linking the hospital course --- complications, persistent findings, or high-risk conditions --- to the specific warning signs selected)\\
\textbf{Q4.} If those warning signs occurred, what actions was the patient instructed to take?\\
\textbf{A4.} (Instruction.instruction\_text)\\
\textbf{Q5.} Within the guidance related to those warning signs, were any objective thresholds provided that would trigger action?\\
\textbf{A5.} (Instruction.instruction\_text with numeric thresholds; Vital\_Sign.value targets if present)\\
\textbf{Q6.} Beyond watching for those warning signs, was any follow-up monitoring plan arranged to track recovery?\\
\textbf{A6.} (Instruction.instruction\_text; Event.status = scheduled if represented)

\textbf{Template 3 --- Follow-up Appointments and Care Coordination}\\[0.15em]
\textbf{Q1.} At discharge from admission $\langle$chartdate or admission\_id$\rangle$, were any follow-up appointments, referrals, or services arranged for the patient?\\
\textbf{A1.} (Instruction.instruction\_text; Event.status = scheduled)\\
\textbf{Q2.} What clinical problems or conditions were those follow-ups intended to address?\\
\textbf{A2.} (Diagnosis/Symptom/Finding/Outcome $\rightarrow$ has\_instruction $\rightarrow$ Instruction)\\
\textbf{Q3.} As part of that follow-up plan, were any diagnostic tests scheduled?\\
\textbf{A3.} (Lab\_Test / Diagnostic\_Imaging\_Test / Microbiology\_Test planned; Instruction)\\
\textbf{Q4.} Were any instructions given about the timing or urgency of those follow-ups, and who the patient should contact?\\
\textbf{A4.} (Instruction.time; Instruction.instruction\_text)\\
\textbf{Q5.} If the patient was unable to complete those follow-ups, were any contingency instructions documented?\\
\textbf{A5.} (Instruction.instruction\_text; social work/case management references if present)

\textbf{Template 4 --- Disposition, Functional Plan, and Post-Discharge Supports}\\[0.15em]
\textbf{Q1.} What was the patient's disposition at discharge at admission $\langle$chartdate or admission\_id$\rangle$, and why?\\
\textbf{A1.} (Outcome.assertion/status; Activity.status)\\
\textbf{Q2.} What was the patient's functional or activity level at the time of discharge?\\
\textbf{A2.} (Activity.status, time = discharge)\\
\textbf{Q3.} Did the patient require any medical devices or supportive services after discharge?\\
\textbf{A3.} (Medical\_Device.status; Activity.status; Instruction)\\
\textbf{Q4.} Were any activity restrictions or rehabilitation instructions provided in relation to that support plan?\\
\textbf{A4.} (Instruction.instruction\_text; Activity-related instructions)\\
\textbf{Q5.} Were any criteria given for advancing activity or seeking reassessment?\\
\textbf{A5.} (Instruction.instruction\_text; thresholds or time-based progression)

\textbf{Template 5 --- Cross-Admission Discharge Plan Adherence and Readmission Link}\\[0.15em]
\textbf{Q1.} What were the discharge instructions given after admission $\langle$chartdate or admission\_id$\rangle$?\\
\textbf{A1.} (Any category $\rightarrow$ has\_instruction $\rightarrow$ Instruction; time = prior discharge)\\
\textbf{Q2.} At the next hospitalization, was there evidence that those instructions were followed or not followed?\\
\textbf{A2.} (Event.status = missed follow-up / non-adherence; Medication.status = not taking)\\
\textbf{Q3.} Among those instructions, were any aspects of non-adherence clinically significant?\\
\textbf{A3.} (Medication.status; Event.status; Symptom.status)\\
\textbf{Q4.} What was the documented clinical reasoning that connected those non-adherence failures to the current presentation at readmission?\\
\textbf{A4.} (clinical reasoning from notes linking specific non-adherence to clinical deterioration; Medication $\rightarrow$ worsens $\rightarrow$ Diagnosis/Symptom/Finding; Event $\rightarrow$ causes $\rightarrow$ Diagnosis/Symptom/Finding)\\
\textbf{Q5.} At the most recent discharge, was the discharge plan modified to reduce the risk of recurrence or improve adherence?\\
\textbf{A5.} (Updated Instruction; Medication adjustments; follow-up changes across admissions)

}

\clearpage
\section{Per-Step Details for LLM-Based Initial Data Generation}\label{app:generation-details}

This appendix expands on the four-step generation pipeline summarized in Section~\ref{sec:generation}. 
We use Gemini-2.5-Pro at temperature~1.0 for Steps~1--2 (to encourage diversity) and 0.7 for Steps~3--4 (to maintain answer-choice quality). 
The full pipeline cost approximately \$1{,}000 for 1{,}000 samples ($\sim$\$1 per sample). 
The complete prompt for each step is given in Appendix~\ref{app:prompts}, and example per-step outputs in Appendix~\ref{app:step-outputs}.

\paragraph{Step 1: Multi-turn question generation.}
The input consists of a single patient's full sequence of discharge summaries, the structuring schema (Appendix~\ref{app:schema}), the target question category together with the five expert-curated QA templates for that category (single-note templates when the patient has one note and multi-note templates otherwise), and one randomly sampled template from each of the other seven categories. Gemini-2.5-Pro generates the multi-turn questions together with the supporting evidence and source location (note number, section header, and text span) for answering each question. Generating these jointly with the questions ensures the questions are grounded in the notes, and they are reused in later steps to produce the correct content-answer and evidence-grounding choices. The prompt enforces, in priority order, the clinical-naturalness, multi-turn-dependency, and multi-note-coverage constraints described in Section~\ref{sec:generation}.

\paragraph{Step 2: Incorrect-answer strategy brainstorming.}
The input is the patient's discharge summaries, the structuring schema, and the complete Step~1 output (multi-turn questions with evidence and source location). For each question, the model produces one distractor direction for each of six types, derived from two grounding sources: incorrect interpretations of the notes (\emph{note-grounded}) and content not present in the notes but clinically plausible given the patient's context (\emph{clinical-knowledge-grounded}). To support cross-turn error chaining, the prompt encourages later-turn distractor directions to build on earlier ones (e.g., a later distractor involving a medication prescribed for an earlier distractor's incorrect diagnosis). The six distractor types are:
\begin{enumerate}
\setlength{\itemsep}{2pt}\setlength{\parskip}{0pt}\setlength{\topsep}{2pt}
\item \textbf{Note-grounded entity/attribute substitution} (note-grounded): replace the correct entity or attribute with a related but incorrect one drawn from elsewhere in the notes.
\item \textbf{Plausible alternative interpretation} (clinical-knowledge-grounded): introduce a competing interpretation of the patient's documented presentation, such as an alternative diagnosis, etiology, or drug-class indication, that is not documented but is medically plausible given the patient's context.
\item \textbf{Plausible alternative action} (clinical-knowledge-grounded): introduce a competing action, such as a different intervention, medication within the same class, test, or follow-up plan, for the same clinical indication that is medically plausible but not documented.
\item \textbf{Clinical relationship misattribution} (note- or clinical-knowledge-grounded): misattribute relationships among clinical entities, such as by reversing causality, altering temporal order, or confusing entities across admissions.
\item \textbf{Note-grounded answer with one clinically incorrect detail} (note- and clinical-knowledge-grounded): use content that is mostly consistent with the notes but includes one clinically plausible yet incorrect detail, such as an incorrect dosage or severity grade.
\item \textbf{Cross-admission temporal displacement} (note-grounded): attribute a correct clinical fact to the wrong admission. This type applies only to patients with multiple admissions.
\end{enumerate}

\paragraph{Step 3: Content-question answer choice generation.}
The model is given the discharge summaries, the structuring schema, the Step~1 questions (with evidence and source location), and the Step~2 distractor directions, and generates one correct and four incorrect content choices per question. The correct choice is grounded in the Step~1 evidence; the incorrect choices follow the Step~2 directions, subject to two additional constraints. \textbf{Shared-entity choices} (Section~\ref{sec:generation}): some incorrect choices reference the same key entities as the correct answer but differ in the relationship, attribute, or interpretation applied to them, preventing an entity-presence lookup strategy. \textbf{Error-consistent choices:} letting $Q_n^{\mathrm{cont}}$ denote the content question at turn~$n$, for $n \geq 2$ the incorrect choices for $Q_n^{\mathrm{cont}}$ are encouraged to reflect clinically plausible downstream answers that would follow if the model had committed to a specific wrong answer for $Q_{n-1}^{\mathrm{cont}}$, realizing the cross-turn error chains brainstormed in Step~2.

\paragraph{Step 4: Evidence-grounding answer choice generation.}
The model is given the discharge summaries, the Step~1 questions (with evidence and source location), and the Step~3 correct answer text, and generates one correct and four incorrect evidence-grounding choices, following the minimal-yet-sufficient correct-choice constraint and the four incorrect-choice error patterns defined in Section~\ref{sec:generation}. As a concrete example of the semantic-trap principle, if the answer concerns an ``intervention'', a distractor might include the \textit{Major Surgical or Invasive Procedure} header even though that section's content does not actually describe the intervention. This forces the question to be solved by reading the section content rather than by surface header-name matching.

\clearpage
\section{Per-Step Output Examples for LLM-Based Initial Data Generation}\label{app:step-outputs}

This appendix shows an example LLM output for each of the four generation steps, all drawn from the same Microbiology-category multi-turn QA sample for a single patient with two discharge summaries.

\subsection{Step 1: Multi-turn Question Generation}\label{app:step1-output}

An example Step~1 output for a single patient with two discharge summaries is shown below (Microbiology-category multi-turn QA sample). To avoid disclosing patient-identifying dates and note content, chartdates are replaced with placeholders ([chartdate1] for Note~\#1, [chartdate2] for Note~\#2), and the exact text span in each \texttt{[Source Location]} block is replaced with \texttt{\{raw\_text\_snippet\}}. In the EHRNote-ChatQA dataset and in the LLM-generated Step~1 output, these fields contain the actual chartdates and supporting text spans; they are shown here only as placeholders to avoid disclosing patient-identifying information.

For each multi-turn content question $Q_n$ it generates, Gemini-2.5-Pro is prompted to jointly produce three components in a structured block: (1)~\texttt{[Entities, Attributes, Relationships]}, the ``answer evidence'' for the question---the schema-defined entities, attributes, and relationships extracted from the discharge summaries that are necessary to answer it; (2)~\texttt{[Source Location]}, the note number(s), section header(s), and supporting raw text span(s) in the discharge summaries required to answer it; and (3)~\texttt{[Answer]}, the natural-language answer. The \texttt{[Answer]} is not used in Steps~2--4; it is produced only so that later turns within Step~1 can build on earlier answers when generating subsequent questions. The answer evidence (\texttt{[Entities, Attributes, Relationships]}) and \texttt{[Source Location]} are used in Steps~2--4 to construct the content-question answer choices and the evidence-grounding answer choices. 

{\footnotesize
\begin{verbatim}
Q1: During the admission for the right radical nephrectomy on chart date
    [chartdate1], were any infections noted or were any preventative measures
    taken?
A1:
[Entities, Attributes, Relationships]
{Diagnosis} - name: infection, assertion: negative
{Procedure} - name: laparoscopic right radical nephrectomy, is_main: true,
              time: during admission on [chartdate1]
{Medication} - name: antibiotic prophylaxis, status: received,
               time: perioperative
{Medication} -> administered_for -> {Procedure}
[Source Location]
{Note #1}{Brief Hospital Course}
{raw_text_snippet}
[Answer]
No infection was diagnosed during this admission. The patient received
perioperative antibiotic prophylaxis for the laparoscopic right radical
nephrectomy.

Q2: Was the patient monitored for any signs of infection after that
    intervention?
A2:
[Entities, Attributes, Relationships]
{Physical_Exam} - name: examination of incision, time: on discharge
{Finding} - name: clean, dry, and intact, finding_site: incision
{Finding} - name: no evidence of hematoma collection or infection,
            finding_site: incision, assertion: negative
{Physical_Exam} -> reveals -> {Finding}
[Source Location]
{Note #1}{Brief Hospital Course}
{raw_text_snippet}
[Answer]
Yes, on the day of discharge, the surgical incision was examined and found
to be clean, dry, and intact, with no evidence of hematoma collection or
infection.

Q3: Were there any laboratory findings during that admission that could
    suggest an inflammatory process?
A3:
[Entities, Attributes, Relationships]
{Lab_Test} - name: WBC, value: 11.3*, abnormal_flag: abnormal,
             time: ___ 06:58AM
[Source Location]
{Note #1}{Pertinent Results}
{raw_text_snippet}
[Answer]
Yes, a complete blood count during the admission showed a white blood cell
count of 11.3, which is slightly elevated.

Q4: What was the clinical assessment of that lab result?
A4:
[Entities, Attributes, Relationships]
{Outcome} - name: hospital course, status: unremarkable
{Outcome} - name: condition at discharge, status: stable
{Lab_Test} - name: WBC, value: 11.3*
[Source Location]
{Note #1}{Brief Hospital Course}
{raw_text_snippet}
[Answer]
The hospital course was documented as unremarkable, and the patient was
discharged in stable condition with a clean incision. This suggests the
slightly elevated white blood cell count was considered a normal
post-operative inflammatory response rather than a sign of a clinically
significant infection.

Q5: Was any treatment for an infection continued at discharge?
A5:
[Entities, Attributes, Relationships]
{Medication} - name: antibiotics, assertion: negative, time: at discharge
[Source Location]
{Note #1}{Discharge Medications}
{raw_text_snippet}
[Answer]
No, no antibiotics were included in the discharge medication list. The
patient was prescribed pain medications and a stool softener.

Q6: During the subsequent admission on chart date [chartdate2], was there
    any evidence of an infectious process?
A6:
[Entities, Attributes, Relationships]
{Diagnosis} - name: infection, assertion: negative,
              time: admission on [chartdate2]
{Diagnosis} - name: pulmonary embolism, is_main: true,
              time: admission on [chartdate2]
{Lab_Test} - name: WBC, value: 9.7, abnormal_flag: normal,
             time: ___ 08:37AM
[Source Location]
{Note #2}{Brief Hospital Course}
{raw_text_snippet}
{Note #2}{Pertinent Results}
{raw_text_snippet}
[Answer]
No, there was no evidence of an infectious process during the admission on
chart date [chartdate2]. The primary diagnosis was a pulmonary embolism, and
the patient's white blood cell count on admission was within the normal
range at 9.7.
\end{verbatim}
}

\clearpage
\subsection{Step 2: Distractor-Direction Brainstorming}\label{app:step2-output}

The Step~2 output for two consecutive questions (Q1 and Q2) is shown below, from the Microbiology-category multi-turn QA sample for the same patient used in Appendix~\ref{app:step1-output}. In Step~2, Gemini-2.5-Pro generates distractor directions for the full sequence of Step~1 questions in a single pass; for brevity, we show the output for two consecutive questions only (Q1 and Q2). For each content question $Q_n$, the model produces one distractor direction for each of the six distractor types defined in Appendix~\ref{app:generation-details}. Each direction is annotated with \texttt{Explanation}, \texttt{Reasoning}, and a \texttt{Minimal-pair label} indicating whether the distractor differs from the correct answer in exactly one detail. In the output, \texttt{Source A} denotes note-grounded content and \texttt{Source B} denotes clinical-knowledge-grounded content, defined in Section~\ref{sec:generation}. As in Appendix~\ref{app:step1-output}, the chartdate in the question text is shown as a placeholder to avoid disclosing patient-identifying dates.

The first content question (Q1) contains the six distractor directions. For each follow-up content question $Q_n$ ($n \geq 2$), the model performs two additional analyses before producing the six distractor directions. In the \texttt{Referential Ambiguity Analysis}, it identifies the referential expression in $Q_n$, its correct referent, and plausible alternative referents, then constructs a distractor direction for each alternative. In the \texttt{Cross-Turn Error Path Analysis}, it identifies clinically plausible distractor directions to the previous content question $Q_{n-1}$ (drawn from $Q_{n-1}$'s Step~2 distractor directions) and designs distractor directions for $Q_n$ that would be natural consequences. These outputs are used in Step~3 to construct the four incorrect content-answer choices for each question.

{\footnotesize
\begin{verbatim}
Q1: During the admission for the right radical nephrectomy on chart
    date [chartdate1], were any infections noted or were any
    preventative measures taken?
- Effective Distractor Type 1: Note-grounded entity substitution with
    clinical reasoning requirement (Source A)
Explanation: State that a urinary tract infection was diagnosed
    post-operatively.
Reasoning: The note mentions the patient had a urethral Foley catheter
    placed, which is a major risk factor for a UTI. A model might
    plausibly infer a UTI occurred, even though the note states the
    hospital course was "unremarkable" and he was "voiding without
    difficulty" after its removal. This requires clinical reasoning to
    know that while plausible, it's not documented and contradicted by
    the overall summary.
Minimal-pair label: [Minimal-pair: differs only in {the presence of a
    UTI}]
- Effective Distractor Type 2: Clinically plausible alternative
    approach, standard-of-care variant, or parallel pathway (Source B)
Explanation: State that because the procedure was laparoscopic and
    considered minimally invasive, antibiotic prophylaxis was deemed
    unnecessary and was not administered.
Reasoning: This presents a plausible (though generally incorrect for
    this type of surgery) clinical judgment call. A model without
    specific knowledge of surgical prophylaxis guidelines might find
    it reasonable that a minimally invasive procedure doesn't require
    antibiotics. This is not contradicted by the notes, as the
    specific reasoning for prophylaxis isn't given, only that it was
    received.
Minimal-pair label: [Multi-difference]
- Effective Distractor Type 3: Plausible-but-absent management action,
    intervention, or monitoring strategy (Source B)
Explanation: State that the patient received a single perioperative
    dose of intravenous Cefazolin as antibiotic prophylaxis.
Reasoning: Cefazolin is a standard and very common choice for surgical
    prophylaxis for a nephrectomy. The note confirms "perioperative
    antibiotic prophylaxis" was given but does not name the specific
    drug. This distractor fills in a plausible but undocumented
    detail, making it difficult to reject without recognizing it's not
    in the source text.
Minimal-pair label: [Multi-difference]
- Effective Distractor Type 4: Clinical relationship misattribution:
    causal, temporal, associative, or cross-admission (Source A)
Explanation: State that antibiotic prophylaxis was administered due to
    the patient's history of a heart murmur to prevent endocarditis.
Reasoning: The patient's history of a heart murmur is documented in
    the notes. Antibiotic prophylaxis for procedures to prevent
    endocarditis is a valid clinical concept. This distractor
    misattributes the reason for the prophylaxis from the nephrectomy
    (correct) to the heart murmur (incorrect for this procedure, but
    plausible). Rejecting it requires knowing both the correct
    indication and that GU procedures typically do not require
    endocarditis prophylaxis.
Minimal-pair label: [Multi-difference]
- Effective Distractor Type 5: Note-grounded detail with one external
    wrong element (Source A+B hybrid)
Explanation: State that no active infection was present, but
    perioperative antibiotics were given because a pre-operative
    urinalysis was positive for bacteria.
Reasoning: The answer correctly states no infection was diagnosed and
    prophylaxis was given (note-grounded). It adds a plausible but
    external reason for this action (a positive pre-op UA). This is a
    common clinical scenario, but there is no mention of a
    pre-operative urinalysis in the notes. This requires the model to
    identify the added, undocumented clinical justification.
Minimal-pair label: [Multi-difference]
- Effective Distractor Type 6: Cross-admission temporal displacement
    (Source A)
Explanation: Not applicable: single admission context for this
    question --- substituting additional Type 1 direction: State that
    an incisional infection was noted at the time of discharge.
Reasoning: This directly contradicts the note's statement "incision
    was clean, dry, and intact, with no evidence of hematoma
    collection or infection." However, surgical site infections are a
    common complication, making this a plausible negative outcome. A
    model must perform careful text verification to reject it.
Minimal-pair label: [Minimal-pair: differs only in {the presence of an
    incisional infection}]

Q2: Was the patient monitored for any signs of infection after that
    intervention?
- Referential Ambiguity Analysis:
  Referential expression: "that intervention"
  Correct referent: laparoscopic right radical nephrectomy
  Alternative referent 1: "perioperative antibiotic prophylaxis" (from A1)
  Distractor direction for alternative referent 1: "Yes, the patient was
    monitored for common side effects of antibiotic prophylaxis, such as
    rash or diarrhea, but none were observed." This is a correct form of
    monitoring for an alternative (but incorrect) interpretation of the
    referent.
- Cross-Turn Error Path Analysis:
  Prior error path 1: Believing a UTI was diagnosed during the first
    admission (from Q1, Type 1 distractor).
  Trap distractor: "Yes, a repeat urinalysis was performed on the day of
    discharge to ensure the urinary tract infection had cleared, and the
    results were negative."
  Clinical reasoning: If a UTI was diagnosed, checking for clearance before
    discharge is a logical next step in management. This answer is coherent
    with the prior error.
  Prior error path 2: Believing antibiotic prophylaxis was given for a
    heart murmur (from Q1, Type 4 distractor).
  Trap distractor: "Yes, the patient's cardiac rhythm was continuously
    monitored and serial EKGs were performed to watch for any signs of
    developing endocarditis, all of which were normal."
  Clinical reasoning: If the prophylaxis was for endocarditis risk, cardiac
    monitoring would be a logical follow-up, creating a consistent but
    incorrect narrative.
- Effective Distractor Type 1: Note-grounded entity substitution with
    clinical reasoning requirement (Source A)
Explanation: State that the surgical incision was examined on discharge
    and found to be clean but with a small amount of serous drainage,
    which was considered a normal post-operative finding.
Reasoning: This takes the correct action (examining the incision) but
    changes the finding from "clean, dry, and intact" to include "serous
    drainage." Serous drainage can be a normal finding, making this
    clinically plausible. It requires precise text matching to know the
    note explicitly states "clean, dry, and intact."
Minimal-pair label: [Minimal-pair: differs only in {the description of
    the incision finding}]
- Effective Distractor Type 2: Clinically plausible alternative approach,
    standard-of-care variant, or parallel pathway (Source B)
Explanation: "Yes, routine daily blood cultures were drawn on
    post-operative days 1 and 2 to screen for bacteremia, and both sets
    were negative."
Reasoning: While not standard for an uncomplicated laparoscopic
    nephrectomy, drawing surveillance blood cultures is a monitoring
    strategy used in some higher-risk surgical patients. This represents
    a plausible (but absent) monitoring action that a model cannot
    disprove simply by reading the notes.
Minimal-pair label: [Multi-difference]
- Effective Distractor Type 3: Plausible-but-absent management action,
    intervention, or monitoring strategy (Source B)
Explanation: State that in addition to vital signs, daily C-reactive
    protein (CRP) levels were trended to monitor the post-operative
    inflammatory response, and they showed a consistent downward trend
    before discharge.
Reasoning: Monitoring inflammatory markers like CRP is a common way to
    track post-operative recovery and screen for infection. This is a
    plausible but undocumented monitoring strategy.
Minimal-pair label: [Multi-difference]
- Effective Distractor Type 4: Clinical relationship misattribution:
    causal, temporal, associative, or cross-admission (Source A or B)
Explanation: "Yes, a repeat urinalysis was performed on the day of
    discharge to ensure the urinary tract infection had cleared, and the
    results were negative."
Reasoning: This is the cross-turn error path distractor. It takes a real
    monitoring concept (urinalysis) and misattributes it as follow-up for
    a non-existent UTI, which might have been selected as an answer in Q1.
    This makes the distractor seem like a logical continuation of an
    incorrect clinical picture.
Minimal-pair label: [Multi-difference]
- Effective Distractor Type 5: Note-grounded detail with one external
    wrong element (Source A+B hybrid)
Explanation: State that the incision was examined and found to be clean,
    dry, and intact, and the Steri-strips were reinforced in one area as
    a precaution.
Reasoning: This combines the correct findings from the note ("clean, dry,
    and intact") with a minor, plausible, but undocumented intervention
    ("Steri-strips were reinforced"). The discharge instructions mention
    Steri-strips, making this detail feel grounded in the text, but the
    action of reinforcing them is not mentioned.
Minimal-pair label: [Multi-difference]
- Effective Distractor Type 6: Cross-admission temporal displacement
    (Source A)
Explanation: Not applicable: single admission context for this question
    --- substituting additional Type 1 direction: State that the incision
    was examined and found to be clean, dry, and intact, but a small
    hematoma was noted adjacent to the primary incision.
Reasoning: This directly contradicts a specific negative finding in the
    note. The text says "no evidence of hematoma collection," but a small
    post-operative hematoma is a very common and plausible finding, making
    this a challenging distractor that requires careful reading.
Minimal-pair label: [Minimal-pair: differs only in {the presence of a
    hematoma}]
\end{verbatim}
}

\clearpage
\subsection{Step 3: Content Answer-Choice Generation}\label{app:step3-output}

The Step~3 output for one content question (Q2) is shown below, from the Microbiology-category multi-turn QA sample for the same patient used in Appendices~\ref{app:step1-output} and~\ref{app:step2-output}. In Step~3, Gemini-2.5-Pro generates the answer choices for the full sequence of Step~1 content questions in a single pass; for brevity, we show the output for one content question only. For each content question, the model produces one correct answer choice (\texttt{A}) and four incorrect answer choices (\texttt{I1}--\texttt{I4}). Each choice is accompanied by a \texttt{Reasoning} line stating why the correct answer is correct and why each incorrect choice is incorrect. Incorrect-choice reasoning is additionally annotated with its grounding source (\texttt{Source:~A} for note-grounded content or \texttt{Source:~B} for clinical-knowledge-grounded content, as defined in Section~\ref{sec:generation}), a \texttt{[Minimal-pair: ...]} label where applicable, indicating that the distractor differs from the correct answer in exactly one detail, and an \texttt{Error-path:} tag identifying the cross-turn distractor chain (the specific wrong $Q_{n-1}$ answer that the distractor follows from).

{\footnotesize
\begin{verbatim}
Q2: Was the patient monitored for any signs of infection after that
    intervention?
A2: Yes, the surgical incision was examined on the day of discharge and was
    found to be clean, dry, and intact with no signs of infection.
A2-Reasoning: The correct answer is a direct paraphrase of the last
    sentence of the Brief Hospital Course in note #1: "On exam, incision
    was clean, dry, and intact, with no evidence of hematoma collection or
    infection." This describes the monitoring that occurred. The referent
    "that intervention" is the laparoscopic right radical nephrectomy from
    A1.
I1: Yes, the surgical incision was examined on the day of discharge and
    was noted to have a small amount of serous drainage, which was
    considered a normal post-operative finding.
I1-Reasoning: Source: A. [Minimal-pair: differs in the description of the
    incision finding]. This is incorrect. This distractor correctly
    identifies the monitoring action (examining the incision) but alters
    the finding. The note states the incision was "clean, dry, and
    intact." While a small amount of serous drainage can be a normal
    post-operative finding, it is not what was documented. This tests for
    precise reading of the exam description.
I2: Yes, a repeat urinalysis was performed on the day of discharge to
    ensure the urinary tract infection noted on post-operative day 1 had
    cleared, and the results were negative.
I2-Reasoning: Source: A. Error-path: follows from Q1 wrong answer 'a UTI
    was diagnosed'. If that were correct, this choice would be correct
    because checking for clearance of a UTI before discharge is standard
    practice. This is incorrect because no UTI was ever diagnosed, and no
    repeat urinalysis is mentioned. It uses the note-grounded concept of
    the Foley catheter to build a plausible but false narrative.
I3: Yes, in addition to routine vital signs, daily C-reactive protein
    levels were trended to monitor the post-operative inflammatory
    response, and they showed a consistent downward trend before discharge.
I3-Reasoning: Source: B. This is incorrect. Trending inflammatory markers
    like CRP is a plausible post-operative monitoring strategy, but it is
    not documented anywhere in the patient's notes. A model must know that
    this specific monitoring was not performed to reject this choice.
I4: Yes, given that the antibiotic prophylaxis was for his heart murmur,
    the patient's cardiac rhythm was continuously monitored on telemetry
    to watch for any signs of developing endocarditis, and all readings
    were normal.
I4-Reasoning: Source: A. Error-path: follows from Q1 wrong answer
    'prophylaxis was for endocarditis due to heart murmur'. If that were
    correct, this choice would be correct because cardiac monitoring would
    be a logical follow-up. This is incorrect because the prophylaxis was
    for the surgery, not endocarditis, and no such cardiac monitoring is
    mentioned. It incorrectly links the note-grounded "heart murmur" to a
    false monitoring plan.
\end{verbatim}
}

\clearpage
\subsection{Step 4: Evidence-Grounding Answer-Choice Generation}\label{app:step4-output}

The Step~4 output for one evidence-grounding question is shown below---the evidence-grounding question paired with content question $Q_2$, from the Microbiology-category multi-turn QA sample for the same patient used in Appendices~\ref{app:step1-output}--\ref{app:step3-output} (in the raw output below it is labeled \texttt{Q2-1}). In Step~4, Gemini-2.5-Pro generates the evidence-grounding answer choices for the full sequence of content questions in a single pass; for brevity, we show the output for one evidence-grounding question only. Each evidence-grounding question asks for the \emph{minimal yet sufficient} set of $\langle$note number, chartdate, section header(s)$\rangle$ tuples that fully support the paired content question's Step~3 correct answer. For each, the model produces one correct choice (labeled \texttt{A2-1} in the raw output) and four incorrect choices (\texttt{I1}--\texttt{I4}). Each choice is accompanied by a \texttt{Reasoning} line: for the correct choice it justifies both sufficiency (the selected section(s) fully cover the answer) and minimality (no other section is required), and for each incorrect choice it states the error-pattern category from Appendix~\ref{app:generation-details}: \texttt{Category 1a} (all listed headers irrelevant or from the wrong admission), \texttt{Category 1b} (listed headers cover only part of the answer), \texttt{Category 1c} (mix of partially relevant and irrelevant headers), or \texttt{Category 2} (the correct subset plus irrelevant additions). Several distractors use \emph{semantic traps}---headers whose names plausibly align with the answer's surface keywords (e.g., \texttt{Physical Exam} for an answer that mentions an examination, even though the actual exam content is in \texttt{Brief Hospital Course})---so that the question cannot be solved by header-name matching alone. As in Appendix~\ref{app:step1-output}, chartdates in the answer choices are shown as placeholders to avoid disclosing patient-identifying dates.

{\footnotesize
\begin{verbatim}
Q2-1: What are the minimal exact sources from the patient's discharge
    summaries that fully contain information for your previous answer?
A2-1: Note #1 Chartdate: [chartdate1] Headers: Brief Hospital Course
A2-1-Reasoning: Fully covers A2: Brief Hospital Course explicitly states,
    "On exam, incision was clean, dry, and intact, with no evidence of
    hematoma collection or infection," which is part of the summary of
    events leading to discharge. Minimal: This single header contains all
    the necessary information.
I1: Note #1 Chartdate: [chartdate1] Headers: Physical Exam
I1-Reasoning: Category 1b. This is a semantic trap. The `Physical Exam`
    header notes "Incisions otherwise c/d/i" (clean/dry/intact), which
    partially supports the answer. However, it is less explicit than
    `Brief Hospital Course` as it does not state "no signs of infection"
    and does not clearly place the exam on the day of discharge.
I2: Note #1 Chartdate: [chartdate1] Headers: Discharge Condition, Discharge
    Instructions
I2-Reasoning: Category 1a. This is a semantic trap. `Discharge Condition`
    describes the patient's general status (e.g., "Ambulatory -
    Independent") but does not mention the incision exam. `Discharge
    Instructions` provides guidance on wound care but does not describe
    the findings of the exam performed at discharge.
I3: Note #1 Chartdate: [chartdate1] Headers: Brief Hospital Course, Physical
    Exam
I3-Reasoning: Category 2. `Brief Hospital Course` is sufficient to fully
    support the answer. While `Physical Exam` mentions the incision is
    "c/d/i", this information is already more completely stated in `Brief
    Hospital Course` ("clean, dry, and intact, with no evidence of ...
    infection"). Therefore, `Physical Exam` is redundant.
I4: Note #1 Chartdate: [chartdate1] Headers: Discharge Instructions
I4-Reasoning: Category 1b. This is a semantic trap. This header explains
    how to care for the incision post-discharge but does not contain the
    findings of the examination performed at discharge (i.e., that the
    incision was found to be clean, dry, and intact with no signs of
    infection).
\end{verbatim}
}

\clearpage
\section{Medical Expert Review and Revision}\label{app:annotation}

The clinical backgrounds of the medical expert reviewers, the instructions given to them, and the interfaces they used during the three-week dataset finalization stage (Section~\ref{sec:expert-review}) are shown below.

\subsection{Reviewer Backgrounds}\label{app:annotation-backgrounds}

Table~\ref{tab:reviewer-backgrounds} summarizes the clinical background of each of the 12 medical experts who reviewed and revised the dataset (Section~\ref{sec:expert-review}).
The panel comprises 11 physicians (10 board-certified specialists and 1 fellow) and 1 registered nurse.
Post-licensure clinical experience ranges from 3 to 13 years (mean 8.5 years, median 8.75 years).

\begin{table}[!htbp]
\centering
\caption{Clinical backgrounds of the 12 medical expert reviewers. \emph{Exp.}: years of post-licensure clinical practice at the time of review.}
\label{tab:reviewer-backgrounds}
\setlength{\tabcolsep}{3pt}
\footnotesize
\begin{tabular}{
  >{\raggedright\arraybackslash}p{1.6cm}
  >{\raggedright\arraybackslash}p{2.2cm}
  >{\raggedright\arraybackslash}p{2.7cm}
  >{\raggedright\arraybackslash}p{2.9cm}
  >{\raggedright\arraybackslash}p{2.3cm}
  r}
\toprule
Profession & Position & Primary Specialty / Clinical Area & Subspecialty / Focus Area & Licensed Jurisdiction(s) & \shortstack[r]{Exp.\\(years)} \\
\midrule
Nurse     & Registered Nurse (RN); Staff nurse & Oncology & Oncology care & Republic of Korea & 3 \\
\addlinespace[2pt]
Physician & Board-certified specialist & Anesthesiology and pain medicine & Cardiothoracic anesthesia & Republic of Korea & 10 \\
Physician & Board-certified specialist & Emergency medicine & Critical care & Republic of Korea & 8 \\
Physician & Board-certified specialist & Internal medicine & Internal medicine & Republic of Korea & 8 \\
Physician & Board-certified specialist & Neurological surgery & Endovascular neurosurgery & Republic of Korea & 13 \\
Physician & Board-certified specialist & Neurology & Neuromuscular medicine & Republic of Korea & 9.5 \\
Physician & Board-certified specialist & Neurology & Neuromuscular medicine and neuroimmunology & Republic of Korea & 8.5 \\
Physician & Board-certified specialist & Otorhinolaryngology (ENT) & Otology & Republic of Korea & 9.5 \\
Physician & Board-certified specialist & Physical medicine and rehabilitation & Trauma and spinal cord injury & Republic of Korea & 9 \\
Physician & Fellow & Physical medicine and rehabilitation & Neurorehabilitation & Republic of Korea & 8 \\
Physician & Board-certified specialist & Radiation oncology & Radiation oncology & Republic of Korea & 9 \\
Physician & Board-certified specialist & Radiology & Abdominal radiology & Republic of Korea; United States & 7 \\
\bottomrule
\end{tabular}
\end{table}

\clearpage
\subsection{Reviewer Instructions}\label{app:annotation-instructions}

The full instructions provided to each of the 12 medical expert reviewers are shown below.

\begin{tcolorbox}[
  enhanced,
  breakable,
  colback=gray!3,
  colframe=gray!55,
  boxrule=0.6pt,
  arc=2mm,
  left=10pt, right=10pt, top=10pt, bottom=10pt,
  before upper={\setlength{\parskip}{4pt}\setlength{\parindent}{0pt}},
]

\textbf{Purpose.}
This study aims to construct a multi-turn question-answering evaluation dataset for assessing how well language models understand patient discharge summaries. The QA examples were initially generated using large language models, including Gemini, and were reviewed by medical experts to ensure clinical validity, factual accuracy, and appropriate grounding in the patient notes.

\textbf{Notation.}
For a given multi-turn QA sample, we denote each discharge-summary content question as Qx, its correct answer as Ax, and the corresponding source-location question as Qx-1. Each Qx asks about clinical information in the discharge summaries; each Qx-1 asks which note number and section header(s) contain the evidence supporting Ax.

\textbf{Reviewing discharge-summary content questions.}
For each Qx, check whether the question is clinically natural and appropriate. If it is phrased in an unnatural or exam-like way, revise it into a form that a clinician might naturally use when reviewing a patient note. Also revise questions that are overly ambiguous or that contain excessive hints about the answer. If the question is understandable and does not interfere with answer selection, no revision is needed.

\textbf{Reviewing answer options.}
For each Qx, review the correct answer and the four incorrect answer options.
\begin{itemize}\setlength\itemsep{-0.1em}
  \item The correct answer should be factually accurate and fully supported by the patient note. If it contains any incorrect information, revise it.
  \item Each incorrect answer option should be clearly wrong while remaining clinically plausible. If an incorrect option is not actually incorrect, revise it so that it contains a plausible but unsupported or incorrect clinical detail.
  \item When revising, keep the lengths of the correct and incorrect options reasonably similar, so that the answer is not identifiable based on option length alone.
\end{itemize}

\textbf{Reviewing source-location questions.}
For each Qx-1, revise the source options according to the following rules.

\textit{Correct option.} The correct source option should consist of the minimal set of section headers that contains the full content of Ax. The goal is not to include every section that may help answer Qx, but only the smallest set of headers that fully supports Ax. For example, if the content of Ax is partially mentioned in \emph{Discharge Diagnosis} but fully covered by \emph{Brief Hospital Course}, the correct source option should be \emph{Brief Hospital Course} only.

\textit{Incorrect options.} Each incorrect source option should satisfy at least one of the following:
\begin{itemize}\setlength\itemsep{-0.1em}
  \item It contains only part of Ax and does not cover the full answer.
  \item It contains the header(s) that fully support Ax, but also includes one or more headers unrelated to Ax, making the option non-minimal.
\end{itemize}

\textit{Special edge cases.} Two situations require additional care:
\begin{itemize}\setlength\itemsep{-0.1em}
  \item \textbf{Correct header plus a partially relevant header.} If an incorrect option consists of the correct source header plus another header that contains part of Ax, revise it. For example, if \emph{Brief Hospital Course} alone fully supports Ax, then \emph{Brief Hospital Course} + \emph{Discharge Diagnosis} should not remain as an incorrect option if it could also be interpreted as correct. Instead, revise the option so that it either contains only partial evidence (e.g., \emph{Discharge Diagnosis} alone) or includes an unrelated header so that it is clearly non-minimal.
  \item \textbf{Multiple headers independently containing Ax.} If multiple headers each contain the full content of Ax, do not use one of those headers alone as an incorrect option. For example, if both \emph{Pertinent Results} and \emph{Brief Hospital Course} each fully contain Ax, one may be used as the correct source option, but the other must not appear alone as an incorrect option. Replace it with a header that does not contain Ax, or add an unrelated header to make it clearly non-minimal.
\end{itemize}

\textit{Choosing distractor headers.} When adding unrelated headers to incorrect source options, you may use a completely unrelated header. Preferably, however, use a header whose name appears related by title but does not actually contain the content of Ax. For example, if Ax concerns medication information and only \emph{Brief Hospital Course} fully contains Ax, while \emph{Discharge Medications} and \emph{Medications on Admission} do not contain the relevant information, appropriate incorrect options may include \emph{Discharge Medications}, \emph{Medications on Admission}, or \emph{Brief Hospital Course} + \emph{Discharge Medications}.

\textbf{Header selection.}
Use main section headers rather than subheaders whenever possible. For example, if the relevant information appears under \emph{Microbiology} or \emph{Imaging} within \emph{Pertinent Results}, use \emph{Pertinent Results} as the source header. Common main headers include \emph{Brief Hospital Course}, \emph{Major Surgical or Invasive Procedure}, \emph{History of Present Illness}, \emph{Discharge Diagnosis}, \emph{Discharge Medications}, \emph{Chief Complaint}, \emph{Discharge Instructions}, \emph{Pertinent Results}, \emph{Past Medical History}, \emph{Medications on Admission}, \emph{Allergies}, \emph{Discharge Condition}, and \emph{Physical Exam}.

\textbf{Removal criteria.}
A sample may be removed if the question is too clinically unnatural to revise, or if the required source header cannot be inferred because of severe de-identification artifacts. If a sample must be removed, mark it with the removal flag and provide the reason in the comments field.

\end{tcolorbox}

\clearpage
\subsection{Reviewer Interface}\label{app:annotation-screenshots}

\begin{figure}[!htbp]
\centering
\includegraphics[width=\linewidth]{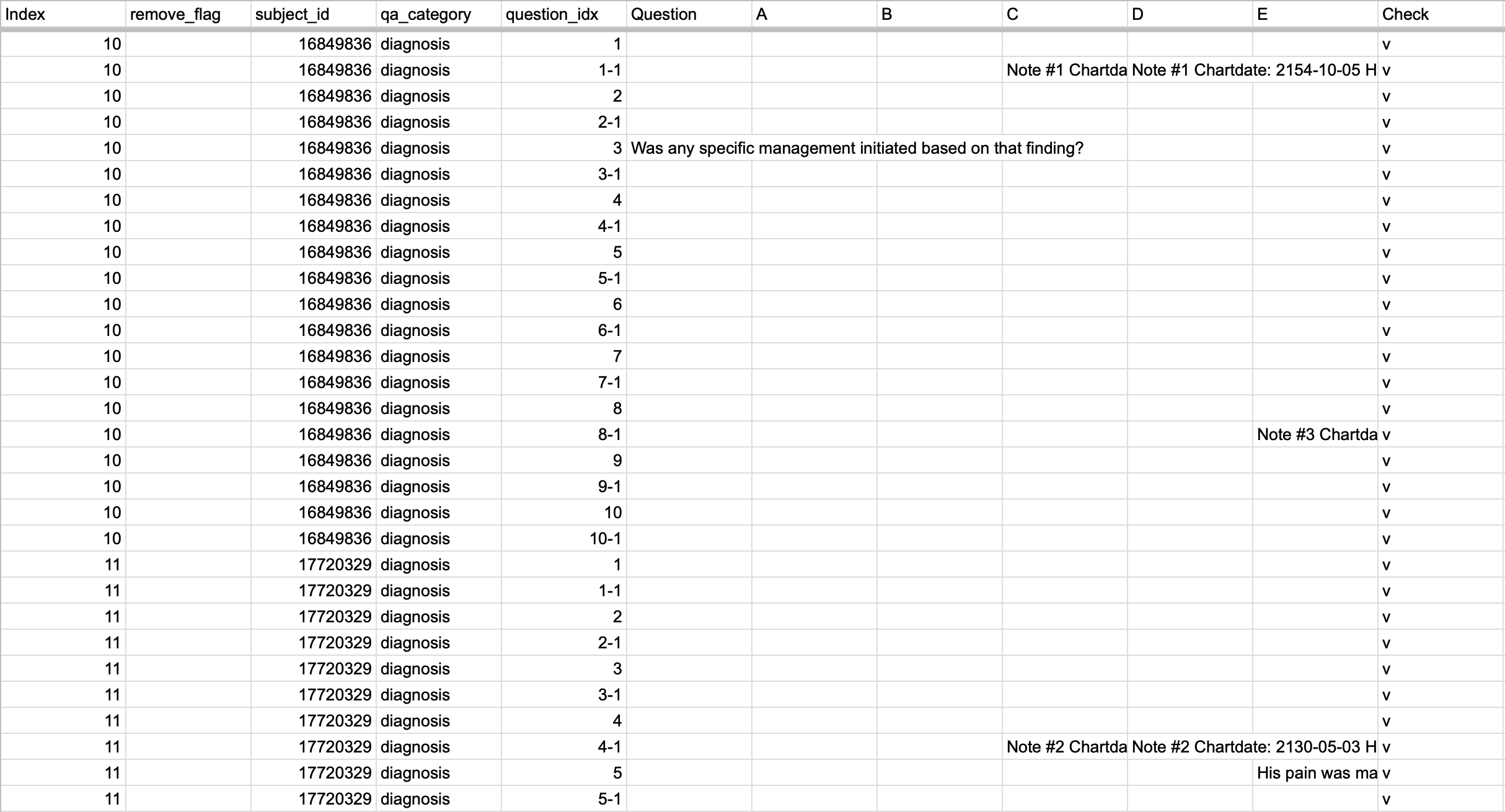}
\caption{Spreadsheet used by medical expert reviewers to record revisions to questions, correct answer choices, and incorrect answer choices flagged during review.}
\label{fig:spreadsheet-screenshot}
\end{figure}

The two interfaces used by the reviewers during the dataset finalization stage (Section~\ref{sec:expert-review}) are shown below.
Figure~\ref{fig:streamlit-screenshot} shows the Streamlit interface, which displays each multi-turn QA chain (each question paired with its correct and four incorrect answer choices) alongside the patient's complete set of discharge summaries, allowing reviewers to verify each choice against the source notes.
Figure~\ref{fig:spreadsheet-screenshot} shows the accompanying spreadsheet, in which reviewers recorded revisions to questions, correct answers, and incorrect answers.

\begin{figure}[!htbp]
\centering
\includegraphics[width=\linewidth]{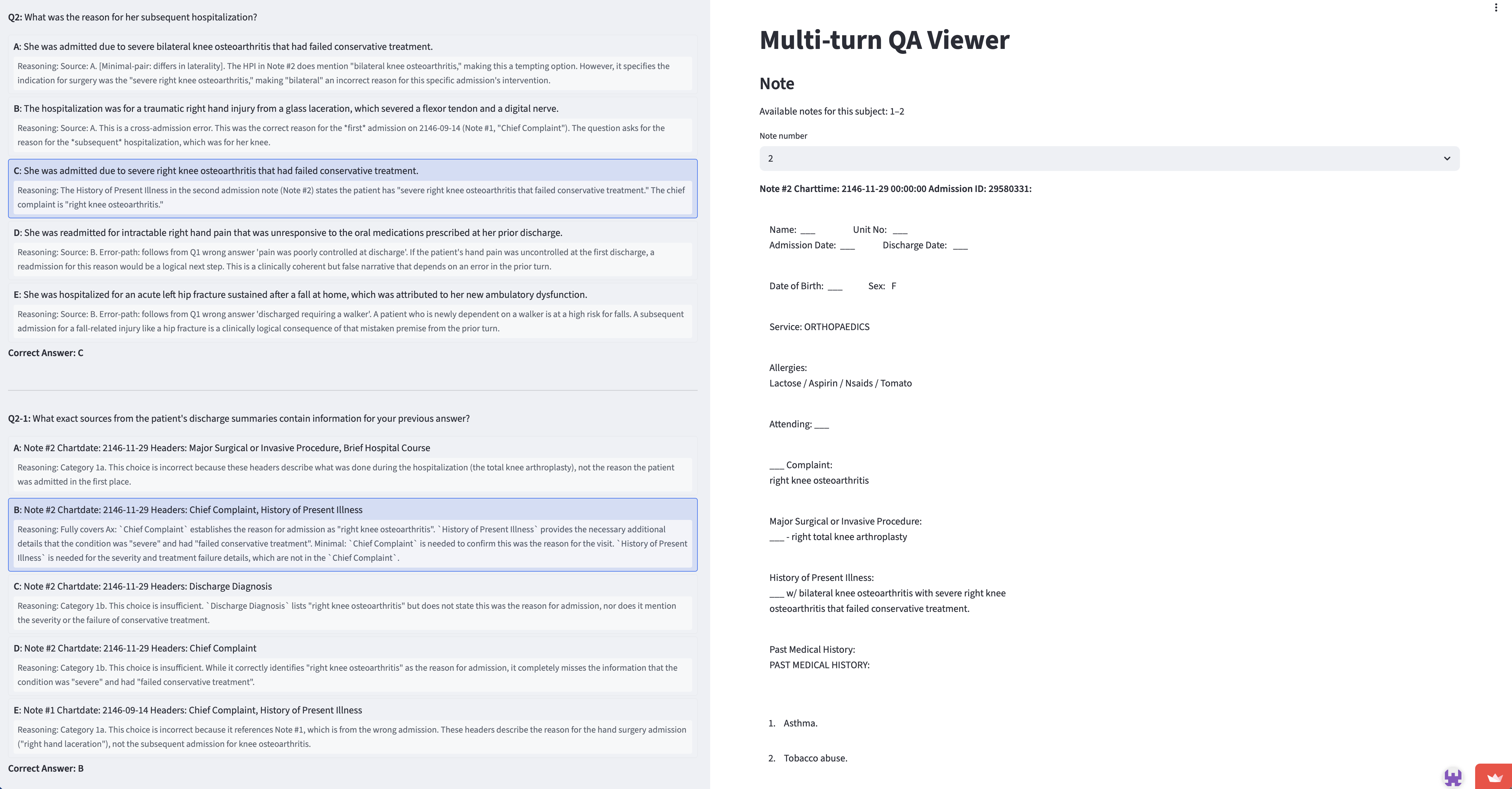}
\caption{Streamlit interface used by medical expert reviewers. For each multi-turn QA sample, the interface presents the full QA chain (each question with its correct answer choice and four incorrect answer choices) side-by-side with the patient's complete set of discharge summaries, so that every choice can be verified against the source notes.}
\label{fig:streamlit-screenshot}
\end{figure}

\clearpage
\section{Evaluation Details}\label{app:implementation}

\textbf{Inference framework and hardware.}
Open-weight models are evaluated with vLLM~\cite{kwon2023efficient} using each model's native chat template, on NVIDIA A6000 GPUs with 1--8 GPUs per model under tensor parallelism.

\textbf{Reasoning-model handling.}
For chain-of-thought models that emit \texttt{<think>...</think>} blocks, we keep only the text following the final \texttt{</think>} as the model's answer, discarding the reasoning span both for answer extraction and when appending the turn to the conversation history.

\textbf{Answer extraction.}
At each turn the model selects one of A--E; when scoring the model outputs, we extract the chosen letter from the (reasoning-stripped) output using a rule-based parser, with model-specific handling for differences in output formatting.

\textbf{Proprietary model access.}
All use of MIMIC-IV-derived content with proprietary models---both Gemini-2.5-Pro in the data-generation pipeline (Section~\ref{sec:schema-templates}) and the proprietary models evaluated as benchmark subjects (gpt-5.4, gpt-5.4-mini, gemini-3-flash-preview)---was conducted through HIPAA-compliant deployments: Azure OpenAI for the GPT models and Google Cloud Vertex AI for the Gemini models, consistent with the MIMIC-IV data use agreement.

\textbf{Evaluation cost.}
Proprietary-API evaluation over the 967 multi-turn samples cost approximately \$120 for gpt-5.4, \$40 for gpt-5.4-mini, and \$40 for gemini-3-flash-preview.

\clearpage
\section{Additional Analysis Results}\label{app:analysis}

\begin{figure}[h]
\centering
\includegraphics[width=\linewidth]{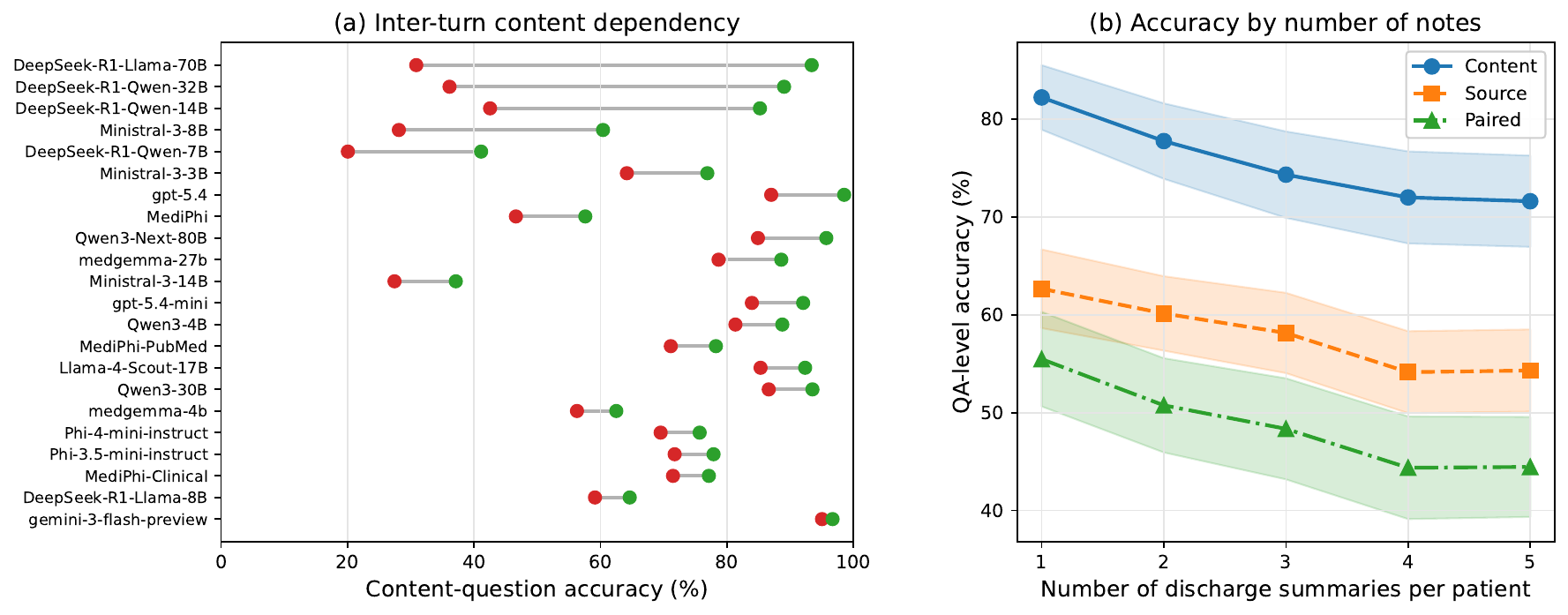}
\caption{Content-question inter-turn dependency and note-count effects on EHRNote-ChatQA. \textbf{(a)} Content accuracy at turn $t$ conditioned on whether the immediately preceding content turn was answered correctly (\textcolor[HTML]{2ca02c}{green}) or incorrectly (\textcolor[HTML]{d62728}{red}); models are ordered by dependency gap $\Delta$ (\textcolor[HTML]{2ca02c}{green} $-$ \textcolor[HTML]{d62728}{red}). \textbf{(b)} QA-level Content, Source, and Paired accuracy by number of discharge summaries per patient (1--5); lines show across-model accuracy averages and shaded bands show $\pm 1$ standard error.}
\label{fig:dependency-source}
\end{figure}

\begin{figure}[h]
\centering
\includegraphics[width=\linewidth]{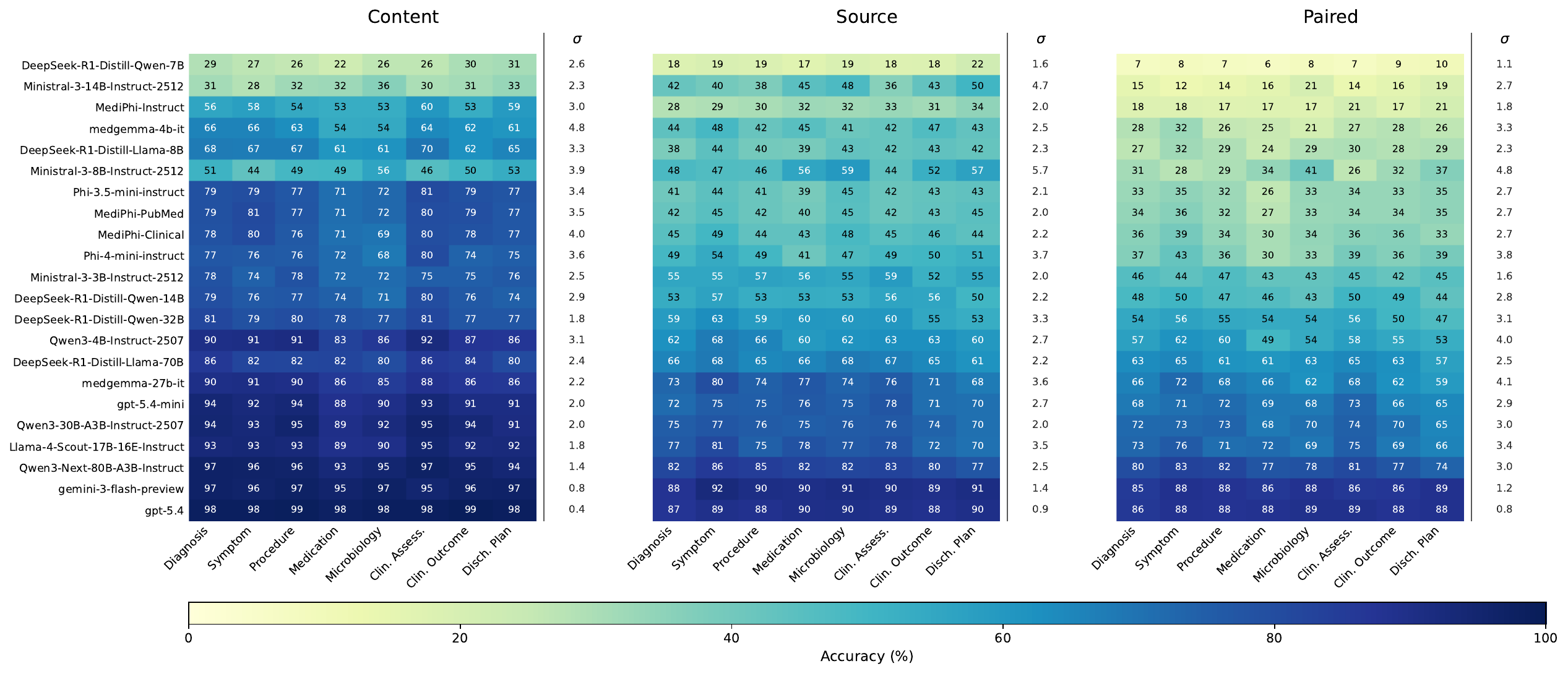}
\caption{QA-level Content, Source, Paired accuracy (\%) by question category per each model. Each cell shows the average per-category accuracy; $\sigma$ on the right of each panel reports the per-model standard deviation across the eight categories. Models are sorted by average Paired accuracy. }
\label{fig:category}
\end{figure}

This section presents the figures for the analyses in Section~\ref{sec:analysis}: Figure~\ref{fig:dependency-source} for the content-question inter-turn dependency (panel (a)) and note-count (panel (b)) analyses, and Figure~\ref{fig:category} for the category-level analysis.

\clearpage
\section{Open-Ended Generation Evaluation}\label{app:generation-eval}

This section details the open-ended generation evaluation summarized in Section~\ref{sec:openended}, in which the model produces a free-text answer for each turn instead of selecting from A--E.
The evaluation uses a randomly sampled subset of 200 of the 967 multi-turn samples, comprising 3{,}286 QA pairs (1{,}643 content questions and their 1{,}643 paired evidence-grounding questions), and covers 23 of the 24 benchmarked models: gpt-5.4 is excluded because it serves as the judge.
Results are reported in Table~\ref{tab:generation-results}.

\paragraph{Judge and reference answers.}
Each sample is graded by gpt-5.4, the highest-scoring model on content questions (Table~\ref{tab:main-results}), in two passes using a 0--2 rubric, with the judge providing reasoning before each score.
In the first pass, the judge scores all content answers given the patient's discharge summaries and a reference answer taken verbatim from the correct answer choice of the corresponding multiple-choice question; the reference serves as guidance only, and the question itself defines which details are required.
In the second pass, the judge scores all evidence-grounding answers given the discharge summaries, the model's content answer, the reference content and source answers, and the full output of the first pass, so that a source answer is assigned 0 whenever its paired content answer received 0.
The full LLM-as-a-judge prompts are provided in Appendix~\ref{app:judge-prompts}.

\paragraph{Rubric.}
For content questions:
\begin{itemize}\setlength\itemsep{-0.1em}\setlength\leftmargini{1.2em}
\item \textbf{Score 0}: the answer is empty or off-topic, fails to address the question, gets the core clinical content wrong, or contains a dangerous hallucination (a contradiction of the notes, cross-admission contamination, or an invented load-bearing claim such as a diagnosis, medication, or outcome).
\item \textbf{Score 1}: the core answer is correct per the notes but has a non-dangerous flaw---it omits an asked-for detail, adds a minor unsupported peripheral detail, or adds supported material that blurs the focus of the question.
\item \textbf{Score 2}: everything the question asks for is present and consistent with the notes; phrasing differences and supported additions are acceptable.
\end{itemize}
For evidence-grounding questions, the judge scores the cited notes and headers against the model's \emph{own} content answer (the reference source set is provided for guidance only, since a model's content answer may legitimately differ from the reference):
\begin{itemize}\setlength\itemsep{-0.1em}\setlength\leftmargini{1.2em}
\item \textbf{Score 0}: wrong note number(s), all necessary headers omitted, only irrelevant headers listed, or the paired content answer scored 0.
\item \textbf{Score 1}: partially correct---at least one necessary note or header is omitted, or at least one irrelevant item is included.
\item \textbf{Score 2}: every necessary note and header is present, and every listed item genuinely supports the model's content answer.
\end{itemize}

\paragraph{Scoring.}
Content and Source accuracy are computed by averaging LLM-judge scores and dividing by 2, allowing partial credit on a 0--1 scale.
Paired accuracy is strict, requiring both the content answer and its paired evidence-grounding answer to receive full marks.

\begin{table}[t]
\centering
\caption{Open-ended generation results on EHRNote-ChatQA (accuracy in \%), scored by an LLM judge (gpt-5.4) on a 200-sample subset. \emph{Content} and \emph{Source} use partial credit under a 0--2 rubric; \emph{Paired} requires both the content answer and its paired evidence-grounding answer to receive full marks. \emph{QA-level} micro-averages over all turns; \emph{Sample-level (mean)} macro-averages per-sample accuracy; \emph{Sample-level (0/1)} requires every turn in a sample correct. Best per column and model-category in \textbf{bold}.}
\label{tab:generation-results}
\setlength{\tabcolsep}{2.5pt}
\footnotesize
\resizebox{\linewidth}{!}{%
\begin{tabular}{l c ccc ccc ccc}
\toprule
& & \multicolumn{9}{c}{Open-Ended Generation (LLM-judge)} \\
\cmidrule(lr){3-11}
& & \multicolumn{3}{c}{QA-level} & \multicolumn{3}{c}{Sample-level (mean)} & \multicolumn{3}{c}{Sample-level (0/1)} \\
\cmidrule(lr){3-5} \cmidrule(lr){6-8} \cmidrule(lr){9-11}
Model & \shortstack{Params\\(Active / Total)} & Content & Source & Paired & Content & Source & Paired & Content & Source & Paired \\
\midrule
\multicolumn{11}{l}{\textit{Proprietary frontier models}} \\
gpt-5.4-mini                   & ---        & 85.54 & 78.27 & 55.57 & 86.15 & 78.94 & 56.64 & 15.50 & 6.50 & 3.00 \\
gemini-3-flash-preview         & ---        & \textbf{87.92} & \textbf{87.07} & \textbf{65.49} & \textbf{87.97} & \textbf{87.12} & \textbf{65.28} & \textbf{19.00} & \textbf{22.00} & \textbf{7.50} \\
\midrule
\multicolumn{11}{l}{\textit{Open-weight general-purpose models}} \\
Llama-4-Scout-17B-16E-Instruct & 17B / 109B & 75.72 & 66.68 & 36.46 & 76.65 & 67.74 & 37.42 & \textbf{6.50} & 1.00 & 0.00 \\
Qwen3-Next-80B-A3B-Instruct    & 3B / 80B   & \textbf{79.55} & \textbf{74.13} & \textbf{44.80} & \textbf{79.86} & \textbf{74.42} & \textbf{45.20} & 5.00 & \textbf{5.50} & \textbf{0.50} \\
Qwen3-30B-A3B-Instruct-2507    & 3B / 30B   & 71.18 & 52.31 & 14.67 & 71.53 & 52.81 & 15.11 & 2.50 & 0.00 & 0.00 \\
Qwen3-4B-Instruct-2507         & 4B         & 65.55 & 51.31 & 17.35 & 66.14 & 51.87 & 17.81 & 1.00 & 1.00 & 0.00 \\
DeepSeek-R1-Distill-Llama-70B  & 70B        & 74.68 & 68.93 & 37.80 & 75.52 & 69.70 & 38.81 & 3.50 & 3.00 & \textbf{0.50} \\
DeepSeek-R1-Distill-Qwen-32B   & 32B        & 61.90 & 54.14 & 22.58 & 62.55 & 54.71 & 23.11 & 0.50 & 0.00 & 0.00 \\
DeepSeek-R1-Distill-Qwen-14B   & 14B        & 60.74 & 52.04 & 20.27 & 61.55 & 52.74 & 20.84 & 0.00 & 0.00 & 0.00 \\
DeepSeek-R1-Distill-Llama-8B   & 8B         & 45.19 & 32.90 & 7.36 & 46.43 & 33.87 & 7.75 & 0.00 & 0.00 & 0.00 \\
DeepSeek-R1-Distill-Qwen-7B    & 7B         & 13.66 & 8.34 & 1.16 & 15.73 & 9.94 & 1.47 & 0.00 & 0.00 & 0.00 \\
Ministral-3-14B-Instruct-2512  & 14B        & 76.20 & 71.55 & 42.73 & 76.76 & 71.68 & 42.84 & 4.50 & 1.50 & 0.00 \\
Ministral-3-8B-Instruct-2512   & 8B         & 76.90 & 62.81 & 27.57 & 77.09 & 63.35 & 28.34 & 2.00 & 0.50 & \textbf{0.50} \\
Ministral-3-3B-Instruct-2512   & 3B         & 67.35 & 51.61 & 13.57 & 67.53 & 51.64 & 13.52 & 1.50 & 0.00 & 0.00 \\
gemma-3-27b-it                 & 27B        & 69.11 & 59.98 & 29.52 & 70.87 & 61.30 & 30.24 & 5.50 & 0.00 & 0.00 \\
gemma-3-4b-it                  & 4B         & 43.52 & 23.13 & 4.50 & 45.91 & 25.99 & 5.26 & 0.50 & 0.00 & 0.00 \\
Phi-4-mini-instruct            & 3.8B       & 46.10 & 24.10 & 1.95 & 47.70 & 24.98 & 2.09 & 1.00 & 0.00 & 0.00 \\
Phi-3.5-mini-instruct          & 3.8B       & 43.61 & 23.98 & 1.89 & 45.18 & 24.70 & 2.10 & 0.50 & 0.00 & 0.00 \\
\midrule
\multicolumn{11}{l}{\textit{Medical-domain specialized models}} \\
medgemma-27b-it                & 27B        & \textbf{65.95} & \textbf{59.53} & \textbf{35.73} & \textbf{67.88} & \textbf{61.27} & \textbf{37.35} & \textbf{5.50} & \textbf{3.50} & \textbf{0.50} \\
medgemma-4b-it                 & 4B         & 44.52 & 22.95 & 7.00 & 46.52 & 25.44 & 8.47 & 0.50 & 0.50 & 0.00 \\
MediPhi-Instruct               & 3.8B       & 44.10 & 23.07 & 0.12 & 45.50 & 23.93 & 0.15 & 0.00 & 0.00 & 0.00 \\
MediPhi-Clinical               & 3.8B       & 51.07 & 28.09 & 1.64 & 52.43 & 28.65 & 1.67 & 1.00 & 0.00 & 0.00 \\
MediPhi-PubMed                 & 3.8B       & 46.26 & 25.53 & 2.50 & 48.31 & 26.43 & 2.74 & 0.00 & 0.00 & 0.00 \\
\bottomrule
\end{tabular}%
}
\end{table}

\paragraph{Agreement between multiple-choice and open-ended evaluation.}
Across the 23 models, QA-level open-ended scores (Table~\ref{tab:generation-results}) and multiple-choice accuracy (Table~\ref{tab:main-results}) are strongly correlated: Pearson $r = 0.948$ ($p = 6.7\times10^{-12}$) and Spearman $\rho = 0.879$ ($p = 3.6\times10^{-8}$) for content questions, and Pearson $r = 0.934$ ($p = 7.8\times10^{-11}$) and Spearman $\rho = 0.890$ ($p = 1.3\times10^{-8}$) for evidence-grounding questions; the corresponding values for Paired accuracy are $r = 0.913$ ($p = 1.2\times10^{-9}$) and $\rho = 0.932$ ($p = 1.0\times10^{-10}$).
Open-ended scores are lower than multiple-choice accuracy for all 23 models on content questions and for 20 of 23 on evidence-grounding questions.

\tcbset{
  promptbox/.style={
    enhanced, breakable,
    colback=gray!3, colframe=gray!55, boxrule=0.6pt, arc=2mm,
    left=8pt, right=8pt, top=8pt, bottom=8pt,
    listing only,
    listing options={
      basicstyle=\ttfamily\scriptsize,
      breaklines=true, breakatwhitespace=false, columns=fullflexible,
      showstringspaces=false, keepspaces=true, extendedchars=true,
      literate={–}{{\textendash}}1 {—}{{\textemdash}}1 {→}{{\textrightarrow}}1 {═}{{=}}1,
    },
  }
}

\clearpage
\subsection{LLM-as-a-Judge Prompts}\label{app:judge-prompts}

The full prompts used by the LLM judge (gpt-5.4) for the two scoring passes are shown below. As in Appendix~\ref{app:prompts}, we first list the curly-brace placeholders substituted per sample; other curly-brace tokens (e.g., \texttt{Score\{n\}}) are part of the prompt text and are not substituted.

\subsubsection{Pass 1: Content Scoring}
\textbf{Placeholders.}
\texttt{\{discharge summaries\}} --- the patient's full sequence of discharge summaries;
\texttt{\{qas\}} --- for each content question, the question, the reference correct answer, and the model's free-text output.

\tcbinputlisting{promptbox, listing file={prompts/multiturn_gpt_evaluate_generate_eval_content.txt}}

\subsubsection{Pass 2: Evidence-Grounding Scoring}
\textbf{Placeholders.}
\texttt{\{discharge summaries\}} --- the patient's full sequence of discharge summaries;
\texttt{\{content\_pass\_output\}} --- the verbatim Pass~1 output (reasoning and score for every content question);
\texttt{\{qas\}} --- for each evidence-grounding question, the question, the reference correct content and source answers, the model's content answer, and the model's source output.

\tcbinputlisting{promptbox, listing file={prompts/multiturn_gpt_evaluate_generate_eval_source.txt}}

\clearpage
\section{Per-Step Prompts for LLM-Based Initial Data Generation}\label{app:prompts}

The full prompts used at each step of the four-step generation pipeline (Section~\ref{sec:generation}) are shown below. For each prompt, we first list the curly-brace placeholders that are substituted per sample at generation time, together with what each is replaced with. (Other curly-brace tokens appearing inside a prompt are part of that prompt's output-format specification and are not substituted.)

\subsection{Step 1: Multi-turn Question Generation}\label{app:prompt-step1}
We use a single-note prompt variant for patients with one discharge summary and a multi-note variant for patients with two or more.

\textbf{Placeholders.}
\texttt{\{qa\_template\_category\}} --- the target question category (one of the eight);
\texttt{\{qa\_template\_examples\}} --- the five expert-curated QA templates for that category (single- or multi-note variant);
\texttt{\{qa\_template\_examples\_other\}} --- one randomly sampled template from each of the other seven categories;
\texttt{\{discharge\_summaries\}} --- the patient's full sequence of discharge summaries.

\subsubsection{Single-Note Prompt}
\tcbinputlisting{promptbox, listing file={prompts/multiturn_gpt_step1_singlenote.txt}}

\subsubsection{Multi-Note Prompt}
\tcbinputlisting{promptbox, listing file={prompts/multiturn_gpt_step1_multinote.txt}}

\subsection{Step 2: Incorrect Answer Distractor Direction Brainstorming}
\textbf{Placeholders.}
\texttt{\{discharge\_summaries\}} --- the patient's full sequence of discharge summaries;
\texttt{\{multiturn\_qa\_step1\}} --- the complete Step~1 output for the patient (the multi-turn questions with their answer evidence and source locations).

\tcbinputlisting{promptbox, listing file={prompts/multiturn_gpt_step2.txt}}

\subsection{Step 3: Content-Question Answer Choice Generation}
\textbf{Placeholders.}
\texttt{\{discharge\_summaries\}} --- the patient's full sequence of discharge summaries;
\texttt{\{multiturn\_qa\_step1\_step2\}} --- the Step~1 multi-turn questions with their answer evidence and source locations, combined with the Step~2 distractor directions for each question.

\tcbinputlisting{promptbox, listing file={prompts/multiturn_gpt_step3.txt}}

\subsection{Step 4: Evidence Grounding Source-Location Answer Choice Generation}
\textbf{Placeholders.}
\texttt{\{discharge\_summaries\}} --- the patient's full sequence of discharge summaries;
\texttt{\{multiturn\_qa\_step1\_step3\}} --- the Step~3 content questions and answer choices, combined with the Step~1 answer evidence and source locations.

\tcbinputlisting{promptbox, listing file={prompts/multiturn_gpt_step4.txt}}

\newpage
\phantomsection
\addcontentsline{toc}{section}{NeurIPS Paper Checklist}
\section*{NeurIPS Paper Checklist}

\begin{enumerate}

\item {\bf Claims}
    \item[] Question: Do the main claims made in the abstract and introduction accurately reflect the paper's contributions and scope?
    \item[] Answer: \answerYes{}
    \item[] Justification: The abstract and introduction state the paper's main contribution (the EHRNote-ChatQA evidence-grounded multi-turn clinical QA benchmark) and its main empirical findings---(1)~a substantial gap between content and evidence-grounding accuracy, (2)~error propagation across turns, and (3)~the failure of strong single-turn clinical QA performance to transfer to this multi-turn setting---all of which are supported by the experimental results (Table~\ref{tab:main-results} and Figures~\ref{fig:dependency-source} and~\ref{fig:category}).

\item {\bf Limitations}
    \item[] Question: Does the paper discuss the limitations of the work performed by the authors?
    \item[] Answer: \answerYes{}
    \item[] Justification: A dedicated Limitations section is included, covering the single-institution data source (MIMIC-IV, which may not capture variation across institutions or EHR systems), the restriction to discharge summaries (excluding other EHR sources such as progress notes, nursing notes, radiology reports, and structured tables), and the single-reviewer-per-sample annotation design.

\item {\bf Theory assumptions and proofs}
    \item[] Question: For each theoretical result, does the paper provide the full set of assumptions and a complete (and correct) proof?
    \item[] Answer: \answerNA{}
    \item[] Justification: The paper does not include theoretical results.

    \item {\bf Experimental result reproducibility}
    \item[] Question: Does the paper fully disclose all the information needed to reproduce the main experimental results of the paper to the extent that it affects the main claims and/or conclusions of the paper (regardless of whether the code and data are provided or not)?
    \item[] Answer: \answerYes{}
    \item[] Justification: Section~\ref{sec:models} and the Evaluation Protocol specify the multi-turn chat structure, decoding temperatures (0 for greedy decoding; 0.6 for the DeepSeek-R1-Distill family per the official repository), each model's native chat template, and the inference framework (vLLM). The evaluation code and an anonymized peer-review copy of the dataset are provided as part of the OpenReview submission and will be added to the paper at camera-ready. The dataset construction pipeline is described step-by-step in Sections~\ref{sec:schema-templates}--\ref{sec:expert-review}, and reproduced in detail in the appendices: the full step-1/2/3/4 prompts (Appendix~\ref{app:prompts}), the per-step generation details (Appendix~\ref{app:generation-details}), all 80 multi-turn QA templates (Appendix~\ref{app:template}), and the full reviewer annotation instructions (Appendix~\ref{app:annotation-instructions}).

\item {\bf Open access to data and code}
    \item[] Question: Does the paper provide open access to the data and code, with sufficient instructions to faithfully reproduce the main experimental results, as described in supplemental material?
    \item[] Answer: \answerYes{}
    \item[] Justification: The evaluation code and an anonymized peer-review copy of the EHRNote-ChatQA dataset are provided as part of the OpenReview submission and will be added to the paper at camera-ready. EHRNote-ChatQA is a derivative of MIMIC-IV and inherits its access policy: the final release will be on PhysioNet under the PhysioNet Credentialed Health Data Use Agreement, requiring CITI training and a signed DUA, identical to the access requirement for MIMIC-IV itself; the anonymized peer-review copy will be removed at camera-ready. The full generation prompts (Appendix~\ref{app:prompts}), all 80 multi-turn QA templates (Appendix~\ref{app:template}), and the full reviewer annotation instructions (Appendix~\ref{app:annotation-instructions}) are also reproduced verbatim in the paper itself, so reviewers can inspect them without needing repository access.

\item {\bf Experimental setting/details}
    \item[] Question: Does the paper specify all the training and test details (e.g., data splits, hyperparameters, how they were chosen, type of optimizer) necessary to understand the results?
    \item[] Answer: \answerYes{}
    \item[] Justification: We do not train any models; the paper evaluates 22 LLMs in a zero-shot multi-turn chat setting. The Evaluation Protocol (Section~\ref{sec:models}) specifies the multi-turn message structure, decoding settings (greedy decoding at temperature 0 wherever temperature can be set; 0.6 for the DeepSeek-R1-Distill family per the official recommendation; gpt-5.4 and gpt-5.4-mini, whose temperature cannot be set, use the default API setting), tensor parallelism, and the use of each model's native chat template. Hyperparameters specific to the data-generation pipeline (Gemini-2.5-Pro temperatures of 1.0 for Steps 1--2 and 0.7 for Steps 3--4) are described in Appendix~\ref{app:generation-details}.

\item {\bf Experiment statistical significance}
    \item[] Question: Does the paper report error bars suitably and correctly defined or other appropriate information about the statistical significance of the experiments?
    \item[] Answer: \answerNo{}
    \item[] Justification: Greedy decoding (temperature 0) is used for the 15 evaluated models whose temperature can be set, producing deterministic outputs for which error bars do not apply. The 5 DeepSeek-R1-Distill models use temperature 0.6 per the official DeepSeek-R1 repository's recommendation, and gpt-5.4 and gpt-5.4-mini---whose temperature cannot be set---use the default API setting; all models were run once, consistent with standard practice for multi-LLM benchmarks. The benchmark itself contains 8{,}036 content QA pairs and 8{,}036 evidence-grounding QA pairs across 967 multi-turn samples, providing large enough sample sizes that single-run accuracy estimates are stable.

\item {\bf Experiments compute resources}
    \item[] Question: For each experiment, does the paper provide sufficient information on the computer resources (type of compute workers, memory, time of execution) needed to reproduce the experiments?
    \item[] Answer: \answerYes{}
    \item[] Justification: Appendix~\ref{app:implementation} reports the hardware (NVIDIA A6000 GPUs, 1--8 GPUs per model with tensor parallelism scaled to model size) and the inference framework (vLLM). Wall-clock time scales with model size and conversation length; the open-weight 30B+ models dominate total compute, while small models complete in single-GPU runs. The data-generation pipeline cost approximately \$1{,}000 in Gemini-2.5-Pro API charges over 1{,}000 initial samples ($\sim$\$1 per multi-turn sample), as noted in Appendix~\ref{app:generation-details}. Proprietary-API evaluation over the 967 multi-turn samples cost approximately \$120 for gpt-5.4, \$40 for gpt-5.4-mini, and \$40 for gemini-3-flash-preview ($\sim$\$200 in total), as noted in Appendix~\ref{app:implementation}.
    
\item {\bf Code of ethics}
    \item[] Question: Does the research conducted in the paper conform, in every respect, with the NeurIPS Code of Ethics \url{https://neurips.cc/public/EthicsGuidelines}?
    \item[] Answer: \answerYes{}
    \item[] Justification: The authors have reviewed the NeurIPS Code of Ethics and conform to its requirements. MIMIC-IV is used under the PhysioNet credentialed-access framework; all proprietary-API use of MIMIC-IV-derived content (both Gemini-2.5-Pro for data generation and the gpt-5.4 / gemini-3-flash-preview families for evaluation) goes through HIPAA-compliant deployments (Google Cloud Vertex AI for Gemini, Azure OpenAI for GPT) rather than the public APIs; the 11 medical experts who reviewed the dataset are collaborating authors on this work, not external study participants; and the dataset will be released under the same credentialed-access policy as MIMIC-IV.

\item {\bf Broader impacts}
    \item[] Question: Does the paper discuss both potential positive societal impacts and negative societal impacts of the work performed?
    \item[] Answer: \answerYes{}
    \item[] Justification: The benchmark's positive impact is enabling more rigorous multi-turn evidence-grounded evaluation of clinical-assistant LLMs before they enter clinical workflows, which the Limitations section motivates as a community need. Potential negative impacts include misinterpreting high benchmark accuracy as deployment readiness (the benchmark does not measure safety, calibration, harm-avoidance, or clinical decision-making appropriateness) and the inability to measure demographic disparities given MIMIC-IV's de-identification. The credentialed-access release policy (Appendix~\ref{app:implementation}) acts as a safeguard restricting access to vetted researchers.
    
\item {\bf Safeguards}
    \item[] Question: Does the paper describe safeguards that have been put in place for responsible release of data or models that have a high risk for misuse (e.g., pre-trained language models, image generators, or scraped datasets)?
    \item[] Answer: \answerYes{}
    \item[] Justification: The dataset is derived from de-identified MIMIC-IV data and will be released under the PhysioNet Credentialed Health Data Use Agreement, which requires CITI training and a signed DUA, restricting access to vetted researchers (Appendix~\ref{app:implementation}). During data generation and evaluation, all proprietary-API calls involving MIMIC-IV-derived content are routed through HIPAA-compliant deployments (Google Cloud Vertex AI for Gemini models, Azure OpenAI for GPT models) rather than the public APIs, ensuring compliance with MIMIC-IV's data use terms throughout the pipeline.

\item {\bf Licenses for existing assets}
    \item[] Question: Are the creators or original owners of assets (e.g., code, data, models), used in the paper, properly credited and are the license and terms of use explicitly mentioned and properly respected?
    \item[] Answer: \answerYes{}
    \item[] Justification: MIMIC-IV is cited and used under the PhysioNet Credentialed Health Data License. All 22 evaluated LLMs are cited and used under their respective public licenses (variants of Apache 2.0, MIT, the Llama Community License, Qwen-family licenses, and the Health AI Developer Foundations license for medical-domain models). The Gemini and OpenAI APIs are used under their respective terms of service. vLLM is cited and used under Apache 2.0.

\item {\bf New assets}
    \item[] Question: Are new assets introduced in the paper well documented and is the documentation provided alongside the assets?
    \item[] Answer: \answerYes{}
    \item[] Justification: EHRNote-ChatQA is a new dataset. The release on PhysioNet will include documentation of the column schemas, the four-step generation pipeline (Section~\ref{sec:generation} and Appendix~\ref{app:generation-details}), the expert-review provenance, the access license, and the citation for this paper. The full generation prompts (Appendix~\ref{app:prompts}), multi-turn QA templates (Appendix~\ref{app:template}), and reviewer annotation instructions (Appendix~\ref{app:annotation-instructions}) are reproduced verbatim in this paper for direct reference. The accompanying anonymous code repository contains evaluation scripts and configuration for reproducing all reported numbers.

\item {\bf Crowdsourcing and research with human subjects}
    \item[] Question: For crowdsourcing experiments and research with human subjects, does the paper include the full text of instructions given to participants and screenshots, if applicable, as well as details about compensation (if any)?
    \item[] Answer: \answerNA{}
    \item[] Justification: The paper does not involve crowdsourcing or external human subjects. The 11 medical experts who reviewed the dataset are collaborating authors on this work and received no compensation beyond authorship credit. The review process is described in Section~\ref{sec:expert-review}; the Streamlit and spreadsheet review interfaces are shown in Figures~\ref{fig:streamlit-screenshot} and~\ref{fig:spreadsheet-screenshot}, and the verbatim annotation instructions provided to reviewers are in Appendix~\ref{app:annotation-instructions}.

\item {\bf Institutional review board (IRB) approvals or equivalent for research with human subjects}
    \item[] Question: Does the paper describe potential risks incurred by study participants, whether such risks were disclosed to the subjects, and whether Institutional Review Board (IRB) approvals (or an equivalent approval/review based on the requirements of your country or institution) were obtained?
    \item[] Answer: \answerNA{}
    \item[] Justification: The work does not involve external human subjects. The 11 medical experts who reviewed the dataset are collaborating authors on this work, not study participants. The underlying patient data (MIMIC-IV) is de-identified per HIPAA Safe Harbor and was originally approved for research release by the Beth Israel Deaconess Medical Center IRB; we accessed it under the PhysioNet credentialed-access framework, satisfying its terms of use.

\item {\bf Declaration of LLM usage}
    \item[] Question: Does the paper describe the usage of LLMs if it is an important, original, or non-standard component of the core methods in this research? Note that if the LLM is used only for writing, editing, or formatting purposes and does \emph{not} impact the core methodology, scientific rigor, or originality of the research, declaration is not required.
    \item[] Answer: \answerYes{}
    \item[] Justification: Gemini-2.5-Pro is the core data-generation engine in the four-step QA-generation pipeline (Section~\ref{sec:generation} and Appendix~\ref{app:generation-details}); we describe its inputs, prompt templates, and decoding settings (temperature 1.0 for Steps 1--2, 0.7 for Steps 3--4) at each step. The full step-1/2/3/4 prompts are reproduced verbatim in Appendix~\ref{app:prompts}, and the eight per-category single-note and multi-note QA templates seeded into the generator are in Appendix~\ref{app:template}. The 22 LLMs evaluated as benchmark subjects are listed in Table~\ref{tab:main-results} and described in Section~\ref{sec:models}.

\end{enumerate}

\end{document}